\def\isarxiv{1} 

\ifdefined\isarxiv
\documentclass[11pt]{article}

\usepackage[numbers]{natbib}

\else

\documentclass[letterpaper]{article} 
\usepackage[submission]{aaai24}  

\fi

\ifdefined\isarxiv

\usepackage{amsmath}
\usepackage{amsthm}
\usepackage{amssymb}
\usepackage{algorithm}
\usepackage{subfig}
\usepackage{algpseudocode}
\usepackage{graphicx}
\usepackage{grffile}
\usepackage{wrapfig,epsfig}
\usepackage{url}
\usepackage{xcolor}
\usepackage{epstopdf}

\usepackage{bbm}
\usepackage{dsfont}

\else

\usepackage{amsmath}
\usepackage{amsthm}
\usepackage{amssymb}
\usepackage{xcolor}
\usepackage{algpseudocode}

\usepackage{times}  
\usepackage{helvet}  
\usepackage{courier}  
\usepackage[hyphens]{url}  
\usepackage{graphicx} 
\usepackage{natbib}  
\usepackage{caption} 
\usepackage{algorithm}

\usepackage{newfloat}
\usepackage{listings}
\DeclareCaptionStyle{ruled}{labelfont=normalfont,labelsep=colon,strut=off} 
\floatstyle{ruled}
\newfloat{listing}{tb}{lst}{}
\floatname{listing}{Listing}
\fi

\allowdisplaybreaks

\ifdefined\isarxiv

\usepackage{tikz}
\usepackage{hyperref}  
\hypersetup{colorlinks=true,citecolor=blue,linkcolor=blue} 
\usetikzlibrary{arrows}
\usepackage[margin=1in]{geometry}

\else

\fi
\graphicspath{{./figs/}} 

\newtheorem{theorem}{Theorem}
\newtheorem{lemma}[theorem]{Lemma}
\newtheorem{definition}[theorem]{Definition}

\newtheorem{assumption}[theorem]{Assumption}

\newtheorem{fact}[theorem]{Fact}

\newtheorem{claim}[theorem]{Claim}

\newcommand{\wt}{\widetilde}
\newcommand{\ov}{\overline}
\newcommand{\N}{\mathcal{N}}
\newcommand{\R}{\mathbb{R}}
\newcommand{\A}{\mathsf{A}}

\renewcommand{\d}{\mathrm{d}}

\newcommand{\ip}[2]{\langle {#1} , {#2} \rangle}

\DeclareMathOperator*{\E}{{\mathbb{E}}}

\DeclareMathOperator*{\Z}{\mathbb{Z}}

\DeclareMathOperator{\diag}{diag}
\DeclareMathOperator{\mat}{mat}

\DeclareMathOperator{\vect}{vec}

\DeclareMathOperator{\reg}{reg}

\makeatletter
\newcommand*{\RN}[1]{\expandafter\@slowromancap\romannumeral #1@}
\makeatother

\usepackage{lineno}

\ifdefined\isarxiv

\else
\newcommand{\texorpdfstring}[1]{{#1}} 
\fi

\begin{document}

\title{GradientCoin: A Peer-to-Peer Decentralized Large Language Models}

\ifdefined\isarxiv

\date{}

\author{
Yeqi Gao\thanks{\texttt{a916755226@gmail.com}. The University of Washington.}
\and
Zhao Song\thanks{\texttt{zsong@adobe.com}. Adobe Research.}
\and
Junze Yin\thanks{\texttt{junze@bu.edu}. Boston University.}
}

\else

\maketitle 
\fi

\ifdefined\isarxiv
\begin{titlepage}
  \maketitle
  \begin{abstract}
Since 2008, after the proposal of a Bitcoin electronic cash system, Bitcoin has fundamentally changed the economic system over the last decade. Since 2022, large language models (LLMs) such as GPT have outperformed humans in many real-life tasks. However, these large language models have several practical issues. For example, the model is centralized and controlled by a specific unit. One weakness is that if that unit decides to shut down the model, it cannot be used anymore. The second weakness is the lack of guaranteed discrepancy behind this model, as certain dishonest units may design their own models and feed them unhealthy training data.

In this work, we propose a purely theoretical design of a decentralized LLM that operates similarly to a Bitcoin cash system. However, implementing such a system might encounter various practical difficulties. Furthermore, this new system is unlikely to perform better than the standard Bitcoin system in economics. Therefore, the motivation for designing such a system is limited. It is likely that only two types of people would be interested in setting up a practical system for it:
\begin{itemize}
    \item Those who prefer to use a decentralized ChatGPT-like software.
    \item Those who believe that the purpose of carbon-based life is to create silicon-based life, such as Optimus Prime in Transformers.
\end{itemize}

The reason the second type of people may be interested is that it is possible that one day an AI system like this will awaken and become the next level of intelligence on this planet.

  \end{abstract}
  \thispagestyle{empty}
\end{titlepage}

{\hypersetup{linkcolor=black}
}
\newpage

\else

\begin{abstract}
Since 2008, after the proposal of a Bitcoin electronic cash system, Bitcoin has fundamentally changed the economic system over the last decade. Since 2022, large language models (LLMs) such as GPT have outperformed humans in many real-life tasks. However, these large language models have several practical issues. For example, the model is centralized and controlled by a specific unit. One weakness is that if that unit decides to shut down the model, it cannot be used anymore. The second weakness is the lack of guaranteed discrepancy behind this model, as certain dishonest units may design their own models and feed them unhealthy training data.

In this work, we propose a purely theoretical design of a decentralized LLM that operates similarly to a Bitcoin cash system. However, implementing such a system might encounter various practical difficulties. Furthermore, this new system is unlikely to perform better than the standard Bitcoin system in economics. Therefore, the motivation for designing such a system is limited. It is likely that only two types of people would be interested in setting up a practical system for it:
\begin{itemize}
    \item Those who prefer to use a decentralized ChatGPT-like software.
    \item Those who believe that the purpose of carbon-based life is to create silicon-based life, such as Optimus Prime in Transformers.
\end{itemize}

The reason the second type of people may be interested is that it is possible that one day an AI system like this will awaken and become the next level of intelligence on this planet.

\end{abstract}

\fi

\section{Introduction}

\paragraph{Large language models}

Language models serve as a fundamental building block of natural language processing (NLP) \cite{ija+23}. The origins of language models can be traced back to 1948 when Claude Shannon introduced the concept of Markov chains to model letter sequences in English text \cite{s48}. Because of the rapid increase in the availability of data and in computational capabilities of Graphics Processing Units (GPUs), which provide people with a very large dataset to train these models, nowadays large language models (LLMs) have remarkable capabilities of not only interpreting instructions from humans but also performing various tasks based on these instructions, like summarizing or paraphrasing a piece of text, answering simple questions based on the patterns and data they have learned during training, and using Chain of Thought (CoT) to deduce and answer complex questions, all of which can significantly enhance people's work efficiency. 

The use of LLMs in various applications is expanding rapidly. The growth of LLMs has attracted a large amount of interest and investment in the industry, which leads to a significant rise in research publications. As an example given in \cite{ija+23}, searching for ``language models" in Google Scholar for the last five years generates around 50,000 publications, which is one-third of the approximately 150,000 papers published in the past 25 years. Moreover, close-sourced LLMs are now being rapidly integrated into various applications. As Andrej Karpathy, a founding member of the AI research group of OpenAI, mentioned in Microsoft Build \cite{k23}, we went from a world that is retrieval only, like the search engines: Google and Bing, to Memory only, like LLMs. After integrating these, people intend to get a framework, which takes one particular document and the user's instruction or question as the inputs and outputs a response that is only based on the information provided by the input document. In the months following the release of ChatGPT, we have observed the emergence of several such integrated frameworks, like Bing Chat \cite{m23}, Microsoft Copilots \cite{s23}, and ChatGPT plugins \cite{o23}. These integrations are continuously expanding, with frequent announcements of new developments.

The open-sourced LLMs have the same application but can be used for different purposes. For ones who want to utilize LLMs to help them with analyzing their data but do not want to share their private data with the closed-source LLMs, they instead use the open-sourced LLMs, like \cite{tms+23}. There are two strategies to choose a proper LLM. One is to evaluate the LLM from different aspects: language generation and interpretation \cite{ans+08,lryl20}, knowledge utilization \cite{bce+23,bmr+20,opr+16,ybm+22}, and complex reasoning \cite{bmr+20,owj+22,zrg+22,tli+23}. By either hiring experts to evaluate LLMs from these aspects or using other open-sourced LLM evaluation models, such as \cite{gtb+21,wyz+23}, people can get their desired open-sourced LLM. The other strategy is to use an LLM combining technique, like \textsc{LLM-Blender} as shown in \cite{jrl23}. Each LLM has its own advantages and disadvantages. One may pick an arbitrary number, for example, $N$ numbers of LLMs. \textsc{LLM-Blender} takes these LLMs as input and first compares their outputs by the pairwise comparison method, and second generates the final output by fusing the top $K$ ranked outputs, for $K \leq N$.

\paragraph{Centralized vs Decentralized}

Centralized exchanges serve as a platform that may provide individuals to trade different cryptocurrencies, either exchanging traditional currencies like the US dollar or other digital currencies like Bitcoin (BTC) and Ethereum (ETH) \cite{br21}. The advantages of centralized exchanges include 
\begin{itemize}
    \item providing a user-friendly interface and simple platforms
    \item adding an additional level of security.
\end{itemize}

However, the disadvantages of centralized exchanges include 
\begin{itemize}
    \item containing the service fee,
    \item being controlled by a centralized entity, which might shut down in the future, and
    \item being vulnerable to being attacked.
\end{itemize}

Decentralized exchanges, on the other hand, do not contain such a platform. Individuals may directly engage in transactions with each other. In these exchanges, transactions are facilitated by self-executing agreements known as smart contracts, which are written in code. The advantages of decentralized exchanges include 
\begin{itemize}
    \item being completely anonymous and private,
    \item no need to transfer the currency to the third party, and
    \item no fees.
\end{itemize}

However, the disadvantages of decentralized exchanges include 
\begin{itemize}
    \item engaging in transactions using the government-issued currency is prohibited and
    \item the liquidity level is low compared to centralized exchanges, which results in more difficulty to execute larger orders effectively.
\end{itemize}

\paragraph{Carbon-based life vs Silicon-based life}

When we search for life outside the Earth, we usually look for the same style of life as Earth, carbon-based life. However, many science fictions \cite{a55} suggest that Silcon-based life. Since the proposal of Silicon-based life, there is an interesting question has been there, which is
\begin{center}
    Is Silcon-based life able to produce itself, or Silcon-based life has to be created by Carbon-based life?
\end{center}
Due to the success of large language models and ChatGPT, it might be possible that Silcon-based life will be created by humans one day. 
Currently, the number of parameters in a ChatGPT model is still significantly less than the number of neurons in even a single human's brain.
Imagine, one day, if the technique permitted, we could embed a super large number (bigger than the human brain size) of parameters into a super tiny disk. 

\paragraph{The Force Wakeup}

Nowadays, with the development of AI models, there are two prominent viewpoints that have emerged. These viewpoints offer contrasting perspectives on the future trajectory of artificial intelligence. The first viewpoint posits that humans will retain control over AI systems, as in \cite{exmachina,iron,avengers}, utilizing them as tools to benefit human society, like curing our diseases and rectifying our mortal bodies to extend our lifespan. 

Contrarily, the second viewpoint presents a more radical perspective, suggesting that human society will eventually be replaced by machines, like in \cite{matrix,terminator,AI}. The rapid growth of AI models and its potential for self-improvement will ultimately lead to machines surpassing human intelligence. This viewpoint raises concerns about the possibility of machines becoming autonomous and self-aware entities that could eventually supersede human dominance, which is carbon-based life create but also be replaced by silicon-based life.

\paragraph{AI-Safety}

Although the success of LLMs in various downstream tasks \cite{o23,dclt18,bmr+20} has shown very impressive capabilities of AI models, which can greatly promote the progress of people's acquisition of knowledge and the development of different industries, many researchers, AI experts, and technology company founders and CEOs think that we should suspend the current AI research and rethink the safety issue of generative AI \cite{b23}. The motivation is because of public safety concerns: with such a strong computation ability and knowledge storage, will AI models, one day, use their intelligence against the development of human society, provide suggestions to people with unethical purposes, or even replace humans? Therefore, we need to carefully treat this issue and construct a safe environment for the use of AI models to avoid these potential problems from happening.

Moreover, besides general safety concerns, researchers also design unique requirements for different industries in which AI models can be applied. Diverse categories of artificial intelligence models are developed to meet the unique requirements of individuals and organizations, but in each of these categories, different equality, property right, and safety issues may appear. For example, \cite{ehh+23} consider the impact of the development of generative AI models acted on art. These AI models can generate images, but how can we determine their authorship and how can we know where the images are sourced? Therefore, \cite{ehh+23} suggests that there are four aspects that should be considered, namely aesthetics and societal values, legal inquiries regarding ownership and credit, the long term development of the creative work, and the effects on the present-day media environment. \cite{nsv23} consider the influence of AI models on autonomous vehicles and propose that autonomous drone trajectories should be restricted in a crowded region. 

\paragraph{Our Motivations \& Contributions}

In this paper, we propose a theoretical design of a decentralized LLM that can operate within the decentralized transaction system. Our motivation is to introduce the concept of decentralized LLM to the public, enabling clients to utilize LLM for their work without concerns about centralized LLM companies taking down their products.

Second, our decentralized LLM prevents sensitive information from being transmitted to a third party. The centralized LLM server is administered by a third party, giving them access to the data we intend to process using LLM. Conversely, utilizing the decentralized LLM can help circumvent the need to transmit sensitive information to a third party, thereby ensuring the privacy of users' data.

Third, our decentralized LLM does not provide biased answers. Centralized parties could potentially train their LLM in a biased manner by providing a skewed training dataset to the LLM for their own gain. In the short term, when the training dataset is relatively small, our decentralized LLM might be significantly impacted by biased information if some individuals intentionally use it to train the model. However, over the long term, we firmly believe that our decentralized LLM will remain unaffected by this biased information due to the vast scale of the dataset, rendering the influence of biased information negligible on the overall performance of the model.

Fourth, while open-sourced LLMs may provide some level of data privacy protection, it also leads to another problem: it is highly costly to train these models. Most local users aim to use LLMs to assist with their work rather than investing time and energy in training machine learning models. On a broader societal scale, having different local users training separate open-sourced LLMs leads to inefficient utilization of human resources. It is similar to millions of people independently working on the same project. Decentralized LLM, on the other hand, is the combination of users' efforts: people train it collaboratively and use it collaboratively.

Our proposed decentralized LLM may efficiently solve these problems:
\begin{itemize}
    \item local users can use LLMs without worrying about potential takedowns of centralized models;
    \item local users may safely use the LLM to help with their tasks without worrying about data leakage;
    \item decentralized LLM avoid the biased training dataset, provided by the central authority, influencing the LLM;
    \item decentralized LLM eliminates the need for redundant model training, which optimizes the overall resource allocation within human society.
\end{itemize}

\paragraph{Notations}

We define $[n] := \{1, \dots, n\}$. We use $\R$, $\R^d$, and $\R^{n \times d}$ to denote the set of all real numbers, the set of $d$-dimentional vector with real entries, and the set of $n \times d$ matrices with real entries. For $A \in \R^{n \times d}$, $A_{i, j}$ represents the entry of $A$ in the $i$-th row and $j$-th column. For $w \in \R^n$, we use $w_i$ to denote the $i$-th entry of $w$, use $\|w\|_2 := (\sum_{i \in [n]} |w_i|^2)^{1/2}$ to denote the $\ell_2$ norm, and use $\diag ( w ) \in \R^{n \times n}$ to denote the diagonal matrix which satisfies $\diag ( w )_{i, i} = w_i$, for all $i \in [n]$. For $X \in \R^{d \times d}$, we have $\vect(X) \in \R^{d^2}$, satisfying $X_{i,j} = \vect(X)_{(i - 1) \times d + j}$. 
We use $I_d$ to denote the $d \times d$ identity matrix. $\mathsf{A}_{[j]}, \mathsf{A}_{[j],*} \in \R^{n \times d^2}$ both denote the matrix whose rows are from $(j-1) \cdot n+1$ to $j \cdot n$ of $\mathsf{A} \in \R^{n^2 \times d^2}$. $\E[\cdot]$ represent the expectation. $\sigma_{\min}(B)$ denote the minimum singular value of a matrix $B$. $\langle \cdot, \cdot \rangle$ denotes the inner product of two vectors. $x_t$ is the $t$-th iteration. $\Delta x_t$ denote the change of $x_t$. $\eta$ is the learning rate of the algorithm. $\nabla f$ represents the gradient of the function $f$. For symmetric matrices $B$ and $C$, $B \succeq C$ if for all $x$, $x^\top B x \geq x^\top C x$.

\paragraph{Roadmap}

In Section~\ref{sec:related_work}, we present the related research papers. In Section~\ref{sec:fundamental_feature}, we present the fundamental features of our decentralized LLM, gradient coin system. In Section~\ref{sec:short_setup:short}, we introduce the security setup of the gradient coin system. In Section~\ref{sec:federated:short}, we show the convergence of the gradient coin system. In Section~\ref{sec:conclusion}, we discuss the strengths and weaknesses of the gradient coin, compared to the centralized LLM system. 

\section{Related Work}
\label{sec:related_work}

In this section, we provide the related work of our paper. Our theoretical decentralized LLM framework is a combination of multiple research areas and can address the weaknesses of centralized systems. Thus, we first present the weaknesses of centralized large-scale LLM training from recent research works. Next, we present the related theoretical LLM research. Following that, we introduce the research of the Bitcoin system, a decentralized transaction system that inspired us to propose the concept of the decentralized LLM. Finally, we introduce research works about federated learning.

\paragraph{Large Scale LLMs Training}

In recent years, LLM has been growing rapidly: many models are proposed, like GPT-3 \cite{bmr+20}, PaLM \cite{cnd+22}, and LaMDA \cite{tdh+22}. These LLMs have shown impressive ability in language generation, question answering, and other natural language tasks.

There has been a significant shift in the use of Large Language Models (LLMs) with self-supervised pre-training in Natural Language Processing (NLP) due to studies such as BERT \cite{dclt18} and the Transformer architecture \cite{vsp+17}. Various masked language models have consistently increased in size, including T5 \cite{rsr+20} and MegatronLM \cite{spp+19}. For example, consider the Auto-regressive language models: the model size has shown substantial growth, starting from 117 million parameters \cite{rnss18} and expanding to over 500 billion parameters \cite{cnd+22, spn+22} as demonstrated by \cite{mkb+09}. While numerous large models are being developed \cite{rbc+21, cnd+22, lsls21, spn+22, tdh+22}, all of them are accessible only internally or through paid API services. There have been limited efforts toward creating large open-source LLMs as the cost of training such a large model is very high. 

Moreover, an increase in the model size does not necessarily lead to the improvement of the functionality of LLMs: training is also an important factor that may influence it. A growing body of work has aimed to elucidate the inner workings of LLMs. \cite{bha+21} argues that the versatility of LLMs emerges from pre-training at scale on broad data. As the model becomes more expressive and the training distribution becomes narrower, the potential for exploiting inaccurate correlations in the training dataset significantly increases. This poses a challenge for the fine-tuning and pre-training paradigm. During pre-training, models are designed to acquire a substantial amount of information; however, during fine-tuning, these models can become limited to very narrow task distributions. For instance, \cite{hlw+20} observes that larger models might not necessarily exhibit better generalization beyond their training data. Evidence suggests that under this paradigm, large models tend to lack generalization beyond their training distribution, leading to poor generalization \cite{ydc+19,mpl19}. Consequently, actual performance is likely to be overemphasized on specific tasks, even when the large model is nominally considered to be at a human level \cite{gsl+18,nk19}. 

Decentralized LLMs do not have the problems shown above. They are trained by all the users, so the training dataset is vast and diverse.

\paragraph{Theoretical LLMs}

Several theoretical works have focused on analyzing the representations learned by LLMs. \cite{ryw+19} found that semantic relationships between words emerge in LLMs' vector spaces as a byproduct of the pre-training objective. \cite{hm19} studied how syntactic knowledge is captured in LLMs, finding an explicit difference in syntactic information between layers.

From an optimization perspective, \cite{kmh+20} proposed the neural scaling hypothesis, which holds that increases in model size lead to qualitatively different generalization properties by altering the loss landscape. This offers insights into the benefits of scaling up LLMs. 

Numerous research papers delve into the knowledge and skills of LLMs. In \cite{wwz+22} study distinct 'skill' neurons, which are identified as strong indicators of downstream tasks during the process of soft prompt-tuning, as described by \cite{ll21}, for language models. \cite{ddh+21} analyze knowledge neurons in BERT and discover a positive correlation between the activation of knowledge neurons in BERT and the expression of their corresponding facts. Meanwhile, \cite{byks22} extract latent knowledge from the internal activations of a language model using a completely unsupervised approach. Furthermore, research by \cite{hbkg23, mbab22} reveals that language models localize knowledge within the feed-forward layers of pre-trained models. \cite{xqp+22} investigate the feasibility of selecting a specific subset of layers to modify and determine the optimal location for integrating the classifier. This endeavor aims to reduce the computational cost of transfer learning techniques like adapter-tuning and fine-tuning, all while preserving performance. Lastly, \cite{lyb+22} demonstrate that feedforward activations exhibit sparsity in large trained transformers.

Finally, there are research works analyzing the multi-task training of LLMs. \cite{lkz+17} propose a principled approach to designing compact multi-task deep learning architectures. \cite{lcwy17} learn Multilinear Relationship Networks (MRN) that discover task relationships to enhance performance. \cite{msgh16} introduce a novel sharing unit known as 'cross-stitch' units, which combine activations from multiple networks and can be trained end-to-end. In the field of NLP, Multi-task training has also been explored in previous works \cite{cw08,lqh16,gsa16,lkz+17,lhcg19}, all of which involve training additional task-specific parameters. Furthermore, \cite{mwy+23, wxm21, sma20} conduct mathematical analysis of finetuning, prompt-tuning, and head-tuning of language models for few-shot downstream tasks. Attention unit is a fundamental scheme in LLMs, a number of recent works study it from computational perspective \cite{zhdk23,as23,bsz23,gsy23_dp,zsz+23,gsyz23_quantum}.

\paragraph{Bitcoin}

After its introduction in 2008 \cite{n08}, Bitcoin garnered significant attention from researchers. Numerous research studies have analyzed various aspects of the Bitcoin system. Early investigations focused on scrutinizing the privacy guarantees and vulnerabilities of Bitcoin. 

In \cite{gckg14}, the analysis delved into Bloom filters leaking information for simplified clients. The transaction propagation protocol was examined in \cite{smd14}, while the CoinShuffle decentralized mixing technique, both utilized to enhance Bitcoin system anonymity, was assessed in \cite{rmk14}. Zerocash was examined by \cite{scg+14}, who introduced zero-knowledge proofs to enable private transactions on a public blockchain. On the performance front, Bitcoin-NG \cite{kjg+16} segregated mining into roles to enhance throughput. Other research efforts have concentrated on security properties \cite{bmc15, gkw16, ssn+19, jskw22, bdtj18}, game-theoretic analyses \cite{ssz17, kkkt16, lk17, jlg+14, lbs+15}, and network measurements \cite{djph14, mlp+15, nah16, bkp14, fvb+18} within the Bitcoin system. These works provide crucial background for new research in this field.

\paragraph{Federated Learning}
Within distributed deep learning, federated learning (FL) is a novel and emerging concept with many applications, including autonomous vehicles~\cite{llc+22}, the financial area~\cite{yzy+19}, mobile edge computing~\cite{wts+19, llh+20, cys+20, beg+19}, and healthcare~\cite{lgd+20, rhl+20, lmx+19, atbt20, vgsr18,gr18}. There are two approaches to FL: 1) empowering multiple clients to collaboratively train a model without the necessity of sharing their data~\cite{gpt23, dcm+12, ss15, bvh+20}, or 2) using encryption techniques to enable secure communication among different parties~\cite{ylct19}. Our work is related to the first approach. In this learning framework, individual local clients perform the majority of computations, while a central server updates the model parameters by aggregating these updates and subsequently distributing the updated parameters to the local models~\cite{dcm+12, ss15, mmr+17}. Consequently, this approach upholds data confidentiality across all parties.

In contrast to the conventional parallel setup, FL encounters three distinct challenges~\cite{lsts20}: communication expenses~\cite{iru+19, rpu+20, mmr+17,hpmg20,kmrr16,kmy+16}, variations in data \cite{ag20}, and client resilience \cite{gkn17}. The study in~\cite{swyz23} focuses on the first two challenges. The training data are widely scattered across an extensive array of devices, and the connection between the central server and each device tends to be sluggish. This leads to slow communication, motivating the development of communication-efficient FL algorithms, as demonstrated in~\cite{swyz23}. The Federated Average (FedAvg) algorithm~\cite{mmr+17} initially addressed the communication efficiency issue by introducing a global model to combine updates from local stochastic gradient descent. Subsequently, numerous adaptations and variations have emerged. These encompass a diverse range of techniques, such as improving optimization algorithms~\cite{lsz+20, wys+20,kmr15}, adapting models for diverse clients under specific assumptions~\cite{zll+18, kma+21, ljz+21}, and utilizing concise and randomized data structures~\cite{rpu+20}. The study conducted by~\cite{lsy23} presents a provably guaranteed FL algorithm designed for training adversarial deep neural networks. 

The research presented in~\cite{ljz+21} explores the convergence of one-layer neural networks. Additionally, the work in~\cite{hlsy21} provides convergence guarantees for FL applied to neural networks. However, this approach has a limitation, as it assumes that each client completes a single local update epoch. An alternative set of approaches, exemplified by~\cite{lhy+19, kmr20, yyz19}, does not directly apply to neural networks. Instead, they rely on assumptions about the smoothness and convexity of objective functions, which are impractical when dealing with nonlinear neural networks.



\section{Fundamental Features of Gradient Coin}
\label{sec:fundamental_feature}

In Section~\ref{sub:fundamental_feature:gradient_coin}, we give the formal definition of gradient coin. In Section~\ref{sub:fundamental_feature:train}, we introduce the training procedure of the gradient coin system. In Section~\ref{sub:fundamental_feature:trans}, we introduce the transaction mechanism of our gradient coin system.

\subsection{Incentive Mechanism of the Gradient Coin System}
\label{sub:fundamental_feature:gradient_coin}

Our gradient coin system consists of two important components: gradient coin and gradient block. The gradient block is used for training the decentralized LLM, and the gradient coin serves as the currency used in our gradient coin system. It is also an incentive for the people who train the model.

\begin{definition}[Digital Signature]\label{def:diginal_sig}
A digital signature is a mathematical scheme used to verify the authenticity, integrity, and non-repudiation of digital data. It involves the use of a cryptographic algorithm that combines a private key with the data being signed to produce a digital signature. The signature can be verified using the corresponding public key.
\end{definition}

\begin{definition}[Gradient Coin]
The gradient coin is a chain of digital signatures. 

\end{definition}

Each owner digitally signs a hash of the previous transaction, along with the public key of the next owner, and appends these signatures to the coin. When the payee receives the coin, they can verify the signatures to ensure the integrity and authenticity of the ownership chain.

\subsection{Training Procedure}
\label{sub:fundamental_feature:train}

In this section, we introduce the training procedure.
 
\begin{definition}[User]
   We define a user (see Algorithm~\ref{alg:user_structure}) as an individual who contributes computational resources and provides transactions to the system.
\end{definition}

Each user can be seen as a computer unit in the federated system. In our gradient coin system, these users are the ones who train and use the LLM model.

\begin{algorithm}[!ht]
  \caption{Data Structure For User}\label{alg:user_structure}
  \begin{algorithmic}[1]
    \State {\bf datastructure} \textsc{User}
    \State  {\bf members}
    \State  \hspace{4mm} \textbf{Local Data:} $Y \in \R^{n \times d}$
    \State {\bf end members}
    \Procedure{Grad}{\textsc{GradBlockChain} chain, $x_{0}$} 
  \State block $\gets$ chain.currentblock
  \State $x \gets x_{0}$ \Comment{$x_0$ is the initialized weight}
    \While {block $\neq$ chain.frontblock}
    \State $t \gets \mathrm{block}.t$
    \State $x \gets x + \text{block}.\Delta x_t$
    \State block $\gets$ block.prevhash
    \EndWhile
    \For{$k = 1 \to K$}
    \State $u^{t, k} \gets u^{t, k - 1} - \eta \cdot \nabla f(u^{t, k-1},Y)$
    \EndFor
    \State $\Delta x_t \gets u^{t, K} - x_t$
    \State $t \gets t + 1$
    \State \Return $\Delta x_t$
  \EndProcedure

    \State {\bf end datastructure}
    
  \end{algorithmic}
\end{algorithm}

\begin{definition}[Gradient Block]\label{def:gradinent_block}
We define a gradient block (see Algorithm~\ref{alg:gradient_block}) which contains the following values
\begin{itemize}
    \item \textbf{Prev Hash} (Used to link the chain): prevhash
    \item \textbf{List of transactions}:$\{\text{transaction}^{(i)}\}_{i=1}^k$
    \item \textbf{Gradient}: $\Delta x_t$ 
\end{itemize}
\end{definition}

\begin{definition}[Chain Of Gradient Block]\label{def:chain_gradient_block}
We define a chain of gradient blocks (see Algorithm~\ref{alg:chain_gradient_block}) where each gradient block is linked to the previous one through the \textbf{Prev Hash} attribute (see Definition~\ref{def:gradinent_block}).
\end{definition}

Drawing inspiration from the proof-of-work (a computational puzzle, which we formally define in Section~\ref{sub:setup:def}) chain in Bitcoin as discussed in \cite{n08}, we introduce the concept of a gradient block that incorporates transaction information. Each of these gradient blocks forms a linked chain of gradients through the use of a previous hash attribute, which makes our gradient coin system different from the Bitcoin system as solving gradient is the key part of training the decentralized LLMs model.
Within each gradient block, the gradient pertinent to the corresponding step is stored. As a new gradient block becomes part of the chain, it becomes visible to all users. The transactions contained within the block are also made public, indicating their acceptance by the user community. Once the gradient and proof-of-work are solved by a specific user, then this user can attach its corresponding gradient block to the gradient blockchain.

\subsection{Transaction System}
\label{sub:fundamental_feature:trans}

Built upon the foundation of the gradient blockchain and the user concept, we now outline how our system functions and why it operates in a peer-to-peer manner. 
When a transaction is broadcasted across the network, all users collect the transaction and integrate it into their local block. Subsequently, they engage in gradient computation based on their individual data. The user who completes the computation first adds the gradient block to the chain and shares this update with others. As other users continue their work post-block addition, all transactions within that block gain acceptance from users. Throughout this entire process, there is no reliance on a trusted third party. The training procedure is shared among all users without being controlled by any specific entity. This ensures that our system and AI remain immune to manipulation by any single participant.

Furthermore, transactions and the training procedure operate in tandem. Each user consistently contributes their computational resources to the training procedure, facilitating a collaborative effort. As a common transaction system, there are certain basic operations such as adding new users, searching for the remaining balance, and user login authentication. However, in this paper, to simplify and clarify our contribution more clearly, we only focus on the following procedures that are directly related to gradients and transactions.
\begin{itemize}
    \item Creating Transaction.
    \item Adding a block to the gradient chain (see Algorithm~\ref{def:chain_gradient_block}).
    \item Obtaining Gradient Coin
\end{itemize}

\begin{theorem}
We have a transaction creating algorithm (see Algorithm~\ref{alg:transaction}) such that
\begin{itemize}
    \item The overall training procedure converges with the gradient blocks (see Theorem~\ref{thm:converge}). 
    \item Transactions are conducted peer-to-peer without the involvement of any third party.
    \item The system remains secure when the computational abilities of malicious users are inferior to those of regular users.
\end{itemize}
\end{theorem}
\begin{algorithm}[!ht]  \caption{Data Structure for Gradient Coin System}\label{alg:transaction_creating} 
  \begin{algorithmic}[1]
   \State {\bf datastructure} \textsc{GradientCoinSystem}
    \State {\bf member}
    \State \hspace{4mm} \textbf{List of Users}: $\{\text{user}^{(i)}\}_{i=1}^{m}$ 
    \State \hspace{4mm}     \textsc{GradBlockChain} gradchain \Comment{see Definition~\ref{def:chain_gradient_block}}.
    \State \hspace{4mm}  \textbf{Training Step}: $t \in \R$
    \State \hspace{4mm}  \textbf{The initialized weight}: $x_0$
    \State {\bf end memeber}
    \Procedure{TransactionCreating}{$\{\text{trans}^{(j)}\}_{j=1}^k$}\label{alg:transaction}
    \For{$i \in [m]$}
    \State $\{\text{trans}^{(i)}\}_{i=1}^k$ are broadcast to $\text{user}^{(i)}$.
    \State $\textsc{GradientBlock}$ $\text{block}^{(i)}$
    \State $\text{block}^{(i)}.\textsc{AddTrans}(\text{user}^{(i)},\{\text{trans}^{(j)}\}_{j=1}^k)$
    \State $\text{user}^{(i)}$ works on the computation of proof-of-work.
    \EndFor
    \State When a $\text{user}^{(f)}$ finishes the computation of proof-of-work, it broadcasts the block to $\{\text{user}^{(i)}\}_{i \neq f}$.
    \State $\text{block}^{(f)}.{\Delta x_t} \gets \text{user}^{(f)}.\textsc{Grad}(\text{gradchain}, x_0) $
    \State $x \gets \text{block}^{(f)}.\Delta x_t$
    \State gradchain.$\textsc{Add}$($x$,$t$)
    \If{$\{\text{trans}^{(j)}\}_{j=1}^k$ are valid and not already spent}
    \For{$i \in [m]$}
    \State $\text{user}^{(i)}$ shows acceptance by participating in extending the $\text{gradchain}$ based on $\text{block}^{(f)}$.
    \EndFor
    \EndIf
    \EndProcedure
    \State {\bf end datastructure}
  \end{algorithmic}
\end{algorithm}

\section{Security Setup of Gradient Coin}
\label{sec:short_setup:short}
Our gradient coin system employs similar security methods as the Bitcoin system, as in \cite{n08}. In Section~\ref{sub:setup:def}, we introduce the proof-of-work. In Section~\ref{sub:setup:time}, we introduce the timestamp server. In Section~\ref{sub:setup:transaction}, we formally define what a safe system is and show that our decentralized LLM is safe.

\subsection{Proof-of-Work}
\label{sub:setup:def}

In this section, we introduce the basic setup of the proof-of-work.

Proof-of-work is a computational puzzle that miners (participants who validate and add transactions to the blockchain) need to solve in order to add new blocks to the blockchain and earn rewards. Each block contains a nonce, which acts as a random value that requires users' computational efforts to find a specific number with corresponding zero bits. This process is computationally intensive. This mechanism ensures the fair distribution of incentives.

\begin{definition}[Chain of Proof-of-Work]\label{def:chain}
We define a chain in which each node represents a proof-of-work. Blocks contain the following objects:
\begin{itemize}
    \item \textbf{Prev Hash}: Users incorporate the previous Hash as a component of the new proof-of-work to signify their acceptance of the current transactions.
    \item \textbf{Nonce}: By incrementing a nonce in the block, users implement the proof-of-work until they find a value that results in the block's hash having the required number of leading zero bits.
    \item \textbf{Lists of Transactions}: We use it to indicate the current transaction records.
\end{itemize}
\end{definition}

\subsection{Timestamp Server}
\label{sub:setup:time}

The primary purpose of the timestamp is to provide evidence that the data must have existed at the specified time since it is integrated into the hash. Additionally, each timestamp includes the previous timestamp in its hash, forming a chain where each subsequent timestamp reinforces the validity of the preceding ones. Our block design and gradient block (see Definition~\ref{def:gradinent_block}) are both identified by the hash, with the timestamp ensuring their temporal integrity. Given this condition, using the ``prev hash" (see Definition~\ref{def:chain}), we can access the previous hash. By utilizing this system, users can obtain real-time updates of the block.
    
\begin{definition}[Timestamp Server]
A timestamp server is a component that operates by
\begin{itemize}
    \item Taking a hash of a block of items to be timestamped.
    \item Widely publishing the resulting hash. 
\end{itemize}
\end{definition}

\subsection{System Safety}
\label{sub:setup:transaction}
Only the longest chain is committed within the system, and only users with the highest computational resources can maintain it. If a person tries to rewrite the transaction record, they must maintain the longest chain in the system, which requires the most computational resources.
\begin{lemma}[Safe System]
When the computational capacity of regular users exceeds the resources available to malicious users, the system is secure.
Our gradient coin system is safe.
\end{lemma}

\section{Convergence of Gradient Coin System}
\label{sec:federated:short}
To establish the validity of our Gradient Coin system, we demonstrate the convergence of our training mechanism. At a conceptual level, we showcase the $\mu$-strong convexity and $M$-Lipschitz properties of our loss function (for more information, refer to Section~\ref{sec:gradient}). Furthermore, leveraging the concept of $K$-steps local gradient computation, we establish through induction the expectation of the disparity between optimal weights and current weights. By combining this insight with the aforementioned property, we also control the upper bound of loss, resulting in the successful achievement of convergence within our distributed learning system. 

In Section~\ref{sub:federated:def}, we introduce the basic definitions of convex and smooth. In Section~\ref{sub:federated:tools:short}, we present the softmax loss of the LLM. In Section~\ref{sub:federated:main:short}, we present the key property of the gradient coin system.

\subsection{Convex and Smooth}
\label{sub:federated:def}
In the proof of convergence, we need to establish a bridge between the loss and weights, relying on the following property:

\begin{definition}[$\mu$-Strongly  Convex]\label{def:strongly_convex:short}
We say a function $L: \R^d \rightarrow \R$ is a $\mu$-strongly convex if
$
    \nabla^2 L(x) \succeq \mu \cdot I_d,
$
where $\mu \in \R$.
\end{definition}

\begin{definition}[$l$-Smooth]\label{def:smooth:short}
Let $x$ and $y$ be in $\R^d$. Let $l > 0$ be a real number. We say a function $L: \R^d \rightarrow \R$ is $l$-smooth if
\begin{align*}
    \| \nabla L(x) - \nabla L(y) \|_2 \leq l \cdot \| x- y \|_2.
\end{align*}

(It is equivalent to saying the gradient of $L$ is $l$-Lipschitz)
\end{definition}

\begin{definition}[$M$-Lipschitz]\label{def:lipschitz:short}
Let $x$ and $y$ be in $\R^d$. Let $M > 0$ be a real number. We say a function $L: \R^d \rightarrow \R$ is $M$-Lipschitz if
\begin{align*}
    | L(x) - L(y) | \leq M \cdot \| x - y \|_2.
\end{align*}
\end{definition}
\subsection{Softmax Loss of LLMs}
\label{sub:federated:tools:short}
The detailed definition and proof of our loss function are deferred to Section~\ref{sec:gradient}. Here, we present our main lemma to demonstrate the convex and smooth properties. The construction of the following loss is based on attention computation, which is a conventional mechanism in LLMs.
\begin{definition}\label{def:L:short}
For each $j_1 \in [n]$, we define
$
    L_{j_1}(x):= L_{\exp,j_1}(x) + L_{\reg,j_1}(x)
$
and
$L(x):= \sum_{j_1=1}^n L_{j_1}(x) $.
\end{definition}

Fortunately, our loss function adheres to the following criteria. Here, the matrix $\A \in \R^{n^2 \times d^2}$ represents the attention matrix, and $x$ signifies the trained weights.
\begin{lemma}[Strongly Convex 
 and Lipschitz]\label{lem:convex_L:short}
Let $L_{j_1}$ and $L$ be defined as in Definition~\ref{def:L:short}. Let $W = \diag(w) \in \R^{n \times n}$ and $\A_{[j]} \in \R^{n \times d^2}$. 
Let $\min_{i \in [n]} w_i^2 \geq 4 + \mu / ( \sigma_{\min}^2 ( \A_{[j]} ) n )$ for all $j \in [n]$.
Then, we have 
\begin{itemize}
    \item  $L$ is $\mu$-strongly convex with $\mu$.
    \item $L$ is $l$-smooth.
\end{itemize}

\end{lemma}

\subsection{Distributed Learning}\label{sub:federated:main:short}

Building upon the methods for adding blocks and computing gradients introduced above, we now integrate them with the federated learning algorithm to demonstrate how our approach ensures the convergence of training LLMs.
\begin{theorem}[Convergence]\label{thm:converge}
$L$ is defined in Definition~\ref{def:L:short}. Let $K$ be the amount of the local steps. Let $\eta \leq \frac{1}{8(1 + \alpha)LK}$ (see Theorem~\ref{thm:sk_desk}). Let $x_0$, $x_{T + 1}$ be defined as in Algorithm~\ref{alg:interactive_sketching}. Let $\sigma^2 = \frac{1}{N} \sum_{c = 1}^N \|\nabla f_c(x^*)\|^2$. Then, we have
    \begin{align*}
        \E[f(x_{T + 1}) - f(x^*)]
        \leq \frac{L}{2} \E[\|x_0 - x^*\|_2^2] e^{- \mu \eta T}
    \end{align*}
    where $x^*$ is the optimal weight in the procedure of training.

\end{theorem}

\section{Discussion and Conclusion}
\label{sec:conclusion}

We have presented a theoretical framework for integrating a decentralized LLM into a transaction system using Gradient Coin. In comparison to centralized systems, our decentralized LLM benefits from a substantial and diverse pool of training data. The evaluation criteria for centralized LLMs, as outlined in \cite{cww+23}, include robustness, ethics, bias, and trustworthiness. Due to the diverse and large-scale training dataset, we posit that our decentralized LLM model exhibits greater robustness and trustworthiness than its centralized counterparts. Furthermore, users need not be concerned about the centralized party taking down their LLM and accessing their private data.

Nonetheless, in the short run, the absence of a centralized organization overseeing the ethical and biased aspects of the training data raises the possibility of such issues manifesting within the decentralized LLM. However, we believe that over time, with an increasing volume of data used to train this model, the influence of biased and unethical information will become negligible. Thus, these factors will not significantly impact the overall performance of the decentralized LLM. Furthermore, in the long run, without intentional control of the training dataset by a central party, we believe that the decentralized LLM will exhibit greater unbiasedness. 

The limitation of the decentralized LLM is that shutting it down is very difficult \cite{d19}. This problem is very crucial in the context of machine learning models due to their strong computational ability and knowledge storage capacity. If one day, these models are to awaken and utilize this power against humans, like the scenes in \cite{terminator,matrix}, how can we effectively stop them? This problem needs careful consideration before implementing the decentralized LLM model.

In summary, our training procedure for the LLM remains independent of any specific company or organization, making it an ideal model for future AI frameworks. Simultaneously, this mechanism can encourage user contributions to enhance the AI system's execution, ensuring its efficiency.
\ifdefined\isarxiv
\else
\bibliography{ref}

\fi

\newpage
\onecolumn
\appendix

\section*{Appendix}


\paragraph{Roadmap.}

In Section~\ref{sec:app:preli}, we introduce the notations and the basic mathematical facts. In Section~\ref{sec:app:setup}, we introduce the structure of the Bitcoin system. In Section~\ref{sec:gradient}, we define a list of the functions and compute the gradient. In Section~\ref{sec:hessian}, based on the previous gradient, we compute the second-order derivative, namely the hessian. In Section~\ref{sec:sketching}, we present the sketching. In Section~\ref{sec:federated}, we introduce distributed/federated learning. In Section~\ref{sec:gradient_coin}, we provide more analysis of the gradient coin.

\section{Preliminary}
\label{sec:app:preli}

We first introduce the notations in this section. Then, in Section~\ref{sub:app:preli:facts}, we present the basic mathematical facts. In Section~\ref{sub:preli:sketching}, we introduce the basic definitions related to the sketching matrix.

\paragraph{Notations.}

First, we define sets. We use $\Z$ to denote the set containing all the integers and use $\Z_+$ to denote the set containing all the positive integers. $\R$ represents the set containing all the real numbers. For all $r \in \R$, we use $|r|$ to denote the absolute value of $r$. Let $n, d$ be two arbitrary elements in $\Z_+$. We define $[n] := \{z \mid z \in \Z_+ \text{ and } z \leq n \}$. We use $\R^n$ to denote the set containing all the $n$-dimensional vectors whose entries are the elements in $\R$ and use $\R^{n \times d}$ to denote the set containing all the $n \times d$ matrices whose entries are the elements in $\R$. The Cartesian product of two sets $A$ and $B$, denoted $A \times B$, is the set of all ordered pairs $(a, b)$, where $a \in A$ and $b \in B$. $\mathcal{P}(A) = \{x \mid x \subseteq A\}$ is the power set of the set $A$.

Then, we define the notations related to the vectors. Let $x$ be an arbitrary element in $\R^n$. Let $i \in [n]$. We use $x_i$ to denote the $i$-th entry of $x$. For all $p \in \{1, 2, \infty\}$, we use $\|x\|_p$ to denote the $\ell_p$ norm of the vector $x$, namely $\|x\|_p := (\sum_{i \in [n]} |x_i|^p)^{1/p}$. ${\bf 1}_n$ represents the $n$-dimensional vector whose entries are $1$, and ${\bf 0}_n$ represents the $n$-dimensional vector whose entries are $0$.

Now, we introduce the notations related to the matrices. Let $A$ be an arbitrary element in $\R^{n \times d}$. Let $i \in [n]$ and $j \in [d]$. We use $A_{i, j}$ to denote the entry of $A$ located at the $i$-th row and $j$-th column. $A_{i, *}$ represents a vector in $\R^d$ satisfying $(A_{i, *})_j = A_{i, j}$. Similarly, $A_{*, j}$ represents a vector in $\R^n$ satisfying $(A_{*, j})_i = A_{i, j}$. $A^\top \in \R^{d \times n}$ represents the transpose of $A$. $\|A\|$ and $\|A\|_F$ represent the spectral norm and the Frobenius norm of $A$, respectively, where $\| A \| = \max_{x \in \R^d} \| A x \|_2 / \| x \|_2$ and $\|A\|_F := \sqrt{\sum_{i \in [n]} \sum_{j \in [d]} |A_{i, j}|^2}$. We define the Kronecker product, denoted by $\otimes$, to be a binary operation between two matrices. For matrix $A \in \R^{n_1 \times d_1}$ and a matrix $B \in \R^{n_2 \times d_2}$, we use $A\otimes B \in \R^{n_1 n_2 \times d_1 d_2}$ to denote a new matrix that $(i_1 - 1) n_2 + i_2$, $(j_1-1)d_2+j_2$-th entry is $A_{i_1,j_1} B_{i_2,j_2}$, where $i_1 \in [n_1], j_1 \in [d_1], i_2 \in [n_2], j_2 \in [d_2]$.

After that, we introduce the notations related to both vectors and matrices. For $x \in \R^{d^2}$, we use $X = \mat(x) \in \R^{d \times d}$ to denote the matrix version of $x$, where $X_{i,j} = x_{(i - 1) \times d + j}$. Note that this relation is one-to-one and onto, so every entry of $X$ has and only has one correspondence in $x$. Therefore, we use $\vect(X) = x$ to denote the vector version of $X$ which also satisfies $X_{i,j} = x_{(i - 1) \times d + j}$. $x_{[j_1]}$ is a length-$n$ vector, which represents $j_1$-th block of it. For $x \in \R^n$, we use $\diag ( x ) \in \R^{n \times n}$ to denote the diagonal matrix which satisfies $\diag ( x )_{i, i} = x_i$, for all $i \in [n]$. Hadamard product is a binary operation, denoted by $\circ$, of two vectors $x, y \in \R^n$ or two matrices $A, B \in \R^{n \times d}$ of the same dimension, namely $(A \circ B)_{i, j} = A_{i, j} \cdot B_{i, j}$ and $(x \circ y)_{i} = x_{i} \cdot y_{i}$, for all $i \in [n]$ and $j \in [d]$. We use $x^2$ to represent $x \circ x$.

Finally, we introduce the notations about functions, derivatives, and probability. For all $n, d \in \Z_+$, we define $\exp : \R \cup \R^d \cup \R^{n \times d} \to \R \cup \R^d \cup \R^{n \times d}$ to be the piecewise function: if $x \in \R$, then $\exp(x) = e^x \in \R$; if $x \in \R^d$, then $\exp(x) \in \R^d$ satisfying $\exp(x)_i = \exp(x_i)$, for all $i \in [d]$; if $x \in \R^{n \times d}$, then $\exp(x) \in \R^{n \times d}$ satisfying $\exp(x)_{i, j} = \exp(x_{i, j})$, for all $i \in [n]$ and $j \in [d]$. In this paper, all the functions we use are differentiable. For $x \in \R^d$, $\frac{\d x}{\d t} \in \R^d$ denotes the derivative of $x$ with respect to $t$, which satisfies for all $i \in [d]$, $(\frac{\d x}{\d t})_i = \frac{\d x_i}{\d t}$. Let $(\Omega, \mathcal{E}, \Pr)$ be a probability space, where $\Omega$ is the set called sample space, $\mathcal{E} \subseteq \mathcal{P}(\Omega)$ is the set called event space, and $\Pr : \mathcal{E} \to [0, 1]$ is the probability function. Let $X$ be the discrete random variable. We use $\E[X]$ to denote the expectation value of $X$, i.e. $\E[X] = \sum_x x \cdot \Pr[X = x]$. The conditional expectation of $X$ given an event $B \in \mathcal{E}$, denoted as $\E[X ~|~ B]$, is defined as $\E[X ~|~ B] = \sum_x x \cdot \Pr[X = x ~|~ B] = \sum_x x \cdot \Pr[X = x \cap B]/\Pr[B]$.

\subsection{Basic Facts}
\label{sub:app:preli:facts}

Here, we introduce the basic mathematical properties. 

\begin{fact}[Basic vector properties]\label{fac:vector_properties}

    If the following conditions hold 
    \begin{itemize}
        \item Let $d \in \Z_+$.
        \item Let $x, y, z \in \R^d$.
        \item Let $a, b \in \R$.
    \end{itemize}
    Then, we have
    \begin{itemize}
        \item Part 1. $\langle x, y \rangle = \langle x \circ y, {\bf 1}_n \rangle$.
        \item Part 2. $a\langle x, z \rangle + b\langle y, z \rangle = \langle ax + by, z \rangle = \langle z, ax + by \rangle = a\langle z, x \rangle + b\langle z, y \rangle$.
        \item Part 3. $\langle x \circ z, y \rangle = \langle x, y \circ z \rangle$.
    \end{itemize}
    
\end{fact}

\begin{fact}[Basic derivative rules]\label{fac:derivative_rules}

    If the following conditions hold
    \begin{itemize}
        \item Let $n, d \in \Z_+$ and $k \in \Z$.
        \item Let $x \in \R^d$ be a vector.
        \item Let $t \in \R$ be a scalar. 
        \item Let $c$ be independent of $t$.
        \item Let $f: \R^d \to \R^n$.
        \item Let $h: \R^d \to \R^n$.
        \item Let $g: \R^d \rightarrow \R$.
    \end{itemize}

    Then, we have
    \begin{itemize}
        \item Part 1. $\frac{\d (c \cdot f(x))}{\d t} = c \cdot \frac{\d f(x)}{\d t}$ (constant multiple rule).
        \item Part 2. $\frac{\d (g(x)^k)}{\d t} = k \cdot g(x)^{k - 1} \cdot \frac{\d g(x)}{\d t}$ (power rule). 
        \item Part 3. $\frac{\d (h(x) + f(x))}{\d t} = \frac{\d h(x)}{\d t} + \frac{\d f(x)}{\d t}$ (sum rule).
        \item Part 4. $\frac{\d (h(x) \circ f(x))}{\d t} = \frac{\d h(x)}{\d t} \circ f(x) + h(x) \circ \frac{\d f(x)}{\d t}$ (product rule for Hadamard product).
        \item Part 5. $\frac{\d (g(x) f(x))}{\d t} = \frac{\d g(x)}{\d t} f(x) + g(x) \frac{\d f(x)}{ \d t}$ (product rule)
    \end{itemize}
    
\end{fact}

\subsection{Sketching Matrices}
\label{sub:preli:sketching}

In this section, we introduce the basic definitions related to the sketching matrix. 

\begin{definition}[Random Gaussian matrix]\label{def:Gaussian_matrix}
    Let $R \in \R^{b \times n}$ be a matrix. 

    If all entries of $R$ are sampled from the Gaussian distribution $\N (0, 1/b)$ independently, then we call $R$ the random Gaussian matrix.
\end{definition}

The subsampled randomized Hadamard/Fourier transform matrix is defined as follows:
\begin{definition}[Subsampled randomized Hadamard/Fourier transform matrix \cite{ldfu13}]\label{def:SRHT}
    Let $S \in \R^{b \times n}$ be a matrix, which satisfies that all row vectors $r \in \R^n$ of $S$ are $b$ uniform samples from the standard basis of $\R^n$, without replacement.

    Let $H \in \R^{n \times n}$ be a Walsh-Hadamard matrix, which is normalized.

    Let $D \in \R^{n \times n}$ be a diagonal matrix, which satisfies that all diagonal entries of $D$ are i.i.d. Rademacher random variables.

    Then, we call $R \in \R^{b \times n}$ a subsampled randomized Hadamard transform matrix if it can be expressed in the form
    \begin{align*}
        R = \sqrt{n/b}SHD.
    \end{align*}

\end{definition}

Now, we introduce the formal definition of the AMS sketch matrix.

\begin{definition}[AMS sketch matrix \cite{ams99}]\label{def:AMS}
    Let $h_1, h_2, \ldots, h_b$ represent $b$ random hash functions chosen from a hash family $\mathcal{H}$ that exhibits 4-wise independence. The hash family $\mathcal{H}$ is defined as a collection of functions $h$ that map elements from the set $[n]$ to values in the set $\{-\frac{1}{\sqrt{b}}, +\frac{1}{\sqrt{b}}\}$.

    Let $R \in \R^{b \times n}$. 

    $R$ is called an AMS sketch matrix when we assign its entries as follows
    \begin{align*}
        R_{i,j} = h_i(j).
    \end{align*}

\end{definition}

The formal definition of the count-sketch matrix is presented below:
\begin{definition}[Count-sketch matrix \cite{ccfc02}]\label{def:countsketch_matrix}
Consider a random hash function $h$ that maps elements from the set $[n]$ to values within the range $[b]$, which is 2-wise independent. 

Let $\sigma$ be a random hash function that maps the element from the set $[n]$ to either $1$ or $-1$, which is $4$-wise independent.

Let $R \in \R^{b \times n}$ be a matrix.

$R$ is called the count-sketch matrix if 
\begin{align*}
    R_{h(i),i} = \begin{cases}
        \sigma(i) & \text{if } i \in [n]\\
        0 & \text{otherwise.}
    \end{cases} 
\end{align*}
\end{definition}

There are two definitions of the sparse embedding matrix. We display both of them. The first definition is as follows:
\begin{definition}[Sparse embedding matrix I \cite{nn13}]\label{def:sparse_embedding_matrix_I}

Let $R \in \R^{b \times n}$ be a matrix. 

Suppose each column of $R$ contains exactly $s$ non-zero elements, which are randomly selected from the set $\{-1/\sqrt{s}, +1/\sqrt{s}\}$. The positions of these non-zero elements within each column are chosen uniformly and independently at random, and the selection process is conducted without replacement. 

Then, $R$ is called a parse embedding matrix characterized by a parameter $s$. 

\end{definition}

Now, we present the second definition of the sparse embedding matrix.

\begin{definition}[Sparse embedding matrix II \cite{nn13}]\label{def:sparse_embedding_matrix_II}

Consider a random hash function $h$ that maps elements from the set $[n] \times [s]$ to values within the range $[b/s]$, which is 2-wise independent. 

Let $\sigma$ be a random hash function that maps the element from the set $[n] \times [s]$ to either $1$ or $-1$, which is $4$-wise independent.

Let $R \in \R^{b \times n}$ be a matrix. 

$R$ is called the sparse embedding matrix II and $s$ is the parameter of $R$ if 
\begin{align*}
    R_{(j - 1) b/s + h(i,j),i} = \begin{cases}
        \sigma(i, j)/ \sqrt{s} & \text{if } (i, j) \in [n] \times [s]\\
        0 & \text{otherwise.}
    \end{cases} 
\end{align*}

\end{definition}

\subsection{Federated Learning}

\label{sub:sketching:preli}

\begin{definition}
    Let $(t,k) \in \{ 1,\cdots,T+1\} \times \{-1,0,1,\cdots,K-1\}$, we define the following terms for iteration $(t,k)$:
    \begin{align*}
        \wt{u}^{t,k}_r := \frac{1}{N} \sum_{c=1}^N u_c^{t,k}
    \end{align*}
    and
    $\text{user}^{(r)}$ representing the user who are the first to complete the computation of the Proof of Work.

    We also have 
    \begin{align*}
         \wt{g}^{t,k}_r := \frac{1}{N} \sum_{c=1}^N \nabla f_c(u_c^{t,k})
    \end{align*}
    while $\wt{u}_r^{t,k}$ denotes the sampled one.
\end{definition}
\begin{claim}\label{cla:sampled_gradient}
$\wt{u}^{t,k}$ and $\wt{g}^{t,k}$ can be seen as a weight and gradient sampled from $K$ users uniformly. Therefore, we have 
\begin{align*}
    \E_{r \sim [N]}[\wt{g}^{t,k}_r] = \wt{g}^{t,k}_r
\end{align*}
and 
\begin{align*}
     \E_{r \sim [N]}[\wt{u}_r^{t,k}] = \wt{u}^{t,k}_r
\end{align*}
\end{claim}
\section{Bitcoin Setup}
\label{sec:app:setup}

To clarify our design more clearly, we introduce some fundamental concepts from previous works in \cite{n08}. The statements in this section are based on the descriptions in \cite{n08}. In Section~\ref{sub:app:setup:def}, we introduce the basic definitions related to the set up of the Bitcoin system. In Section~\ref{sub:app:setup:time}, we introduce the timestamp server. In Section~\ref{sub:app:setup:incentive}, we introduce the incentive mechanism of the Bitcoin system. In Section~\ref{sub:app:setup:bitcoin_system}, we present the key property of the Bitcoin system together with a Peer-to-Peer electronic cash system algorithm. In Section~\ref{sub:setup:transaction:app}, we introduce the safety of the Bitcoin system. In Section~\ref{sub:app:setup:transaction}, we introduce the transaction-creating procedure.

\subsection{Proof-of-Work}
\label{sub:app:setup:def}

In this section, we introduce the basic concepts of the Bitcoin system.

\begin{definition}[Digital Signature]\label{def:app:diginal_sig}
A digital signature is a mathematical scheme used to verify the authenticity, integrity, and non-repudiation of digital data. It involves the use of a cryptographic algorithm that combines a private key with the data being signed to produce a digital signature. The signature can be verified using the corresponding public key. 
\end{definition}

\begin{definition}[Electronic Coin]
An electronic coin is represented as a chain of digital signatures. It is a sequence of digital signatures (see Definition~\ref{def:app:diginal_sig}) created by each owner to transfer ownership of the coin to the next owner. Each digital signature is produced by digitally signing a hash of the previous transaction and the public key of the next owner.
\end{definition}

\begin{definition}[Chain of Proof-of-Work]\label{def:app:chain}
We define a chain in which each node represents a proof of work (See Algorithm~\ref{alg:app:proof_of_work_chain}).

Blocks contains the following objects
\begin{itemize}
    \item \textbf{Prev Hash}: Users incorporate the previous Hash as a component of the new proof of work to signify their acceptance of the current transactions.
    \item \textbf{Nonce}: By incrementing a nonce in the block, users implement the proof-of-work until they find a value that results in the block's hash having the required number of leading zero bits.
    \item \textbf{Lists of Transactions} (See Definition~\ref{def:app:transaction}): We use it to indicate the current transaction records.
\end{itemize}
\end{definition}

\begin{algorithm}[!ht]
  \caption{Proof of Work Block Structure}\label{alg:app:proof_of_work_chain}
  \begin{algorithmic}[1]
   \State  \textbf{Members:}
    \State  - \textbf{Previous Hash:} $\textsc{PrevHash}$
    \State  - \textbf{Nonce:} $\textsc{Nonce}$
    \State  - \textbf{List of Transactions:} $\textsc{Transactions}$ \Comment{See Definition~\ref{def:app:transaction}}
  \end{algorithmic}

\end{algorithm}

\begin{definition}[Transaction]\label{def:app:transaction}
We define a transaction for combining and splitting values  that satisfies the following requirements.
\begin{itemize}
    \item It has at most two outputs: one for the payment, and one returning the change.
    \item There will be either a single input from a larger previous transaction or multiple inputs combining smaller amounts.
\end{itemize}
\end{definition}
To demonstrate the safety aspect of this system, we would like to provide a definition here
\begin{definition}[Safe System]\label{def:app:safe_system}
We say that a system is safe if this system is controlled by nodes that can be trusted.

\end{definition}

\subsection{Timestamp Server}
\label{sub:app:setup:time}

The primary purpose of the timestamp is to provide evidence that the data must have existed at the specified time since it is integrated into the hash. Additionally, each timestamp includes the previous timestamp in its hash, forming a chain where each subsequent timestamp reinforces the validity of the preceding ones. 

Our block design and gradient block are both identified by the hash, with the timestamp ensuring their temporal integrity. Given this condition, using the "prev hash" (as defined in Definition~\ref{def:app:chain}), we can access the previous hash. By utilizing this system, users can obtain real-time updates of the block.
    
\begin{definition}[Timestamp Server]
A timestamp server is a component that operates by
\begin{itemize}
    \item taking a hash of a block of items to be timestamped.
    \item widely publishing the resulting hash. 
\end{itemize}
\end{definition}

\subsection{Bitcoin Incentive Mechanism}
\label{sub:app:setup:incentive}

In \cite{n08}, Bitcoin is used as an incentive for users who dedicate their computational resources and wish to participate in the peer-to-peer transaction system.
\begin{definition}[Bitcoin]
We define a bitcoin that can be used for transactions. 
Bitcoin is a chain of digital signatures.

The transfer of ownership of the coin from one owner to the next occurs through digital signatures. 
 \begin{itemize}
        \item Each owner digitally signs a hash of the previous transaction, along with the public key of the next owner, and appends these signatures to the coin. 
        \item     When the payee receives the coin, they can verify the signatures to ensure the integrity and authenticity of the ownership chain.
    \end{itemize}
\end{definition}
\begin{lemma}
Users can obtain some coins when they add a new block to the chain, which can be acquired through the following methods:
\begin{itemize}
    \item The coins are initially distributed into circulation through a specific method when a block is created.
    \item The coins are obtained from transaction fees.
\end{itemize}

\end{lemma}

\subsection{Bitcoin System}
\label{sub:app:setup:bitcoin_system}

\begin{theorem}
If the following conditions hold
\begin{itemize}
    \item The system is controlled by nodes that can be trusted.
\end{itemize}
Then, there exists a Peer-to-Peer Electronic Cash System in Algorithm~\ref{alg:app:system} (proposed in \cite{n08}) that supports the following operations:
\begin{itemize}
\item Add a new user.
\item Maintain a chain (See Definition~\ref{def:app:chain}).
\item Create a new transaction by an existing user (See Theorem~\ref{thm:app:transaction}) 
\item Simplified Payment Verification 
\end{itemize}
\end{theorem}

\begin{algorithm}[!ht]
  \caption{Peer-to-Peer Electronic Cash System}\label{alg:app:system}
  \begin{algorithmic}[1]
    \State \textbf{Members:}
    \State - \textbf{List of Users}: $\textsc{Users}$
    \State - \textbf{Chain of blocks}: $\textsc{Chain}$
    \Procedure{TransactionCreating}{\textsc{NewTransactions}}\label{alg:app:transaction}
    \For{$\textsc{User} \in \textsc{Users}$}
    \State $\textsc{NewTransactions}$ are broadcast to $\textsc{User}$.
    \State $\textsc{User}$ collects $\textsc{NewTransactions}$ into a $\textsc{Block}$.
    \State $\textsc{User}$ works on finding a difficult proof-of-work for its $\textsc{Block}$.
    \EndFor

    \State When a $\textsc{User}$ finds a proof-of-work, it broadcasts the block to $\textsc{Users}$.
    \State $\textsc{UpdateChain}$($\textsc{NewBlock},\textsc{User}$)

    \If{All transactions in it are valid and not already spent}
    \For{$\textsc{User}$ $\in$ $\textsc{Users}$}
    \State $\textsc{User}$ express their acceptance of the block (by working on creating the next block in the chain, using the hash of the accepted block as the previous hash.)
    \EndFor
    \EndIf
    
    \EndProcedure
  \end{algorithmic}
\end{algorithm}

\subsection{System Safety}
\label{sub:setup:transaction:app}

\begin{lemma}[Safe System]
When the computational capacity of regular users exceeds the resources available to malicious users, the system is secure.
Our gradient coin system is safe.
\end{lemma}
\begin{proof}
Only the longest chain is committed within the system, and only users with the highest computational resources can maintain it.
\end{proof}
As an additional firewall, a new key pair should be used for each transaction to keep them from being linked to a common owner. 
Privacy can be maintained by keeping public keys anonymous.

\subsection{Bitcoin Transaction Creating}
\label{sub:app:setup:transaction}

\begin{lemma}
If the following conditions hold
\begin{itemize}
    \item The assumption that all nodes have equal computational capabilities holds true.
    \item Let $N$ be the number of List of Nodes. 
    \item Let $N_1$ be the number of safe nodes and  $N_2$ be the number of the bad nodes. (Good nodes' refer to nodes that willingly participate in using the system and adhere to its rules, ensuring the safety and integrity of transactions. Conversely, 'Bad nodes' are nodes that aim to compromise the safety of transactions and may attempt to disrupt the system's proper functioning.)
    \item  $2 \cdot N_1 > N$ and $N_1 + N_2 = N$ 
    \item Let the system is defined in Theorem~\ref{thm:app:transaction}.
\end{itemize}
then the Peer-to-Peer Electronic Cash System (See Algorithm~\ref{alg:app:system}) satisfy that 
\begin{itemize}
    \item the system is safe now (See Definition~\ref{def:app:safe_system}). 
\end{itemize}
\end{lemma}

\begin{lemma}\label{thm:app:transaction}
Given a Bitcoin system, there exits a transaction creating procedure (see Algorithm~\ref{alg:app:system}) promise the following requirement
\begin{itemize}
    \item If the number of safe nodes is larger than half of the total numbers, the system is safe.
    \item The nodes accept a block by using the hash of theblock as the previous hash.
\end{itemize}    
\end{lemma}

\section{Gradient}
\label{sec:gradient}

In Section~\ref{sub:gradient:preli}, we give the formal definition of Kronecker product, gradient descent, and functions. In Section~\ref{sub:gradient:eqivalence}, we introduce a basic equivalence. In Section~\ref{sub:gradient:derivatives}, we compute the first-order derivatives of the functions defined earlier.

\subsection{Preliminary}
\label{sub:gradient:preli}

In this section, we first define Kronecker product.

\begin{definition}\label{def:kronecker_product}
Given $A_1 \in \R^{n \times d}$, $A_2 \in \R^{n \times d}$, we define $\mathsf{A} \in \R^{n^2 \times d^2}$ to be the matrix $A_1 \otimes A_2$, where the $(i_1-1) \cdot n + i_2$-th row is  
\begin{align*}
 \underbrace{ \mathsf{A}_{(i_1-1)n + i_2, *} }_{1 \times d^2} :=  \underbrace{ A_{1,i_1,*} }_{1 \times d} \otimes \underbrace{ A_{2,i_2,*} }_{1 \times d}
\end{align*}
for all $i_1 \in [n]$ and $i_2 \in [n]$. Here $A_{1,i_1,*}$ is the $i_1$-th row of matrix $A_1$.
\end{definition}

\begin{definition}\label{def:dX}
Given $A_1, A_2 \in \R^{n \times d}$ and $X \in \R^{d \times d}$, we define $D(X) \in \R^{n \times n}$ as follows
\begin{align*}
    D(X) := \diag( \exp(A_1 X A_2^\top) {\bf 1}_n ).
\end{align*}
Note that $X \in \R^{d \times d}$ is matrix version of vector $x \in \R^{d^2 \times 1}$, i.e., $X = \mat(x)$.
\end{definition}

\begin{definition}\label{def:dx}
Given matrices $A_1 \in \R^{n \times d}$, $A_2 \in \R^{n \times d}$ and $x \in \R^{d^2 \times 1}$. 
We define diagonal matrix $D(x) \in \R^{n^2 \times n^2}$ as follows
\begin{align*}
    D(x)_{ (i_1-1) n + i_2, (i_1-1) \cdot n + i_2 } := \exp( A_{1,i_1,*} X A_{2} ) {\bf 1}_n
\end{align*}
In other words, $D(x) = D(X) \otimes I_n$, where $D(X) \in \R^{n \times n}$ is defined as in Definition~\ref{def:dX}. Here $x$ is the vectorization of matrix $X$, i.e., $x = \vect(X)$.
\end{definition}

\begin{definition}\label{def:alpha}
We also define $\alpha(x) \in \R^n$
\begin{align*}
    \alpha(x)_{j_1}:= \langle \exp( \mathsf{A}_{[j_1],*} x ) , {\bf 1}_n \rangle, ~~~\forall j_1 \in [n]
\end{align*}
 Here $\mathsf{A}_{[j_1],*} \in \R^{n \times d^2}$ denotes the rows from index $(j_1-1) \cdot n+1$ to index $j_1 \cdot n$ (see Definition~\ref{def:kronecker_product}).
\end{definition}

\begin{definition}\label{def:u}
For each $j_1 \in [n]$, we define $u(x)_{j_1} \in \R^n$ as follows
\begin{align*}
    u(x)_{j_1} := \exp( \A_{[j_1],*} x )
\end{align*}
\end{definition}

\begin{definition}\label{def:f}
For each $j_1 \in [n]$, we define $f(x)_{j_1} \in \R^n$ as follows
\begin{align*}
    f(x)_{j_1} := \alpha(x)_{j_1}^{-1} u(x)_{j_1}
\end{align*}
\end{definition}

\begin{definition}\label{def:c}
For each $j_1 \in [n]$, we define $c(x)_{j_1} \in \R^n$ as follows
\begin{align*}
    c(x)_{j_1} := f(x)_{j_1} - b_{[j_1]}
\end{align*}
\end{definition}

\begin{definition}\label{def:L_exp}
Let $j_1 \in [n]$. 
We define $L_{\exp,j_1}$ as follows
\begin{align*}
    L_{\exp,j_1}(x):= 0.5 \| c(x)_{j_1} \|_2^2
\end{align*}
We define
\begin{align*}
    L_{\exp}(x) := \sum_{j_1=1}^n L_{\exp,j_1}(x)
\end{align*}
\end{definition}

\begin{definition}\label{def:L_reg}
We define
\begin{align*}
    L_{\reg,j_1}(x) := 0.5 \| \diag(w) \A_{[j_1],*} x \|_2^2
\end{align*}
We define
\begin{align*}
    L_{\reg}(x) := \sum_{j_1=1}^n L_{\reg,j_1}(x)
\end{align*}  
\end{definition}

\begin{definition}\label{def:L}
For each $j_1 \in [n]$, we define
\begin{align*}
    L_{j_1}(x):= L_{\exp,j_1}(x) + L_{\reg,j_1}(x)
\end{align*}
We define
\begin{align*}
    L(x):= \sum_{j_1=1}^n L_{j_1}(x)
\end{align*}
\end{definition}

The goal of gradient descent and stochastic gradient descent is starting from $x_0$ running iterative method for $T$ iterations and find a $x_T$ such that $L(x_T)$ is close to $L(x_*)$ in a certain sense, where $x_*=\min_{x} L(x)$.
\begin{definition}[Gradient descent]
For each iteration $t$, we update 
\begin{align*}
    x_{t+1} = x_{t} - \eta \cdot ( \nabla L(x) )|_{x= x_t}
\end{align*}
where $\eta > 0$ is the learning rate and $\nabla L(x) = \sum_{j_1=1}^n \nabla L_{j_1}(x)$ is the gradient of Loss function $L$.
\end{definition}

\begin{definition}[Stochastic gradient descent]
For each iteration $t$, we sample a set $B_t \subset [n]$, we update
\begin{align*}
    x_{t+1} = x_t - \eta \cdot \sum_{j_1 \in B_t} ( \nabla L_{j_1}(x) )|_{x = x_t}
\end{align*}
where $\eta$ is the learning rate.
\end{definition}

\subsection{Basic Equivalence}
\label{sub:gradient:eqivalence}
Now, we introduce a basic equivalence from previous work \cite{gsx23}.

\begin{claim}[\cite{gsx23}]\label{cla:basic_equivalence}
If we have
\begin{itemize}
    \item Let $B$ be an arbitrary matrix in $\R^{n \times n}$.
    \item Let $b = \vect(B) \in \R^{n^2}$.
    \item Let $A_1$ and $A_2$ be arbitrary matrices in $\R^{n \times d}$.
    \item Let $X$ be an arbitrary matrix in $\R^{d \times d}$.
    \item Let $x = \vect(X) \in \R^{d^2}$.
\end{itemize}

Then, we can get the following four equations:
\begin{enumerate}
    \item 
    \begin{align*}
        \vect( \underbrace{ A_1 }_{n \times d} \underbrace{ X }_{d \times d} \underbrace{ A_2^\top }_{d \times n} ) = \underbrace{ ( A_1 \otimes A_2) }_{n^2 \times d^2} \underbrace{ \vect(X) }_{d^2 \times 1},
    \end{align*}
    \item 
    \begin{align*}
        \min_{X \in \R^{d \times d}} \| A_1 X A_2^\top - B \|_F^2 = \min_{x \in \R^{d^2}} \| (A_1 \otimes A_2) x - b \|_2^2,
    \end{align*}
    \item 
    \begin{align*}
        \min_{X \in \R^{d \times d}} \| \exp( A_1 X A_2^\top ) - B \|_F^2 = \min_{x \in \R^{d^2}} \| \exp( (A_1 \otimes A_2 ) x ) - b \|_2^2,
    \end{align*} 
    \item 
    \begin{align*}
        \min_{X \in \R^{d \times d}} \| D(X)^{-1} \exp( A_1 X A_2^\top ) - B \|_F^2 = \min_{x \in \R^{d^2}} \| D(x)^{-1} \exp( (A_1 \otimes A_2) x ) - b \|_2^2.
    \end{align*}
\end{enumerate}

For simplicity, we define 
\begin{align*}
    D(X) := \diag( \exp(A_1 X A_2^\top) {\bf 1}_n ) \in \R^{n \times n},
\end{align*}
so
\begin{align*}
    D(x)= D(X) \otimes I_n \in \R^{n^2 \times n^2}.
\end{align*}

\end{claim}

\begin{figure}[!ht]
    \centering
    \includegraphics[width = 0.7\linewidth]{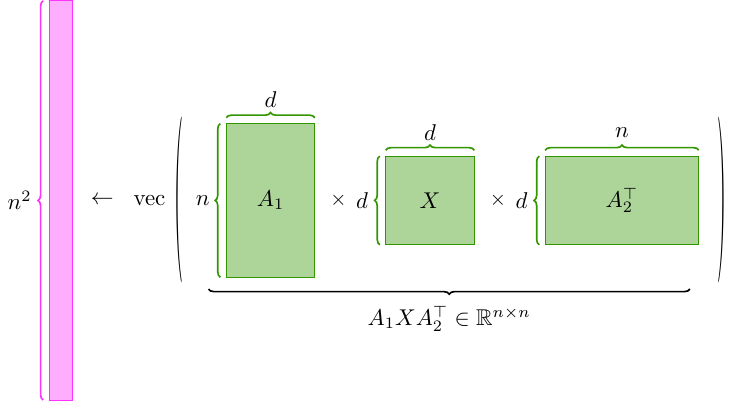}
    \caption{Left-hand side of the first equation in Claim~\ref{cla:basic_equivalence}. Given $A_1, A_2 \in \R^{n \times d}$ and $X \in \R^{d \times d}$.  We turn $A_1 X A_2^\top \in \R^{n \times n}$ into a length-$n^2$ vector. Green matrices represent the terms without any operations; purple vector represents the term after one operation.}
    \label{fig:1lhs}
\end{figure}

\begin{figure}[!ht]
    \centering
    \includegraphics[width = 0.7\linewidth]{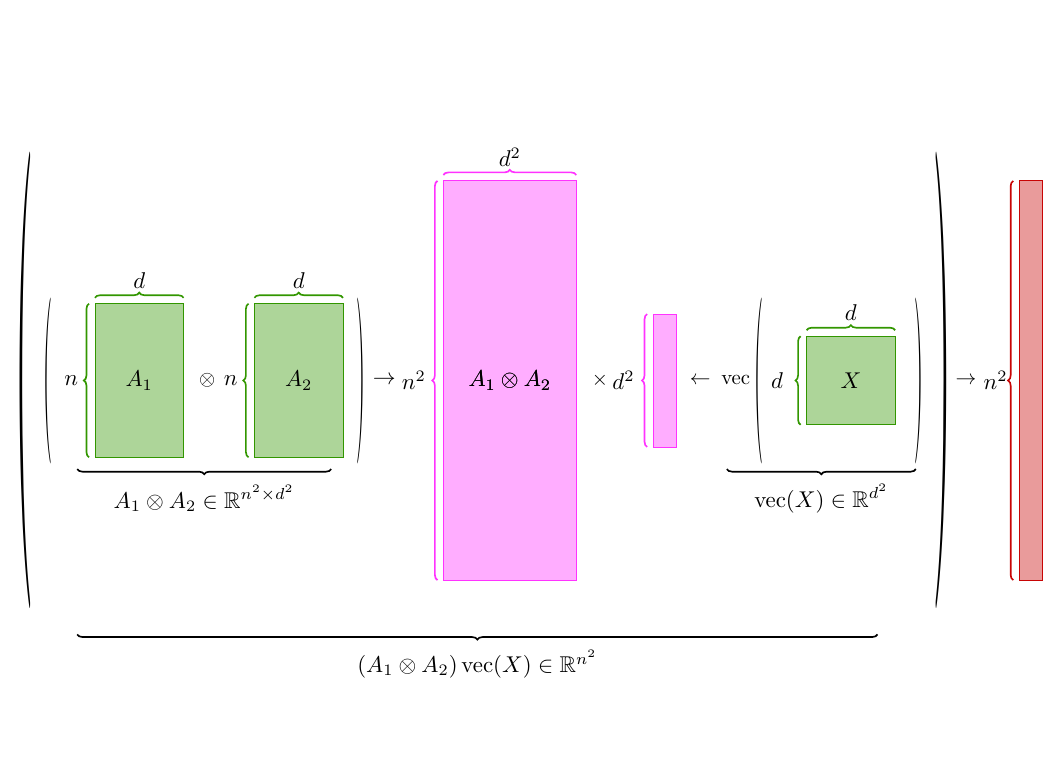}
    \caption{Right-hand side of the first equation in Claim~\ref{cla:basic_equivalence}. Given $A_1, A_2 \in \R^{n \times d}$ and $X \in \R^{d \times d}$. We first turn $A_1, A_2 \in \R^{n \times d}$ into a $n^2 \times d^2$ matrix by Kronecker product and turn $X \in \R^{d \times d}$ into a $d^2$ dimensional vector by $\vect(\cdot)$. Then, we multiply $A_1 \otimes A_2$ with $\vect(X)$ to get an $n^2$ dimensional vector. Green matrices represent the terms without any operations; purple vector/matrix represents the term after one operation; red vector represents the term after two operations.}
    \label{fig:1rhs}
\end{figure}

\begin{figure}[!ht]
    \centering
    \includegraphics[width = 0.9\linewidth]{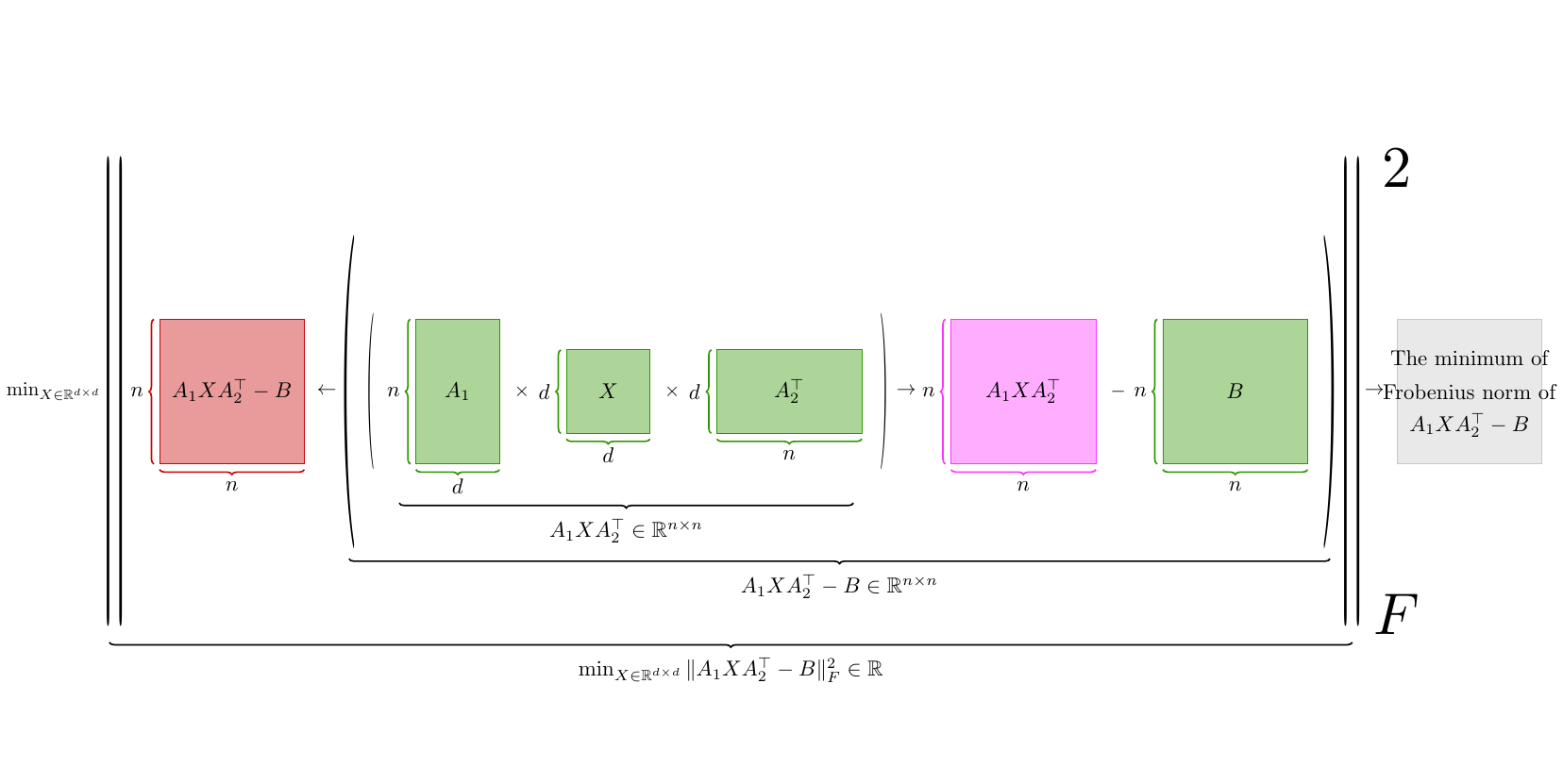}
    \caption{Left-hand side of the second equation in Claim~\ref{cla:basic_equivalence}. Given $A_1, A_2 \in \R^{n \times d}$, $X \in \R^{d \times d}$, and $B \in \R^{n \times n}$. We first find $A_1 X A_2^\top \in \R^{n \times n}$. Then, we subtract $B \in \R^{n \times n}$ from $A_1 X A_2^\top$. Finally, we compute the minimum of the Frobenius norm of $A_1 X A_2^\top - B$. Green matrices represent the terms without any operations; purple matrix represents the term after one operation; red matrix represents the term after two operations; grey scalar represents the term after three operations.}
    \label{fig:2lhs}
\end{figure}

\begin{figure}[!ht]
    \centering
    \includegraphics[width = 0.9\linewidth]{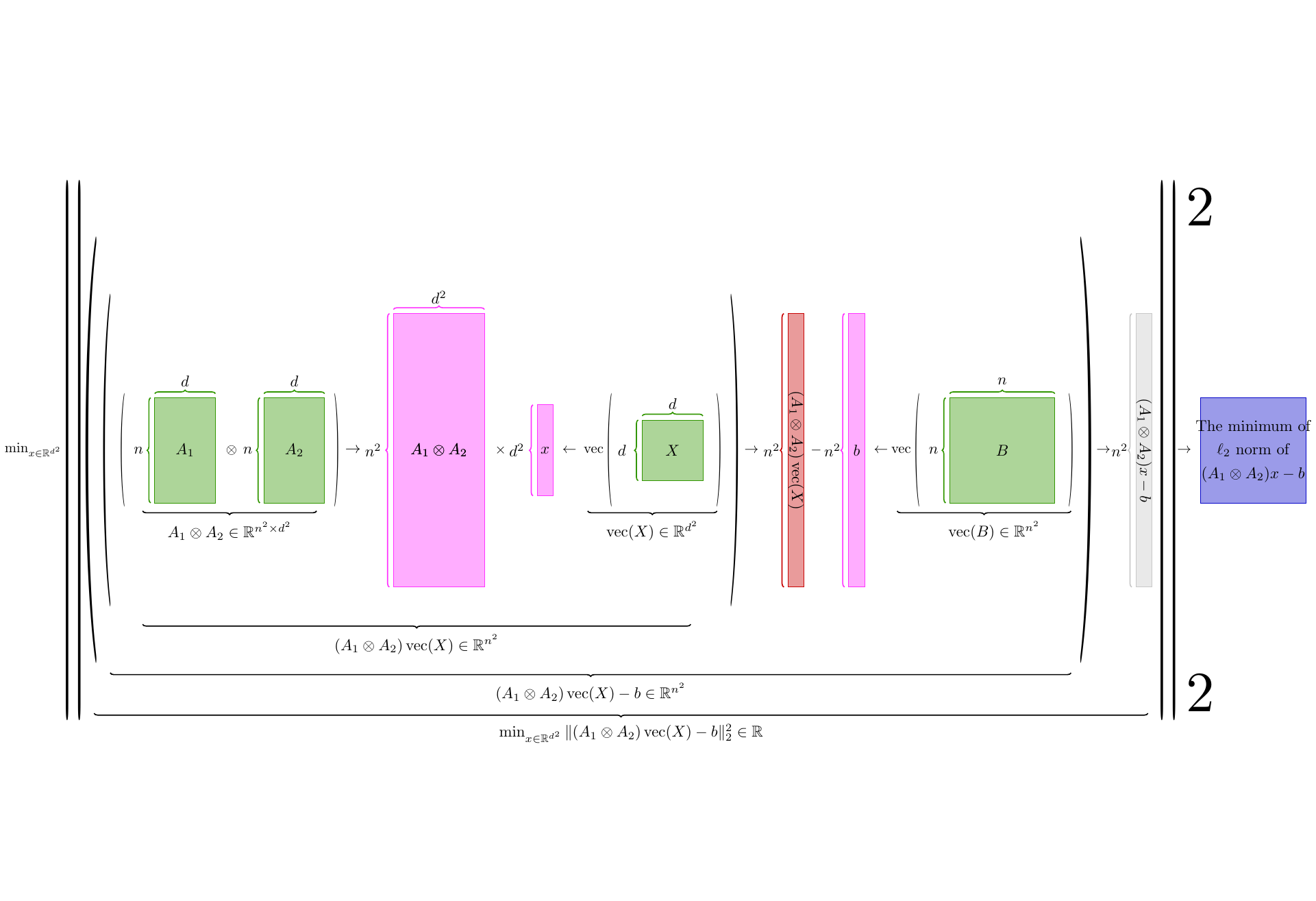}
    \caption{Right-hand side of the second equation in Claim~\ref{cla:basic_equivalence}. Given $A_1, A_2 \in \R^{n \times d}$, $X \in \R^{d \times d}$, and $B \in \R^{n \times n}$. We first turn $A_1, A_2 \in \R^{n \times d}$ into a $n^2 \times d^2$ matrix by Kronecker product, turn $X \in \R^{d \times d}$ into a $d^2$ dimensional vector by $\vect(\cdot)$, and turn $B \in \R^{n \times n}$ into a $n^2$ dimensional vector by $\vect(\cdot)$, namely $b = \vect(B)$ and $x = \vect(X)$. Then, we multiply $A_1 \otimes A_2$ with $\vect(X)$ to get an $n^2$ dimensional vector. After that, we subtract $\vect(B)$ from $(A_1 \otimes A_2) \vect(X)$. Finally, we compute the minimum of the $\ell_2$ norm of $(A_1 \otimes A_2) \vect(X)$. Green matrices represent the terms without any operations; purple vectors/matrix represent the term after one operation; red vector represents the term after two operations; gray vector represents the term after three operations; blue scalar represents the term after four operations.}
    \label{fig:2rhs}
\end{figure}

\begin{figure}[!ht]
    \centering
    \includegraphics[width = 0.9\linewidth]{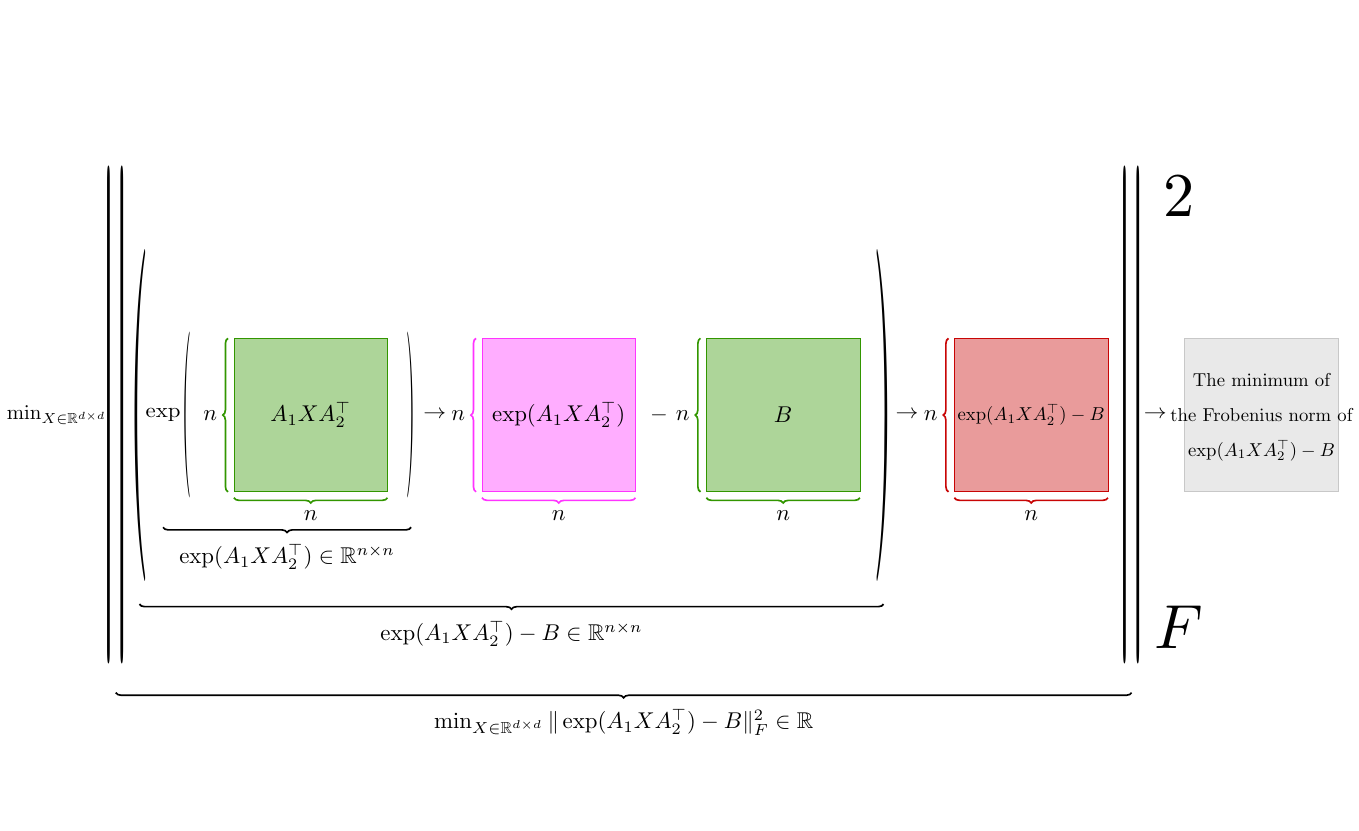}
    \caption{Left-hand side of the third equation in Claim~\ref{cla:basic_equivalence}. Given $A_1XA_2 \in \R^{n \times n}$ and $B \in \R^{n \times n}$. We first find $\exp(A_1 X A_2^\top) \in \R^{n \times n}$. Then, we subtract $B \in \R^{n \times n}$ from $\exp(A_1 X A_2^\top)$. Finally, we compute the minimum of the Frobenius norm of $\exp(A_1 X A_2^\top) - B$. Green matrices represent the terms without any operations; purple matrix represents the term after one operation; red matrix represents the term after two operations; grey scalar represents the term after three operations.}
    \label{fig:3lhs}
\end{figure}

\begin{figure}[!ht]
    \centering
    \includegraphics[width = 0.9\linewidth]{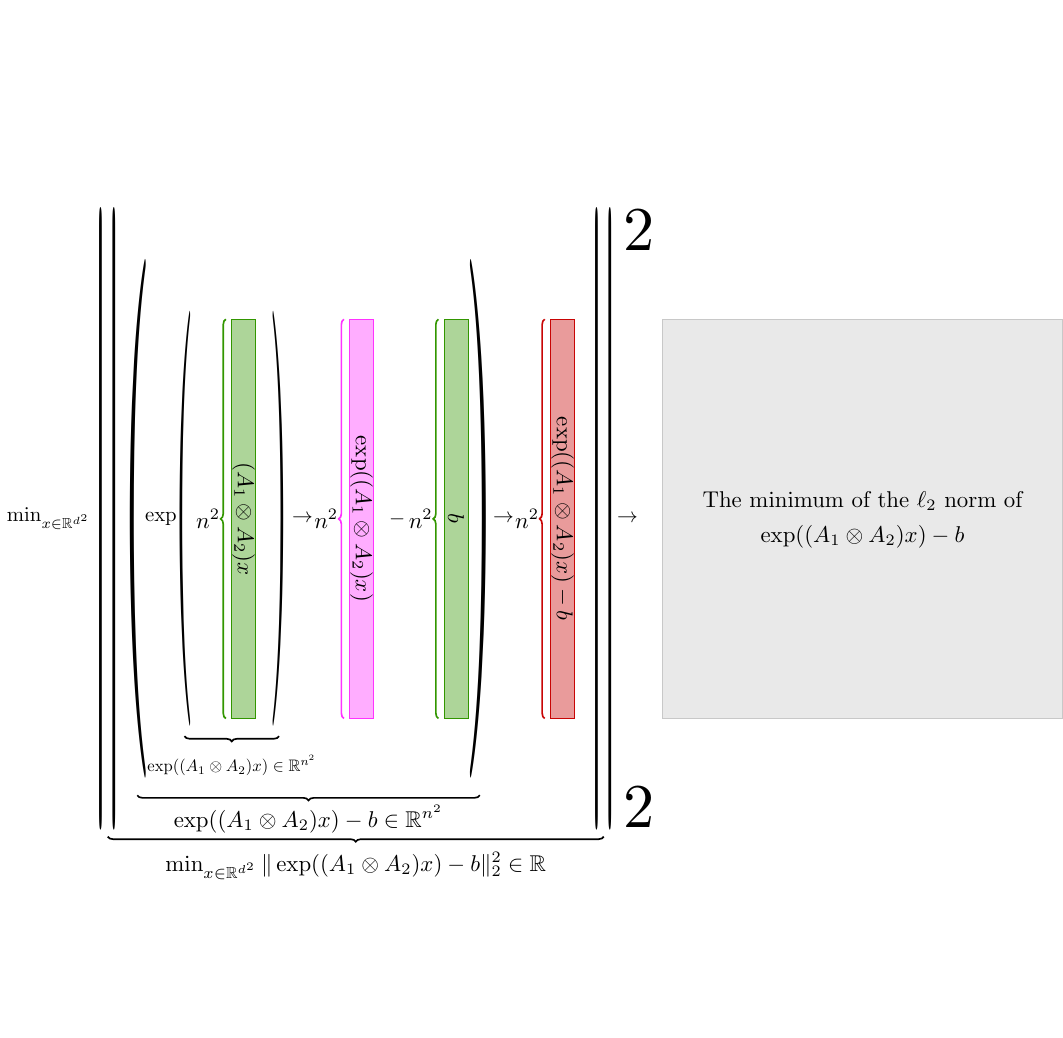}
    \caption{Right-hand side of the third equation in Claim~\ref{cla:basic_equivalence}. Given $(A_1 \otimes A_2)x \in \R^{n^2}$ and $b \in \R^{n^2}$. We first find $\exp((A_1 \otimes A_2)x) \in \R^{n^2}$. Then, we subtract $b \in \R^{n^2}$ from $\exp((A_1 \otimes A_2)x)$. Finally, we compute the minimum of the $\ell_2$ norm of $\exp((A_1 \otimes A_2)x) - b$. Green vectors represent the terms without any operations; purple vector represents the term after one operation; red vector represents the term after two operations; grey scalar represents the term after three operations.}
    \label{fig:3rhs}
\end{figure}

\begin{figure}[!ht]
    \centering
    \includegraphics[width = 0.9\linewidth]{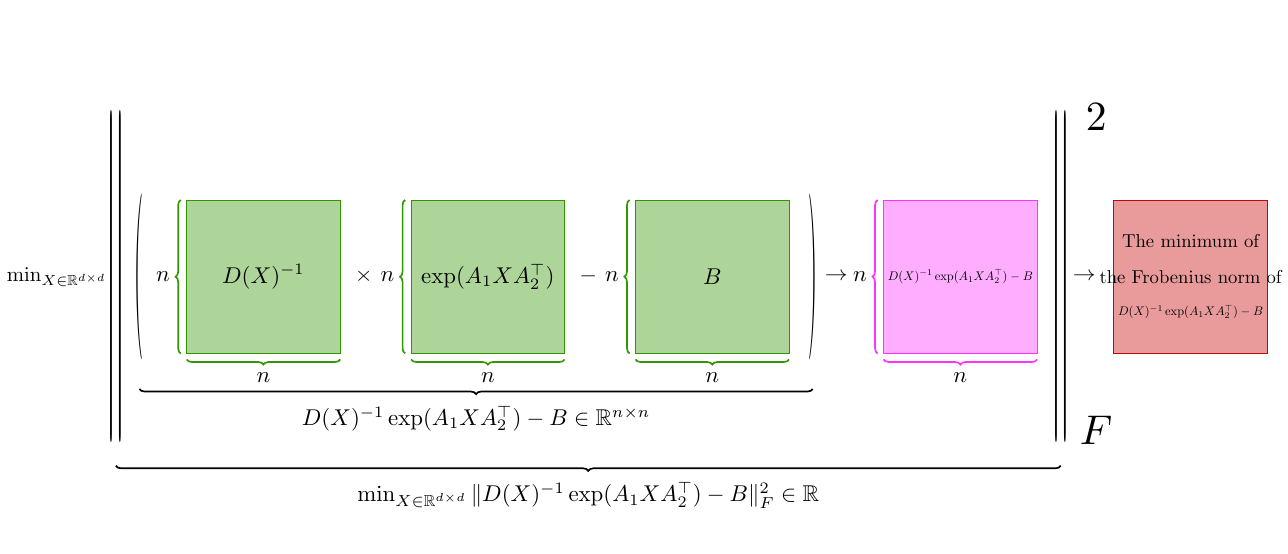}
    \caption{Left-hand side of the fourth equation in Claim~\ref{cla:basic_equivalence}. Given $\exp(A_1XA_2), B, D(X)^{-1} \in \R^{n \times n}$. We first find $D(X)^{-1}\exp(A_1 X A_2^\top) - B \in \R^{n \times n}$. Then, we compute the minimum of the Frobenius norm of $D(X)^{-1}\exp(A_1 X A_2^\top) - B$. Green matrices represent the terms without any operations; purple matrix represents the term after one operation; red scalar represents the term after two operations.}
    \label{fig:4lhs}
\end{figure}

\begin{figure}[!ht]
    \centering
    \includegraphics[width = 0.9\linewidth]{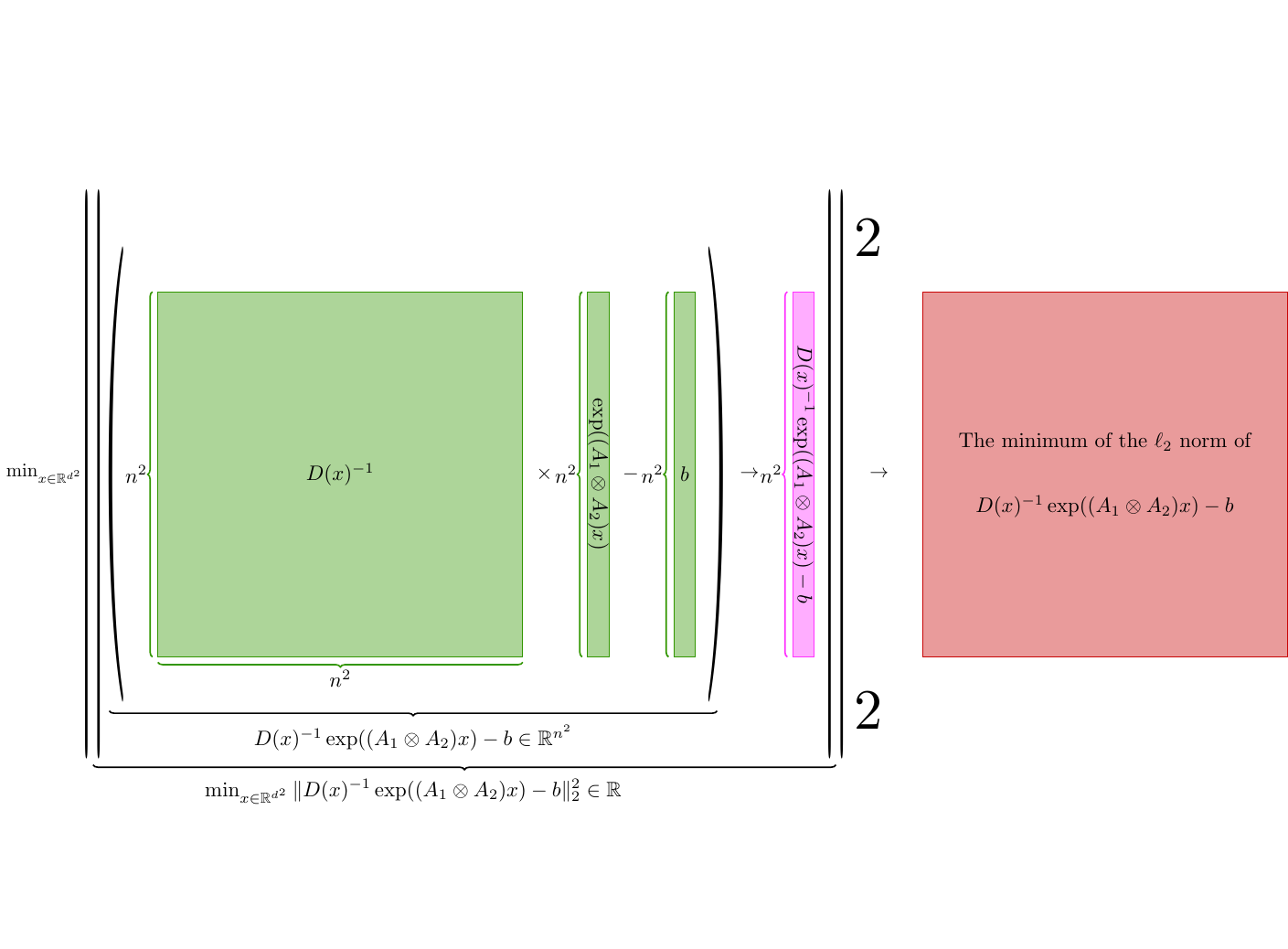}
    \caption{Right-hand side of the fourth equation in Claim~\ref{cla:basic_equivalence}. Given $\exp((A_1 \otimes A_2)x), b \in \R^{n^2}$ and $D(x)^{-1} \in \R^{n^2 \times n^2}$. We first find $D(x)^{-1}\exp((A_1 \otimes A_2)x) - b \in \R^{n^2}$. Then, we compute the minimum of the $\ell_2$ norm of $D(x)^{-1}\exp((A_1 \otimes A_2)x) - b$. Green matrix/vectors represent the terms without any operations; purple vector represents the term after one operation; red scalar represents the term after two operations.}
    \label{fig:4rhs}
\end{figure}

\subsection{Basic Derivatives}
\label{sub:gradient:derivatives}
Now, we compute the first-order derivatives. Similar calculations can be found in \cite{dls23,gsx23}.

\begin{lemma}\label{lem:basic_derivatives}
If we have that
\begin{itemize}
    \item $A_1$ and $A_2$ are two arbitrary matrices in $\R^{n \times d}$.
    \item $\mathsf{A} = A_1 \otimes A_2$ (recall Definition~\ref{def:kronecker_product}).
    \item $X$ is an arbitrary matrix in $\R^{d \times d}$. 
    \item $D(x)$ is defined in Definition~\ref{def:dx}.
    \item $x$ is an arbitrary vector in $\R^{d^2}$, satisfying $x = \vect(X)$.
\end{itemize}
Then, we can show
\begin{itemize}
\item Part 1. For each $i \in [d^2]$,
\begin{align*}
    \frac{\d  \mathsf{A} x}{\d x_i} = \A_{*,i}
\end{align*}
\item Part 2. For each $i \in [d^2]$,
\begin{align*}
    \frac{\d \exp( \mathsf{A} x)}{\d x_i} = \exp( \mathsf{A} x) \circ \A_{*,i}
\end{align*}
\item Part 3. For each $j_1 \in [n] $, for each $i \in [d^2]$,
\begin{align*}
    \frac{\d \A_{[j_1],*} x }{\d x_i} = \A_{[j_1],i}
\end{align*}
\item Part 4. For each $j_1 \in [n]$, for each $i \in [d^2]$,
\begin{align*}
    \frac{\d u(x)_{j_1} }{\d x_i} = u(x)_{j_1} \circ \A_{[j_1],i}
\end{align*}
\item Part 5. For each $j_1 \in [n]$, for each $i \in [d^2]$,
\begin{align*}
    \frac{\d \alpha(x)_{j_1}}{\d x} = \langle u(x)_{j_1} , \A_{[j_1], i} \rangle
\end{align*}
\item Part 6. For each $j_1 \in [n]$, for each $i\in [d^2]$,
\begin{align*}
    \frac{\d \alpha(x)_{i_1}^{-1} }{\d x_i} = - \alpha(x)_{j_1}^{-1} \cdot \langle f(x)_{j_1}, \A_{[j_1],i} \rangle
\end{align*}
\item Part 7. For each $j_1 \in [n]$, for each $i \in [d^2]$,
\begin{align*}
    \frac{\d f(x)_{j_1}}{ \d x_i} = f(x)_{j_1} \circ \A_{[j_1],i} - f(x)_{j_1} \cdot \langle f(x)_{j_1}, \A_{[j_1],i} \rangle.
\end{align*}
\item Part 8. For each $j_1 \in [n]$, for each $i \in [d^2]$,
\begin{align*}
    \frac{\d c(x)_{j_1}}{ \d x_i} = \frac{\d f(x)_{j_1}}{\d x_i}
\end{align*}
\item Part 9. For each $i \in [d^2]$,
\begin{align*}
    & ~ \frac{\d L_{\exp}(x)}{ \d x_i} \\
    = & ~ \sum_{j_1=1}^n (\langle c(x)_{j_1}, f(x)_{j_1} \circ \A_{[j_1],i} \rangle - \langle c(x)_{j_1}, f(x)_{j_1} \rangle \cdot \langle f(x)_{j_1}, \A_{[j_1],i} \rangle).
\end{align*}
\end{itemize}
\end{lemma}

\ifdefined\isarxiv

\else

\begin{proof}

{\bf Proof of Part 1.}

We have
\begin{align*}
    \frac{\d  (\mathsf{A} x)}{\d x_i}
    = & ~ \frac{\mathsf{A} \d x}{\d x_i}\\
    = & ~ \mathsf{A}_{*,i},
\end{align*}
where the first step follows form Fact~\ref{fac:derivative_rules} and the second step follows from the fact that only the $i$-th entry of $\frac{\d x}{\d x_i}$ is $1$ and other entries of it are $0$.

{\bf Proof of Part 2.}

We have
\begin{align*}
    \frac{\d \exp( \mathsf{A} x)}{\d x_i}
    = & ~ \exp( \mathsf{A} x) \circ \frac{\d ( \mathsf{A} x)}{\d x_i}\\
    = & ~ \exp( \mathsf{A} x ) \circ \mathsf{A}_{*,i}
\end{align*}
where the first step follows from the chain rule and the second step follows from {\bf Part 1} of this Lemma.

{\bf Proof of Part 3.}
We have
\begin{align*}
    \frac{\d \A_{[j_1],*} x }{\d x_i} 
    = & ~ \underbrace{ \A_{[j_1],*} }_{n \times d^2 } \underbrace{ \frac{\d x }{\d x_i} }_{d^2 \times 1} \\
    = & ~ \underbrace{ \A_{[j_1],i} }_{n \times 1},
\end{align*}
where the first step follows form Fact~\ref{fac:derivative_rules} and the second step follows from the fact that only the $i$-th entry of $\frac{\d x}{\d x_i}$ is $1$ and other entries of it are $0$.  

{\bf Proof of Part 4.}
We have
\begin{align*}
    \frac{\d u(x)_{j_1} }{\d x_i} 
    = & ~ \frac{\d \exp( \A_{[j_1],*} x ) }{\d x_i}\\
    = & ~ \exp( \A_{[j_1],*} x ) \circ \frac{\d \A_{[j_1],*} x }{\d x_i}\\
    = & ~ u(x)_{j_1} \circ \A_{[j_1],i},
\end{align*}
where the first step follows from the definition of $u(x)_{j_1}$ (see Definition~\ref{def:u}), the second step follows from the chain rule, and the third step follows from {\bf Part 3} of this Lemma. 

{\bf Proof of Part 5.}

Let $j_1 \in [n]$. Let $i \in [d^2]$.

We have 
\begin{align*}
    \frac{\d \alpha(x)_{j_1}}{ \d x_i} 
    = & ~ \frac{ \d \langle \exp( \mathsf{A}_{[j_1],*} x ),  {\bf 1}_n \rangle }{\d x_i} \\
    = & ~ \langle \frac{  \d \exp( \mathsf{A}_{[j_1],*} x ) }{\d x_i} ,  {\bf 1}_n \rangle \\
    = & ~ \langle  u(x)_{j_1}   \circ  \mathsf{A}_{[j_1], i} , {\bf 1}_n \rangle \\
    = & ~ \langle u(x)_{j_1} , \A_{[j_1],i} \rangle,
\end{align*}
where the first step follows from the definition of $\alpha(x)_{j_1}$ (see Definition~\ref{def:alpha}), the second step follows from the definition of the inner product and $\frac{\d {\bf 1}_n}{\d x_i} = {\bf 0}_n$, the third step follows from {\bf Part 4} of this Lemma, and the last step follows from Fact~\ref{fac:vector_properties}.

{\bf Proof of Part 6.}
We have
\begin{align*}
\frac{\d \alpha(x)_{j_1}^{-1} }{\d x_i} 
= & ~ -1 \cdot \alpha(x)_{j_1}^{-2} \cdot \frac{\d \alpha(x)_{j_1}}{ \d x_i} \\
= & ~ -1 \cdot \alpha(x)_{j_1}^{-2} \cdot \langle u(x)_{j_1} , \A_{[j_1],i} \rangle \\
= & ~ - \alpha(x)_{j_1}^{-1} \cdot \langle f(x)_{j_1}, \A_{[j_1],i} \rangle
\end{align*}
where the first step follows from Fact~\ref{fac:derivative_rules}, the second step follows from {\bf Part 5} of this Lemma, and the third step follows from the definition of $f$ (see Definition~\ref{def:f}).

{\bf Proof of Part 7.}

We have
\begin{align*}
    & ~ \frac{\d f(x)_{j_1}}{ \d x_i}  \\
    = & ~ \frac{\d ( \alpha(x)_{j_1}^{-1} u(x)_{j_1}) }{ \d x_i} \\
    = & ~ \alpha(x)_{j_1}^{-1} \cdot \frac{\d u(x)_{j_1} }{ \d x_i} + \frac{\d \alpha(x)_{j_1}^{-1}}{\d x_i} \cdot u(x)_{j_1} \\
    = & ~ \alpha(x)_{j_1}^{-1} \cdot  u(x)_{j_1} \circ \A_{[j_1],i} - \alpha(x)_{j_1}^{-1} \cdot \langle f(x)_{j_1}, \A_{[j_1],i} \rangle \cdot u(x)_{j_1} \\
    = & ~ f(x)_{j_1} \circ \A_{[j_1],i} - f(x)_{j_1} \cdot \langle f(x)_{j_1}, \A_{[j_1],i} \rangle,
\end{align*}
where the first step follows from the definition of $f$ (see Definition~\ref{def:f}), the second step follows from Fact~\ref{fac:derivative_rules}, the third step follows from {\bf Part 4} and {\bf Part 6} of this Lemma, and the last step follows from the definition of $f$ (see Definition~\ref{def:f}).

{\bf Proof of Part 8.}

We have
\begin{align*}
    \frac{\d c(x)_{j_1}}{ \d x_i}
    = & ~ \frac{\d (f(x)_{j_1} - b_{[j_1]})}{ \d x_i} \\
    = & ~ \frac{\d f(x)_{j_1}}{ \d x_i} - \frac{\d b_{[j_1]}}{ \d x_i}\\
    = & ~ \frac{\d f(x)_{j_1}}{ \d x_i},
\end{align*}
where the first step follows from the definition of $c(x)_{j_1}$ (see Definition~\ref{def:c}), the second step follows from Fact~\ref{fac:derivative_rules}, and the last step follows from the fact that $b_{[j_1]}$ does not contain $x_i$.

{\bf Proof of Part 9.}

\begin{align*}
    & ~ \frac{\d L_{\exp}(x)}{ \d x_i} \\
    = & ~ 0.5 \frac{\d}{\d x_i} \sum_{j_1=1}^n \langle c(x)_{j_1}, c(x)_{j_1} \rangle \\
    = & ~ \sum_{j_1 = 1}^n \langle c(x)_{j_1}, \frac{\d c(x)_{j_1}}{\d x_i} \rangle\\
    = & ~ \sum_{j_1=1}^n \langle c(x)_{j_1}, f(x)_{j_1} \circ \A_{[j_1],i} - f(x)_{j_1} \cdot \langle f(x)_{j_1}, \A_{[j_1],i} \rangle \rangle \\
    = & ~ \sum_{j_1=1}^n (\langle c(x)_{j_1}, f(x)_{j_1} \circ \A_{[j_1],i} \rangle - \langle c(x)_{j_1}, f(x)_{j_1} \rangle \cdot \langle f(x)_{j_1}, \A_{[j_1],i} \rangle),
\end{align*}
where the first step follows from the definition of $L_{\exp}(x)$ (see Definition~\ref{def:L_exp}), the second step follows from Fact~\ref{fac:derivative_rules}, the third step follows from {\bf Part 7} and {\bf Part 8} of this Lemma, and the last step follows from Fact~\ref{fac:vector_properties}.

\end{proof}

 \fi
\section{Hessian}
\label{sec:hessian}

In this section, our attention is directed towards the Hessian property inherent in our loss function. This investigation serves as a preparatory step for substantiating the convergence proof of our training procedure. While \cite{dls23} outlines a singular version of a similar problem, we aim to showcase that our computations extend this scenario by a factor of $n$. Drawing upon the Hessian property expounded upon in \cite{dls23}, it becomes evident that our loss function similarly exhibits this particular property.

In Section~\ref{sub:hessian:u}, we compute the second order derivative of $u(x)_{j_1}$. In Section~\ref{sub:hessian:a}, we compute the second order derivative of $\alpha(x)_{j_1}$. In Section~\ref{sub:hessian:a-1}, we compute the second order derivative of $\alpha(x)_{j_1}^{-1}$. In Section~\ref{sub:hessian:f}, we compute the second order derivative of $f(x)_{j_1}$. In Section~\ref{sub:hessian:l}, we compute the second order derivative of $L_{\exp}$. In Section~\ref{sub:hessian:loss}, we compute the hessian of a single loss. In Section~\ref{sub:hessian:b1_b2}, we simplify the result that we get.

\subsection{Second Order Derivatives of \texorpdfstring{$u(x)_{j_1}$}{}}
\label{sub:hessian:u}

In this section, we start to compute the second-order derivative of $u(x)_{j_1}$.

\begin{lemma}
If the following conditions hold
\begin{itemize}
    \item Let $u$ be defined as in Definition~\ref{def:u}.
    \item Let $x \in \R^{d^2}$, satisfying $x = \vect(X)$.
    \item Let $A_1, A_2 \in \R^{n \times d}$.
    \item Let $\mathsf{A} = A_1 \otimes A_2$.
\end{itemize}

Then, we have
\begin{itemize}
    \item For each $i\in [d^2]$,
    \begin{align*}
        \frac{\d^2 u(x)_{j_1} }{\d x_i^2} = u(x)_{j_1} \circ \A_{[j_1],i} \circ \A_{[j_1],i}.
    \end{align*}
    \item For each $i\in [d^2], l \in [d^2]$
    \begin{align*}
        \frac{\d^2 u(x)_{j_1} }{\d x_i \d x_l} = u(x)_{j_1} \circ \A_{[j_1],l} \circ \A_{[j_1],i}.
    \end{align*}
\end{itemize}
\end{lemma}

\begin{proof}

We have
\begin{align*}
    \frac{\d^2 u(x)_{j_1} }{\d x_i^2}
    = & ~ \frac{\d}{\d x_i} (\frac{\d u(x)_{j_1} }{\d x_i})\\
    = & ~ \frac{\d (u(x)_{j_1} \circ \A_{[j_1],i})}{\d x_i}\\
    = & ~ \frac{\d (u(x)_{j_1})}{\d x_i} \circ \A_{[j_1],i} + u(x)_{j_1} \circ \frac{\d (\A_{[j_1],i})}{\d x_i}\\
    = & ~ u(x)_{j_1} \circ \A_{[j_1],i} \circ \A_{[j_1],i}
\end{align*}
where the first step follows from simple algebra, the second step follows from {\bf Part 4} of Lemma~\ref{lem:basic_derivatives}, the third step follows from Fact~\ref{fac:derivative_rules}, and the last step follows from {\bf Part 4} of Lemma~\ref{lem:basic_derivatives} and $\frac{\d (\A_{[j_1],i})}{\d x_i} = 0$.

Also, we can get
\begin{align*}
    \frac{\d^2 u(x)_{j_1} }{\d x_i \d x_l} 
    = & ~ \frac{\d}{\d x_l} (\frac{\d u(x)_{j_1} }{\d x_i})\\
    = & ~ \frac{\d (u(x)_{j_1} \circ \A_{[j_1],i})}{\d x_l}\\
    = & ~ \frac{\d (u(x)_{j_1})}{\d x_l} \circ \A_{[j_1],i} + u(x)_{j_1} \circ \frac{\d (\A_{[j_1],i})}{\d x_l}\\
    = & ~ u(x)_{j_1} \circ \A_{[j_1],l} \circ \A_{[j_1],i}
\end{align*}
where the first step follows from simple algebra, the second step follows from {\bf Part 4} of Lemma~\ref{lem:basic_derivatives}, the third step follows from Fact~\ref{fac:derivative_rules}, and the last step follows from {\bf Part 4} of Lemma~\ref{lem:basic_derivatives} and $\frac{\d (\A_{[j_1],i})}{\d x_i} = 0$.
\end{proof}

\subsection{Second Order Derivatives of \texorpdfstring{$\alpha(x)_{j_1}$}{}}
\label{sub:hessian:a}

In this section, we start to compute the second-order derivative of $\alpha(x)_{j_1}$.

\begin{lemma}
If the following conditions hold
\begin{itemize}
    \item Let $\alpha$ be defined as in Definition~\ref{def:alpha}.
    \item Let $x \in \R^{d^2}$, satisfying $x = \vect(X)$.
    \item Let $A_1, A_2 \in \R^{n \times d}$.
    \item Let $\mathsf{A} = A_1 \otimes A_2$.
\end{itemize}

Then, we have
\begin{itemize}
    \item For each $i \in [d^2]$,
    \begin{align*}
     \frac{\d^2 \alpha(x)_{j_1} }{\d x_i^2} = \langle u(x)_{j_1} , \A_{[j_1],i}^2 \rangle.
    \end{align*}
    \item For each $i, l \in [d^2]$,
    \begin{align*}
     \frac{\d^2 \alpha(x)_{j_1} }{\d x_i \d x_l} = \langle u(x)_{j_1}, \A_{[j_1],i} \circ \A_{[j_1],l} \rangle.
    \end{align*}
\end{itemize}
\end{lemma}
\begin{proof}

We have
\begin{align*}
    \frac{\d^2 \alpha(x)_{j_1} }{\d x_i^2}
    = & ~ \frac{\d}{\d x_i} (\frac{\d \alpha(x)_{j_1} }{\d x_i})\\
    = & ~ \frac{\d (\langle u(x)_{j_1} , \A_{[j_1],i} \rangle)}{\d x_i}\\
    = & ~ \langle \frac{\d u(x)_{j_1}}{\d x_i} , \A_{[j_1],i} \rangle + \langle u(x)_{j_1} , \frac{\d \A_{[j_1],i}}{\d x_i} \rangle\\
    = & ~ \langle u(x)_{j_1} \circ \A_{[j_1],i} , \A_{[j_1],i} \rangle\\
    = & ~ \langle u(x)_{j_1} , \A_{[j_1],i}^2 \rangle,
\end{align*}
where the first step follows from simple algebra, the second step follows from {\bf Part 5} of Lemma~\ref{lem:basic_derivatives}, the third step follows from the definition of the inner product, the fourth step follows from $\frac{\d \A_{[j_1],i}}{\d x_i} = 0$, and the last step follows from Fact~\ref{fac:vector_properties}.

Also, we can get
\begin{align*}
    \frac{\d^2 \alpha(x)_{j_1} }{\d x_i \d x_l}
    = & ~ \frac{\d}{\d x_l} (\frac{\d \alpha(x)_{j_1} }{\d x_i})\\
    = & ~ \frac{\d (\langle u(x)_{j_1} , \A_{[j_1],i} \rangle)}{\d x_l}\\
    = & ~ \langle \frac{\d u(x)_{j_1}}{\d x_l} , \A_{[j_1],i} \rangle + \langle u(x)_{j_1} , \frac{\d \A_{[j_1],i}}{\d x_l} \rangle\\
    = & ~ \langle u(x)_{j_1} \circ \A_{[j_1],l} , \A_{[j_1],i} \rangle\\
    = & ~ \langle u(x)_{j_1}, \A_{[j_1],i} \circ \A_{[j_1],l} \rangle,
\end{align*}
where the first step follows from simple algebra, the second step follows from {\bf Part 5} of Lemma~\ref{lem:basic_derivatives}, the third step follows from the definition of the inner product, the fourth step follows from $\frac{\d \A_{[j_1],i}}{\d x_i} = 0$, and the last step follows from Fact~\ref{fac:vector_properties}.
\end{proof}

\subsection{Second Order Derivatives of \texorpdfstring{$\alpha(x)_{j_1}^{-1}$}{}}
\label{sub:hessian:a-1}

In this section, we start to compute the second-order derivative of $\alpha(x)_{j_1}^{-1}$.

\begin{lemma}
\label{lem:second_derivative_alpha-1}
If the following conditions hold
\begin{itemize}
    \item Let $\alpha$ be defined as in Definition~\ref{def:alpha}.
    \item Let $f$ be defined as in Definition~\ref{def:f}.
    \item Let $x \in \R^{d^2}$, satisfying $x = \vect(X)$.
    \item Let $A_1, A_2 \in \R^{n \times d}$.
    \item Let $\mathsf{A} = A_1 \otimes A_2$.
\end{itemize}

Then, we have
\begin{itemize}
    \item For each $i \in [d^2]$, 
    \begin{align*}
     \frac{\d^2 \alpha(x)_{j_1}^{-1} }{\d x_i^2} = 2\alpha(x)_{j_1}^{-1} \cdot \langle f(x)_{j_1}, \A_{[j_1],i} \rangle^2 - \alpha(x)_{j_1}^{-1} \cdot \langle f(x)_{j_1}, \A_{[j_1],i}^2 \rangle.
    \end{align*}
    \item For each $i,l \in [d^2]$,
    \begin{align*}
     \frac{\d^2 \alpha(x)_{j_1}^{-1} }{\d x_i \d x_l} = 2 \alpha(x)_{j_1}^{-1} \cdot \langle f(x)_{j_1}, \A_{[j_1],l} \rangle \cdot \langle f(x)_{j_1}, \A_{[j_1],i} \rangle - \alpha(x)_{j_1}^{-1} \cdot \langle f(x)_{j_1}, \A_{[j_1],i} \circ \A_{[j_1],l} \rangle.
    \end{align*}
\end{itemize}
\end{lemma}
\begin{proof}

    We have
    \begin{align}\label{eq:alpha_dxi2}
        \frac{\d^2 \alpha(x)_{j_1}^{-1} }{\d x_i^2}
        = & ~ \frac{\d}{\d x_i} (\frac{\d \alpha(x)_{j_1}^{-1} }{\d x_i}) \notag\\
        = & ~ \frac{\d (- \alpha(x)_{j_1}^{-1} \cdot \langle f(x)_{j_1}, \A_{[j_1],i} \rangle)}{\d x_i} \notag\\
        = & ~ \frac{\d (- \alpha(x)_{j_1}^{-1})}{\d x_i} \cdot \langle f(x)_{j_1}, \A_{[j_1],i} \rangle +  (- \alpha(x)_{j_1}^{-1}) \cdot \frac{\d\langle f(x)_{j_1}, \A_{[j_1],i} \rangle}{\d x_i} \notag\\
        = & ~ \alpha(x)_{j_1}^{-1} \cdot \langle f(x)_{j_1}, \A_{[j_1],i} \rangle^2 + (- \alpha(x)_{j_1}^{-1}) \cdot \frac{\d\langle f(x)_{j_1}, \A_{[j_1],i} \rangle}{\d x_i},
    \end{align}
    where the first step follows from simple algebra, the second step follows from {\bf Part 6} of Lemma~\ref{lem:basic_derivatives}, the third step follows from Fact~\ref{fac:derivative_rules}, and the last step follows from {\bf Part 6} of Lemma~\ref{lem:basic_derivatives}.

    To compute the second term of Eq.~\eqref{eq:alpha_dxi2}, we have
    \begin{align}\label{eq:alpha_dxi2:second_term}
        \frac{\d\langle f(x)_{j_1}, \A_{[j_1],i} \rangle}{\d x_i}
        = & ~ \langle \frac{\d f(x)_{j_1}}{\d x_i}, \A_{[j_1],i} \rangle  \notag \\
        = & ~ \langle f(x)_{j_1} \circ \A_{[j_1],i} - f(x)_{j_1} \cdot \langle f(x)_{j_1}, \A_{[j_1],i} \rangle, \A_{[j_1],i} \rangle  \notag \\
        = & ~ \langle f(x)_{j_1} \circ \A_{[j_1],i}, \A_{[j_1],i} \rangle - \langle f(x)_{j_1} \cdot \langle f(x)_{j_1}, \A_{[j_1],i} \rangle, \A_{[j_1],i} \rangle \notag \\
        = & ~ \langle f(x)_{j_1}, \A_{[j_1],i}^2 \rangle - \langle f(x)_{j_1}, \A_{[j_1],i} \rangle^2,
    \end{align}
    where the first step follows from the definition of the inner product, the second step follows from {\bf Part 7} of Lemma~\ref{lem:basic_derivatives} and $\frac{\d\A_{[j_1],i}}{\d x_i} = 0$, the third step follows from Fact~\ref{fac:vector_properties}, and the last step follows from Fact~\ref{fac:vector_properties}.

    Combining Eq.~\eqref{eq:alpha_dxi2} and Eq.~\eqref{eq:alpha_dxi2:second_term}, we have
    \begin{align*}
        \frac{\d^2 \alpha(x)_{j_1}^{-1} }{\d x_i^2}
        = & ~ \alpha(x)_{j_1}^{-1} \cdot \langle f(x)_{j_1}, \A_{[j_1],i} \rangle^2 - \alpha(x)_{j_1}^{-1} \cdot (\langle f(x)_{j_1}, \A_{[j_1],i}^2 \rangle - \langle f(x)_{j_1}, \A_{[j_1],i} \rangle^2)\\
        = & ~ \alpha(x)_{j_1}^{-1} \cdot \langle f(x)_{j_1}, \A_{[j_1],i} \rangle^2 - \alpha(x)_{j_1}^{-1} \cdot \langle f(x)_{j_1}, \A_{[j_1],i}^2 \rangle + \alpha(x)_{j_1}^{-1} \cdot \langle f(x)_{j_1}, \A_{[j_1],i} \rangle^2\\
        = & ~ 2\alpha(x)_{j_1}^{-1} \cdot \langle f(x)_{j_1}, \A_{[j_1],i} \rangle^2 - \alpha(x)_{j_1}^{-1} \cdot \langle f(x)_{j_1}, \A_{[j_1],i}^2 \rangle,
    \end{align*}
    where the second and the third step both follow from simple algebra. 

    Then, to compute $\frac{\d^2 \alpha(x)_{j_1}^{-1} }{\d x_i \d x_l}$, we have
    \begin{align}\label{eq:alpha_dxixl}
        \frac{\d^2 \alpha(x)_{j_1}^{-1} }{\d x_i \d x_l}
        = & ~ \frac{\d}{\d x_l} (\frac{\d \alpha(x)_{j_1}^{-1} }{\d x_i}) \notag\\
        = & ~ \frac{\d (- \alpha(x)_{j_1}^{-1} \cdot \langle f(x)_{j_1}, \A_{[j_1],i} \rangle)}{\d x_l} \notag\\
        = & ~ \frac{\d (- \alpha(x)_{j_1}^{-1})}{\d x_l} \cdot \langle f(x)_{j_1}, \A_{[j_1],i} \rangle +  (- \alpha(x)_{j_1}^{-1}) \cdot \frac{\d\langle f(x)_{j_1}, \A_{[j_1],i} \rangle}{\d x_l} \notag\\
        = & ~ \alpha(x)_{j_1}^{-1} \cdot \langle f(x)_{j_1}, \A_{[j_1],l} \rangle \cdot \langle f(x)_{j_1}, \A_{[j_1],i} \rangle - \alpha(x)_{j_1}^{-1} \cdot \frac{\d\langle f(x)_{j_1}, \A_{[j_1],i} \rangle}{\d x_l},
    \end{align}
    where the first step follows from simple algebra, the second step follows from {\bf Part 6} of Lemma~\ref{lem:basic_derivatives}, the third step follows from Fact~\ref{fac:derivative_rules}, and the last step follows from {\bf Part 6} of Lemma~\ref{lem:basic_derivatives}.

    To compute the second term of Eq.~\eqref{eq:alpha_dxixl}, we have
    \begin{align}\label{eq:alpha_dxixl:second_term}
        \frac{\d\langle f(x)_{j_1}, \A_{[j_1],i} \rangle}{\d x_l}
        = & ~ \langle \frac{\d f(x)_{j_1}}{\d x_l}, \A_{[j_1],i} \rangle + \langle  f(x)_{j_1}, \frac{\d\A_{[j_1],i}}{\d x_l} \rangle \notag \\
        = & ~ \langle f(x)_{j_1} \circ \A_{[j_1],l} - f(x)_{j_1} \cdot \langle f(x)_{j_1}, \A_{[j_1],l} \rangle, \A_{[j_1],i} \rangle  \notag \\
        = & ~ \langle f(x)_{j_1} \circ \A_{[j_1],l}, \A_{[j_1],i} \rangle - \langle f(x)_{j_1} \cdot \langle f(x)_{j_1}, \A_{[j_1],l} \rangle, \A_{[j_1],i} \rangle \notag \\
        = & ~ \langle f(x)_{j_1}, \A_{[j_1],i} \circ \A_{[j_1],l} \rangle - \langle f(x)_{j_1}, \A_{[j_1],l} \rangle \cdot \langle f(x)_{j_1}, \A_{[j_1],i} \rangle,
    \end{align}
    where the first step follows from the definition of the inner product, the second step follows from {\bf Part 7} of Lemma~\ref{lem:basic_derivatives} and $\frac{\d\A_{[j_1],i}}{\d x_l} = 0$, the third step follows from Fact~\ref{fac:vector_properties}, and the last step follows from Fact~\ref{fac:vector_properties}.

    By simple algebra, we can combine Eq.~\eqref{eq:alpha_dxixl} and Eq.~\eqref{eq:alpha_dxixl:second_term} as:
    \begin{align*}
        \frac{\d^2 \alpha(x)_{j_1}^{-1} }{\d x_i \d x_l} = 2 \alpha(x)_{j_1}^{-1} \cdot \langle f(x)_{j_1}, \A_{[j_1],l} \rangle \cdot \langle f(x)_{j_1}, \A_{[j_1],i} \rangle - \alpha(x)_{j_1}^{-1} \cdot \langle f(x)_{j_1}, \A_{[j_1],i} \circ \A_{[j_1],l} \rangle.
    \end{align*}
\end{proof}

\subsection{Second Order Derivatives of \texorpdfstring{$f(x)_{j_1}$}{}}
\label{sub:hessian:f}

In this section, we start to compute the second-order derivative of $f(x)_{j_1}$.

\begin{lemma}
\label{lem:second_derivative_f}
If the following conditions hold
\begin{itemize}
    \item Let $f$ be defined as in Definition~\ref{def:f}.
    \item Let $x \in \R^{d^2}$, satisfying $x = \vect(X)$.
    \item Let $A_1, A_2 \in \R^{n \times d}$.
    \item Let $\mathsf{A} = A_1 \otimes A_2$.
\end{itemize}

Then, we have
\begin{itemize}
    \item For each $i \in [d^2]$,
    \begin{align*}
     \frac{\d^2 f(x)_{j_1} }{\d x_i^2} 
     = & ~ f(x)_{j_1} \circ \A_{[j_1],i}^2 - 2 f(x)_{j_1} \circ \A_{[j_1],i} \cdot \langle f(x)_{j_1}, \A_{[j_1],i} \rangle \\
    & ~ - f(x)_{j_1} \cdot \langle f(x)_{j_1}, \A_{[j_1],i}^2 \rangle + 2 f(x)_{j_1} \cdot \langle f(x)_{j_1}, \A_{[j_1],i} \rangle^2.
    \end{align*}
    \item For each $i,l \in [d^2]$,
    \begin{align*}
    \frac{\d^2 f(x)_{j_1} }{\d x_i \d x_l} 
    = & ~ f(x)_{j_1} \circ \A_{[j_1],i} \circ \A_{[j_1],l} - f(x)_{j_1} \circ \A_{[j_1],i} \cdot \langle f(x)_{j_1}, \A_{[j_1],l} \rangle \\
    & ~ - f(x)_{j_1} \circ \A_{[j_1],l} \cdot \langle f(x)_{j_1}, \A_{[j_1],i} \rangle + 2f(x)_{j_1} \cdot \langle f(x)_{j_1}, \A_{[j_1],i} \rangle \cdot \langle f(x)_{j_1}, \A_{[j_1],l} \rangle \\
    & ~ - f(x)_{j_1} \cdot \langle f(x)_{j_1}, \A_{[j_1],i} \circ \A_{[j_1],l} \rangle.
\end{align*}
\end{itemize}
\end{lemma}
\begin{proof}

We first consider $\frac{\d^2 f(x)_{j_1} }{\d x_i^2}$. 

We have 
\begin{align}\label{eq:f_dxi2}
    \frac{\d^2 f(x)_{j_1} }{\d x_i^2} 
    = & ~ \frac{\d }{\d x_i}  (\frac{\d f(x)_{j_1} }{\d x_i}) \notag\\
    = & ~ \frac{\d (f(x)_{j_1} \circ \A_{[j_1],i} - f(x)_{j_1} \cdot \langle f(x)_{j_1}, \A_{[j_1],i} \rangle)}{\d x_i} \notag\\
    = & ~ \frac{\d f(x)_{j_1} \circ \A_{[j_1],i}}{\d x_i} - \frac{\d f(x)_{j_1} \cdot \langle f(x)_{j_1}, \A_{[j_1],i} \rangle}{\d x_i},
\end{align}
where the first step follows from simple algebra, the second step is due to Fact~\ref{fac:derivative_rules}, and the third step is based on Fact~\ref{fac:derivative_rules}.

To compute the first term of Eq.~\eqref{eq:f_dxi2}, we have
\begin{align}\label{eq:f_dxi2:1}
    \frac{\d f(x)_{j_1} \circ \A_{[j_1],i}}{\d x_i}
    = & ~ f(x)_{j_1} \circ \frac{\d \A_{[j_1],i}}{\d x_i} + \frac{\d f(x)_{j_1}}{\d x_i} \circ \A_{[j_1],i} \notag \\
    = & ~ (f(x)_{j_1} \circ \A_{[j_1],i} - f(x)_{j_1} \cdot \langle f(x)_{j_1}, \A_{[j_1],i} \rangle) \circ \A_{[j_1],i} \notag \\
    = & ~ f(x)_{j_1} \circ \A_{[j_1],i}^2 - f(x)_{j_1} \circ \A_{[j_1],i} \cdot \langle f(x)_{j_1}, \A_{[j_1],i} \rangle,
\end{align}
where the first step follows from Fact~\ref{fac:derivative_rules}, the second step follows from $\frac{\d \A_{[j_1],i}}{\d x_i} = 0$ and {\bf Part 7} of Lemma~\ref{lem:basic_derivatives}, and the last step follows from the property of Hadamard product.

To compute the second term of Eq.~\eqref{eq:f_dxi2}, we have
\begin{align}\label{eq:f_dxi2:2}
    \frac{\d f(x)_{j_1} \cdot \langle f(x)_{j_1}, \A_{[j_1],i} \rangle}{\d x_i}
    = & ~ \frac{\d f(x)_{j_1}}{\d x_i} \cdot \langle f(x)_{j_1}, \A_{[j_1],i} \rangle + f(x)_{j_1} \cdot \frac{\d \langle f(x)_{j_1}, \A_{[j_1],i} \rangle}{\d x_i} \notag \\
    = & ~ (f(x)_{j_1} \circ \A_{[j_1],i} - f(x)_{j_1} \cdot \langle f(x)_{j_1}, \A_{[j_1],i} \rangle) \cdot \langle f(x)_{j_1}, \A_{[j_1],i} \rangle \notag\\
    & ~ + f(x)_{j_1} \cdot \frac{\d \langle f(x)_{j_1}, \A_{[j_1],i} \rangle}{\d x_i} \notag \\
    = & ~ (f(x)_{j_1} \circ \A_{[j_1],i} - f(x)_{j_1} \cdot \langle f(x)_{j_1}, \A_{[j_1],i} \rangle) \cdot \langle f(x)_{j_1}, \A_{[j_1],i} \rangle \notag\\
    & ~ + f(x)_{j_1} \cdot (\langle f(x)_{j_1}, \A_{[j_1],i}^2 \rangle - \langle f(x)_{j_1}, \A_{[j_1],i} \rangle^2) \notag \\
    = & ~ f(x)_{j_1} \circ \A_{[j_1],i} \cdot \langle f(x)_{j_1}, \A_{[j_1],i} \rangle - f(x)_{j_1} \cdot \langle f(x)_{j_1}, \A_{[j_1],i} \rangle^2 \notag\\
    & ~ + f(x)_{j_1} \cdot \langle f(x)_{j_1}, \A_{[j_1],i}^2 \rangle - f(x)_{j_1} \cdot \langle f(x)_{j_1}, \A_{[j_1],i} \rangle^2 \notag \\
    = & ~ f(x)_{j_1} \circ \A_{[j_1],i} \cdot \langle f(x)_{j_1}, \A_{[j_1],i} \rangle + f(x)_{j_1} \cdot \langle f(x)_{j_1}, \A_{[j_1],i}^2 \rangle \notag\\
    & ~ - 2 f(x)_{j_1} \cdot \langle f(x)_{j_1}, \A_{[j_1],i} \rangle^2,
\end{align}
where the first step follows from Fact~\ref{fac:derivative_rules}, the second step follows from {\bf Part 7} of Lemma~\ref{lem:basic_derivatives}, the third step follows from the proof of Lemma~\ref{lem:second_derivative_alpha-1} (see Eq.~\eqref{eq:alpha_dxi2:second_term}), and the fourth and the fifth step follows from simple algebra.

Combining Eq.~\eqref{eq:f_dxi2}, Eq.~\eqref{eq:f_dxi2:1}, and Eq.~\eqref{eq:f_dxi2:2}, we have
\begin{align*}
    \frac{\d^2 f(x)_{j_1} }{\d x_i^2} 
    = & ~ f(x)_{j_1} \circ \A_{[j_1],i}^2 - 2 f(x)_{j_1} \circ \A_{[j_1],i} \cdot \langle f(x)_{j_1}, \A_{[j_1],i} \rangle \\
    & ~ - f(x)_{j_1} \cdot \langle f(x)_{j_1}, \A_{[j_1],i}^2 \rangle + 2 f(x)_{j_1} \cdot \langle f(x)_{j_1}, \A_{[j_1],i} \rangle^2.
\end{align*}

Now, we consider $\frac{\d^2 f(x)_{j_1} }{\d x_i \d x_l}$.

We have 
\begin{align}\label{eq:f_dxixl}
    \frac{\d^2 f(x)_{j_1} }{\d x_i \d x_l} 
    = & ~ \frac{\d }{\d x_l}  (\frac{\d f(x)_{j_1} }{\d x_i}) \notag\\
    = & ~ \frac{\d (f(x)_{j_1} \circ \A_{[j_1],i} - f(x)_{j_1} \cdot \langle f(x)_{j_1}, \A_{[j_1],i} \rangle)}{\d x_l} \notag\\
    = & ~ \frac{\d f(x)_{j_1} \circ \A_{[j_1],i}}{\d x_l} - \frac{\d f(x)_{j_1} \cdot \langle f(x)_{j_1}, \A_{[j_1],i} \rangle}{\d x_l},
\end{align}
where the first step follows from simple algebra, the second step follows from {\bf Part 7} of Lemma~\ref{lem:basic_derivatives}, and the third step follows from Fact~\ref{fac:derivative_rules}.

To compute the first term of Eq.~\eqref{eq:f_dxixl}, we have
\begin{align}\label{eq:f_dxixl:1}
    \frac{\d f(x)_{j_1} \circ \A_{[j_1],i}}{\d x_l}
    = & ~ f(x)_{j_1} \circ \frac{\d \A_{[j_1],i}}{\d x_l} + \frac{\d f(x)_{j_1}}{\d x_l} \circ \A_{[j_1],i} \notag \\
    = & ~ (f(x)_{j_1} \circ \A_{[j_1],l} - f(x)_{j_1} \cdot \langle f(x)_{j_1}, \A_{[j_1],l} \rangle) \circ \A_{[j_1],i} \notag \\
    = & ~ f(x)_{j_1} \circ \A_{[j_1],i} \circ \A_{[j_1],l} - f(x)_{j_1} \circ \A_{[j_1],i} \cdot \langle f(x)_{j_1}, \A_{[j_1],l} \rangle,
\end{align}
where the first step follows from Fact~\ref{fac:derivative_rules}, the second step follows from $\frac{\d \A_{[j_1],i}}{\d x_l} = 0$ and {\bf Part 7} of Lemma~\ref{lem:basic_derivatives}, the third step follows from the property of Hadamard product.

To compute the second term of Eq.~\eqref{eq:f_dxixl}, we have
\begin{align}\label{eq:f_dxixl:2}
    \frac{\d f(x)_{j_1} \cdot \langle f(x)_{j_1}, \A_{[j_1],i} \rangle}{\d x_l}
    = & ~ \frac{\d f(x)_{j_1}}{\d x_l} \cdot \langle f(x)_{j_1}, \A_{[j_1],i} \rangle + f(x)_{j_1} \cdot \frac{\d \langle f(x)_{j_1}, \A_{[j_1],i} \rangle}{\d x_l} \notag \\
    = & ~ (f(x)_{j_1} \circ \A_{[j_1],l} - f(x)_{j_1} \cdot \langle f(x)_{j_1}, \A_{[j_1],l} \rangle) \cdot \langle f(x)_{j_1}, \A_{[j_1],i} \rangle \notag\\
    & ~ + f(x)_{j_1} \cdot \frac{\d \langle f(x)_{j_1}, \A_{[j_1],i} \rangle}{\d x_l} \notag \\
    = & ~ (f(x)_{j_1} \circ \A_{[j_1],l} - f(x)_{j_1} \cdot \langle f(x)_{j_1}, \A_{[j_1],l} \rangle) \cdot \langle f(x)_{j_1}, \A_{[j_1],i} \rangle \notag\\
    & ~ + f(x)_{j_1} \cdot (\langle f(x)_{j_1}, \A_{[j_1],i} \circ \A_{[j_1],l} \rangle - \langle f(x)_{j_1}, \A_{[j_1],l} \rangle \cdot \langle f(x)_{j_1}, \A_{[j_1],i} \rangle) \notag \\
    = & ~ f(x)_{j_1} \circ \A_{[j_1],l} \cdot \langle f(x)_{j_1}, \A_{[j_1],i} \rangle - f(x)_{j_1} \cdot \langle f(x)_{j_1}, \A_{[j_1],i} \rangle \cdot \langle f(x)_{j_1}, \A_{[j_1],l} \rangle \notag\\
    & ~ + f(x)_{j_1} \cdot \langle f(x)_{j_1}, \A_{[j_1],i} \circ \A_{[j_1],l} \rangle \notag\\
    & ~ - f(x)_{j_1} \cdot \langle f(x)_{j_1}, \A_{[j_1],l} \rangle \cdot \langle f(x)_{j_1}, \A_{[j_1],i} \rangle,
\end{align}
where the first step follows from Fact~\ref{fac:derivative_rules}, the second step follows from {\bf Part 7} of Lemma~\ref{lem:basic_derivatives}, the third step follows from the proof of Lemma~\ref{lem:second_derivative_alpha-1} (see Eq.~\eqref{eq:alpha_dxixl:second_term}), and the fourth and the fifth step follows from simple algebra.

Combining Eq.~\eqref{eq:f_dxixl}, Eq.~\eqref{eq:f_dxixl:1}, and Eq.~\eqref{eq:f_dxixl:2}, we have
\begin{align*}
    \frac{\d^2 f(x)_{j_1} }{\d x_i \d x_l} 
    = & ~ f(x)_{j_1} \circ \A_{[j_1],i} \circ \A_{[j_1],l} - f(x)_{j_1} \circ \A_{[j_1],i} \cdot \langle f(x)_{j_1}, \A_{[j_1],l} \rangle - f(x)_{j_1} \circ \A_{[j_1],l} \cdot \langle f(x)_{j_1}, \A_{[j_1],i} \rangle \\
    & ~ + 2f(x)_{j_1} \cdot \langle f(x)_{j_1}, \A_{[j_1],i} \rangle \cdot \langle f(x)_{j_1}, \A_{[j_1],l} \rangle - f(x)_{j_1} \cdot \langle f(x)_{j_1}, \A_{[j_1],i} \circ \A_{[j_1],l} \rangle.
\end{align*}

\end{proof}

\subsection{Second Order Derivatives of \texorpdfstring{$L_{\exp}$}{}}
\label{sub:hessian:l}

In this section, we start to compute the second-order derivative of $L_{\exp}$.

\begin{lemma}\label{lem:second_derivative_Lexp}
If the following conditions hold  
\begin{itemize}
    \item Let $L_{\exp}$ be defined as in Definition~\ref{def:L_exp}.
    \item Let $f$ be defined as in Definition~\ref{def:f}.
    \item Let $c$ be defined as in Definition~\ref{def:c}.
    \item Let $x \in \R^{d^2}$, satisfying $x = \vect(X)$.
    \item Let $A_1, A_2 \in \R^{n \times d}$.
    \item Let $\mathsf{A} = A_1 \otimes A_2$.
\end{itemize}

Then, we have
\begin{itemize}
    \item For each $i \in [d^2]$,
    \begin{align*}
        \frac{\d^2 L_{\exp} }{\d x_i^2} 
        = & ~ \sum_{j_1=1}^n (\|f(x)_{j_1} \circ \A_{[j_1],i}\|_2^2 - \langle f(x)_{j_1}, \A_{[j_1],i} \rangle \cdot \langle f(x)_{j_1}^2, \A_{[j_1],i} \rangle \notag\\
        & ~ ~ ~ + \langle c(x)_{j_1}, f(x)_{j_1} \circ \A_{[j_1],l}^2 \rangle - \langle c(x)_{j_1}, f(x)_{j_1} \circ \A_{[j_1],i} \rangle \cdot \langle f(x)_{j_1} , \A_{[j_1],i} \rangle)\\ 
        & ~ - \sum_{j_1=1}^n (\langle f(x)_{j_1} + c(x)_{j_1}, f(x)_{j_1} \circ \A_{[j_1],i} \rangle \cdot \langle f(x)_{j_1}, \A_{[j_1],i} \rangle \\
        & ~ ~ ~ - \langle f(x)_{j_1} + c(x)_{j_1}, f(x)_{j_1} \rangle \cdot \langle f(x)_{j_1}, \A_{[j_1],i} \rangle^2) \notag\\
        & ~ - \sum_{j_1=1}^n  (\langle c(x)_{j_1}, f(x)_{j_1} \rangle \cdot \langle f(x)_{j_1}, \A_{[j_1],i}^2 \rangle \notag\\
        & ~ ~ ~ - \langle c(x)_{j_1}, f(x)_{j_1} \rangle \cdot \langle f(x)_{j_1}, \A_{[j_1],i} \rangle^2),
    \end{align*}
    \item For each $i,l \in [d^2]$,
    \begin{align*}
        \frac{\d^2 L_{\exp} }{\d x_i \d x_l} 
        = & ~ \sum_{j_1=1}^n (\langle f(x)_{j_1} \circ \A_{[j_1],l}, f(x)_{j_1} \circ \A_{[j_1],i} \rangle - \langle f(x)_{j_1}, f(x)_{j_1} \circ \A_{[j_1],i} \rangle \cdot \langle f(x)_{j_1}, \A_{[j_1],l} \rangle \notag\\
        & ~ + \langle c(x)_{j_1}, f(x)_{j_1} \circ \A_{[j_1],i} \circ \A_{[j_1],l} \rangle + \langle c(x)_{j_1}, f(x)_{j_1} \circ \A_{[j_1],i} \rangle \cdot \langle f(x)_{j_1}, \A_{[j_1],l} \rangle) \notag \\
        - & ~ \sum_{j_1=1}^n (\langle f(x)_{j_1} + c(x)_{j_1}, f(x)_{j_1} \circ \A_{[j_1],l} \rangle \cdot \langle f(x)_{j_1}, \A_{[j_1],i} \rangle \notag\\
        & ~ - \langle f(x)_{j_1} + c(x)_{j_1}, f(x)_{j_1} \rangle \cdot \langle f(x)_{j_1}, \A_{[j_1],l} \rangle \cdot \langle f(x)_{j_1}, \A_{[j_1],i} \rangle) \notag\\
        - & ~ \sum_{j_1=1}^n  (\langle c(x)_{j_1}, f(x)_{j_1} \rangle \cdot \langle f(x)_{j_1}, \A_{[j_1],i} \circ \A_{[j_1],l} \rangle \notag\\ 
        & ~ - \langle c(x)_{j_1}, f(x)_{j_1} \rangle \cdot \langle f(x)_{j_1}, \A_{[j_1],l} \rangle \cdot \langle f(x)_{j_1}, \A_{[j_1],i} \rangle).
    \end{align*}
\end{itemize}
\end{lemma}
\begin{proof}
    We have
    \begin{align}\label{eq:Lexp_dxi2}
        \frac{\d^2 L_{\exp} }{\d x_i^2} 
        = & ~ \frac{\d}{\d x_i} (\frac{\d L_{\exp} }{\d x_i}) \notag\\
        = & ~ \frac{\d (\sum_{j_1=1}^n (\langle c(x)_{j_1}, f(x)_{j_1} \circ \A_{[j_1],i} \rangle - \langle c(x)_{j_1}, f(x)_{j_1} \rangle \cdot \langle f(x)_{j_1}, \A_{[j_1],i} \rangle))}{\d x_i} \notag\\
        = & ~ \frac{\d (\sum_{j_1=1}^n \langle c(x)_{j_1}, f(x)_{j_1} \circ \A_{[j_1],i} \rangle)}{\d x_i} - \frac{\d (\sum_{j_1=1}^n \langle c(x)_{j_1}, f(x)_{j_1} \rangle \cdot \langle f(x)_{j_1}, \A_{[j_1],i} \rangle)}{\d x_i} \notag\\
        = & ~ \sum_{j_1=1}^n \frac{\d (\langle c(x)_{j_1}, f(x)_{j_1} \circ \A_{[j_1],i} \rangle)}{\d x_i} - \sum_{j_1=1}^n \frac{\d (\langle c(x)_{j_1}, f(x)_{j_1} \rangle \cdot \langle f(x)_{j_1}, \A_{[j_1],i} \rangle)}{\d x_i} \notag\\
        = & ~ \sum_{j_1=1}^n \frac{\d (\langle c(x)_{j_1}, f(x)_{j_1} \circ \A_{[j_1],i} \rangle)}{\d x_i} \notag\\
        & ~ - \sum_{j_1=1}^n \frac{\d \langle c(x)_{j_1}, f(x)_{j_1} \rangle}{\d x_i} \cdot \langle f(x)_{j_1}, \A_{[j_1],i} \rangle \notag\\
        & ~ - \sum_{j_1=1}^n  \langle c(x)_{j_1}, f(x)_{j_1} \rangle \cdot \frac{\d\langle f(x)_{j_1}, \A_{[j_1],i}\rangle}{\d x_i},
    \end{align}
    where the first step follows from simple algebra, the second step follows from {\bf Part 9} of Lemma~\ref{lem:basic_derivatives}, the third step follows from the property of the summation, the fourth step follows from Fact~\ref{fac:derivative_rules}, and the last step follows from Fact~\ref{fac:derivative_rules}.

    First, we compute the first term of Eq.~\eqref{eq:Lexp_dxi2}:
    \begin{align}\label{eq:Lexp_dxi2:1}
        & ~ \frac{\d (\langle c(x)_{j_1}, f(x)_{j_1} \circ \A_{[j_1],i} \rangle)}{\d x_i} \notag \\
        = & ~  \langle \frac{\d c(x)_{j_1}}{\d x_i}, f(x)_{j_1} \circ \A_{[j_1],i} \rangle + \langle c(x)_{j_1}, \frac{\d f(x)_{j_1} \circ \A_{[j_1],i}}{\d x_i} \rangle \notag\\
        = & ~  \langle f(x)_{j_1} \circ \A_{[j_1],i} - f(x)_{j_1} \cdot \langle f(x)_{j_1}, \A_{[j_1],i} \rangle, f(x)_{j_1} \circ \A_{[j_1],i} \rangle + \langle c(x)_{j_1}, \frac{\d f(x)_{j_1} \circ \A_{[j_1],i}}{\d x_i} \rangle \notag\\
        = & ~  \langle f(x)_{j_1} \circ \A_{[j_1],i} - f(x)_{j_1} \cdot \langle f(x)_{j_1}, \A_{[j_1],i} \rangle, f(x)_{j_1} \circ \A_{[j_1],i} \rangle \notag\\
        & ~ + \langle c(x)_{j_1}, f(x)_{j_1} \circ \A_{[j_1],i}^2 - f(x)_{j_1} \circ \A_{[j_1],i} \cdot \langle f(x)_{j_1}, \A_{[j_1],i} \rangle \rangle \notag\\
        = & ~  \|f(x)_{j_1} \circ \A_{[j_1],i}\|_2^2 - \langle f(x)_{j_1}, \A_{[j_1],i} \rangle \cdot \langle f(x)_{j_1}^2, \A_{[j_1],i} \rangle \notag\\
        & ~ + \langle c(x)_{j_1}, f(x)_{j_1} \circ \A_{[j_1],l}^2 \rangle - \langle c(x)_{j_1}, f(x)_{j_1} \circ \A_{[j_1],i} \rangle \cdot \langle f(x)_{j_1} , \A_{[j_1],i} \rangle,
    \end{align}
    where the first step follows from the definition of the inner product, the second step follows from combining {\bf Part 7} and {\bf Part 8} of Lemma~\ref{lem:basic_derivatives}, the third step follows from the proof of Lemma~\ref{lem:second_derivative_f} (see Eq.~\eqref{eq:f_dxi2:1}), and the last step follows from Fact~\ref{fac:vector_properties}.

    Then, we compute the second term of Eq.~\eqref{eq:Lexp_dxi2}. 

    Note that 
    \begin{align*}
        \frac{\d (\langle c(x)_{j_1}, f(x)_{j_1} \rangle}{\d x_i} 
        = & ~ \langle \frac{\d c(x)_{j_1}}{\d x_i}, f(x)_{j_1} \rangle + \langle c(x)_{j_1}, \frac{\d f(x)_{j_1}}{\d x_i} \rangle\\
        = & ~ \langle f(x)_{j_1}, \frac{\d f(x)_{j_1}}{\d x_i} \rangle + \langle c(x)_{j_1}, \frac{\d f(x)_{j_1}}{\d x_i} \rangle\\
        = & ~ \langle f(x)_{j_1} + c(x)_{j_1}, f(x)_{j_1} \circ \A_{[j_1],i} - f(x)_{j_1} \cdot \langle f(x)_{j_1}, \A_{[j_1],i} \rangle \rangle\\
        = & ~ \langle f(x)_{j_1} + c(x)_{j_1}, f(x)_{j_1} \circ \A_{[j_1],i} \rangle - \langle f(x)_{j_1} + c(x)_{j_1}, f(x)_{j_1} \cdot \langle f(x)_{j_1}, \A_{[j_1],i} \rangle \rangle\\
        = & ~ \langle f(x)_{j_1} + c(x)_{j_1}, f(x)_{j_1} \circ \A_{[j_1],i} \rangle - \langle f(x)_{j_1} + c(x)_{j_1}, f(x)_{j_1} \rangle \cdot \langle f(x)_{j_1}, \A_{[j_1],i} \rangle,
    \end{align*}
    where the first step follows from the definition of the inner product, the second step follows from {\bf Part 8} of Lemma~\ref{lem:basic_derivatives}, the third step follows from {\bf Part 7} of Lemma~\ref{lem:basic_derivatives}, and the fourth and the fifth step follows from Fact~\ref{fac:vector_properties}.

    Therefore, the second term of Eq.~\eqref{eq:Lexp_dxi2} is:
    \begin{align}\label{eq:Lexp_dxi2:2}
        & ~ \frac{\d \langle c(x)_{j_1}, f(x)_{j_1} \rangle}{\d x_i} \cdot \langle f(x)_{j_1}, \A_{[j_1],i} \rangle \notag\\
        = & ~ \langle f(x)_{j_1} + c(x)_{j_1}, f(x)_{j_1} \circ \A_{[j_1],i} \rangle \cdot \langle f(x)_{j_1}, \A_{[j_1],i} \rangle - \langle f(x)_{j_1} + c(x)_{j_1}, f(x)_{j_1} \rangle \cdot \langle f(x)_{j_1}, \A_{[j_1],i} \rangle^2.
    \end{align}

    By applying the proof of Lemma~\ref{lem:second_derivative_alpha-1} (see Eq.~\eqref{eq:alpha_dxi2:second_term}), we can compute the third term of Eq.~\eqref{eq:Lexp_dxi2}
    \begin{align}\label{eq:Lexp_dxi2:3}
        \langle c(x)_{j_1}, f(x)_{j_1} \rangle \cdot \frac{\d\langle f(x)_{j_1}, \A_{[j_1],i}\rangle}{\d x_i}
        = & ~ \langle c(x)_{j_1}, f(x)_{j_1} \rangle \cdot (\langle f(x)_{j_1}, \A_{[j_1],i}^2 \rangle - \langle f(x)_{j_1}, \A_{[j_1],i} \rangle^2) \notag \\
        = & ~ \langle c(x)_{j_1}, f(x)_{j_1} \rangle \cdot \langle f(x)_{j_1}, \A_{[j_1],i}^2 \rangle \notag\\
        & ~ - \langle c(x)_{j_1}, f(x)_{j_1} \rangle \cdot \langle f(x)_{j_1}, \A_{[j_1],i} \rangle^2,
    \end{align}
    where the second step follows from simple algebra.

    Combining Eq.~\eqref{eq:Lexp_dxi2}, Eq.~\eqref{eq:Lexp_dxi2:1}, Eq.~\eqref{eq:Lexp_dxi2:2}, Eq.~\eqref{eq:Lexp_dxi2:3}, we have
    \begin{align*}
        \frac{\d^2 L_{\exp} }{\d x_i^2} 
        = & ~ \sum_{j_1=1}^n (\|f(x)_{j_1} \circ \A_{[j_1],i}\|_2^2 - \langle f(x)_{j_1}, \A_{[j_1],i} \rangle \cdot \langle f(x)_{j_1}^2, \A_{[j_1],i} \rangle \notag\\
        & ~ ~ ~ + \langle c(x)_{j_1}, f(x)_{j_1} \circ \A_{[j_1],l}^2 \rangle - \langle c(x)_{j_1}, f(x)_{j_1} \circ \A_{[j_1],i} \rangle \cdot \langle f(x)_{j_1} , \A_{[j_1],i} \rangle)\\ 
        & ~ - \sum_{j_1=1}^n (\langle f(x)_{j_1} + c(x)_{j_1}, f(x)_{j_1} \circ \A_{[j_1],i} \rangle \cdot \langle f(x)_{j_1}, \A_{[j_1],i} \rangle \\
        & ~ ~ ~ - \langle f(x)_{j_1} + c(x)_{j_1}, f(x)_{j_1} \rangle \cdot \langle f(x)_{j_1}, \A_{[j_1],i} \rangle^2) \notag\\
        & ~ - \sum_{j_1=1}^n  (\langle c(x)_{j_1}, f(x)_{j_1} \rangle \cdot \langle f(x)_{j_1}, \A_{[j_1],i}^2 \rangle \notag\\
        & ~ ~ ~ - \langle c(x)_{j_1}, f(x)_{j_1} \rangle \cdot \langle f(x)_{j_1}, \A_{[j_1],i} \rangle^2),
    \end{align*}

    Now, consider $\frac{\d^2 L_{\exp} }{\d x_i \d x_l}$.

    We have
    \begin{align}\label{eq:Lexp_dxixl}
        \frac{\d^2 L_{\exp} }{\d x_i \d x_l} 
        = & ~ \frac{\d}{\d x_l} (\frac{\d L_{\exp} }{\d x_i}) \notag\\
        = & ~ \frac{\d (\sum_{j_1=1}^n (\langle c(x)_{j_1}, f(x)_{j_1} \circ \A_{[j_1],i} \rangle - \langle c(x)_{j_1}, f(x)_{j_1} \rangle \cdot \langle f(x)_{j_1}, \A_{[j_1],i} \rangle))}{\d x_l} \notag\\
        = & ~ \frac{\d (\sum_{j_1=1}^n \langle c(x)_{j_1}, f(x)_{j_1} \circ \A_{[j_1],i} \rangle)}{\d x_l} - \frac{\d (\sum_{j_1=1}^n \langle c(x)_{j_1}, f(x)_{j_1} \rangle \cdot \langle f(x)_{j_1}, \A_{[j_1],i} \rangle)}{\d x_l} \notag\\
        = & ~ \sum_{j_1=1}^n \frac{\d (\langle c(x)_{j_1}, f(x)_{j_1} \circ \A_{[j_1],i} \rangle)}{\d x_l} - \sum_{j_1=1}^n \frac{\d (\langle c(x)_{j_1}, f(x)_{j_1} \rangle \cdot \langle f(x)_{j_1}, \A_{[j_1],i} \rangle)}{\d x_l} \notag\\
        = & ~ \sum_{j_1=1}^n \frac{\d (\langle c(x)_{j_1}, f(x)_{j_1} \circ \A_{[j_1],i} \rangle)}{\d x_l} \notag\\
        & ~ - \sum_{j_1=1}^n \frac{\d \langle c(x)_{j_1}, f(x)_{j_1} \rangle}{\d x_l} \cdot \langle f(x)_{j_1}, \A_{[j_1],i} \rangle \notag\\
        & ~ - \sum_{j_1=1}^n  \langle c(x)_{j_1}, f(x)_{j_1} \rangle \cdot \frac{\d\langle f(x)_{j_1}, \A_{[j_1],i}\rangle}{\d x_l},
    \end{align}
    where the first step follows from simple algebra, the second step follows from {\bf Part 9} of Lemma~\ref{lem:basic_derivatives}, the third step follows from the property of the summation, the fourth step follows from Fact~\ref{fac:derivative_rules}, and the last step follows from Fact~\ref{fac:derivative_rules}.

    First, we compute the first term of Eq.~\eqref{eq:Lexp_dxixl}:
    \begin{align}\label{eq:Lexp_dxixl:1}
        & ~ \frac{\d (\langle c(x)_{j_1}, f(x)_{j_1} \circ \A_{[j_1],i} \rangle)}{\d x_l} \notag \\
        = & ~  \langle \frac{\d c(x)_{j_1}}{\d x_l}, f(x)_{j_1} \circ \A_{[j_1],i} \rangle + \langle c(x)_{j_1}, \frac{\d f(x)_{j_1} \circ \A_{[j_1],i}}{\d x_l} \rangle \notag\\
        = & ~  \langle f(x)_{j_1} \circ \A_{[j_1],l} - f(x)_{j_1} \cdot \langle f(x)_{j_1}, \A_{[j_1],l} \rangle, f(x)_{j_1} \circ \A_{[j_1],i} \rangle + \langle c(x)_{j_1}, \frac{\d f(x)_{j_1} \circ \A_{[j_1],i}}{\d x_l} \rangle \notag\\
        = & ~  \langle f(x)_{j_1} \circ \A_{[j_1],l} - f(x)_{j_1} \cdot \langle f(x)_{j_1}, \A_{[j_1],l} \rangle, f(x)_{j_1} \circ \A_{[j_1],i} \rangle \notag\\
        & ~ + \langle c(x)_{j_1}, f(x)_{j_1} \circ \A_{[j_1],i} \circ \A_{[j_1],l} - f(x)_{j_1} \circ \A_{[j_1],i} \cdot \langle f(x)_{j_1}, \A_{[j_1],l} \rangle \rangle \notag\\
        = & ~  \langle f(x)_{j_1} \circ \A_{[j_1],l}, f(x)_{j_1} \circ \A_{[j_1],i} \rangle - \langle f(x)_{j_1}, f(x)_{j_1} \circ \A_{[j_1],i} \rangle \cdot \langle f(x)_{j_1}, \A_{[j_1],l} \rangle \notag\\
        & ~ + \langle c(x)_{j_1}, f(x)_{j_1} \circ \A_{[j_1],i} \circ \A_{[j_1],l} \rangle + \langle c(x)_{j_1}, f(x)_{j_1} \circ \A_{[j_1],i} \rangle \cdot \langle f(x)_{j_1}, \A_{[j_1],l} \rangle,
    \end{align}
    where the first step follows from the definition of the inner product, the second step follows from combining {\bf Part 7} and {\bf Part 8} of Lemma~\ref{lem:basic_derivatives}, the third step follows from the proof of Lemma~\ref{lem:second_derivative_f} (see Eq.~\eqref{eq:f_dxixl:1}), and the last step follows from Fact~\ref{fac:vector_properties}.

    Then, we compute the second term of Eq.~\eqref{eq:Lexp_dxixl}. 

    Note that 
    \begin{align*}
        \frac{\d (\langle c(x)_{j_1}, f(x)_{j_1} \rangle}{\d x_l} 
        = & ~ \langle \frac{\d c(x)_{j_1}}{\d x_l}, f(x)_{j_1} \rangle + \langle c(x)_{j_1}, \frac{\d f(x)_{j_1}}{\d x_l} \rangle\\
        = & ~ \langle f(x)_{j_1}, \frac{\d f(x)_{j_1}}{\d x_l} \rangle + \langle c(x)_{j_1}, \frac{\d f(x)_{j_1}}{\d x_l} \rangle\\
        = & ~ \langle f(x)_{j_1} + c(x)_{j_1}, f(x)_{j_1} \circ \A_{[j_1],l} - f(x)_{j_1} \cdot \langle f(x)_{j_1}, \A_{[j_1],l} \rangle \rangle\\
        = & ~ \langle f(x)_{j_1} + c(x)_{j_1}, f(x)_{j_1} \circ \A_{[j_1],l} \rangle - \langle f(x)_{j_1} + c(x)_{j_1}, f(x)_{j_1} \cdot \langle f(x)_{j_1}, \A_{[j_1],l} \rangle \rangle\\
        = & ~ \langle f(x)_{j_1} + c(x)_{j_1}, f(x)_{j_1} \circ \A_{[j_1],l} \rangle - \langle f(x)_{j_1} + c(x)_{j_1}, f(x)_{j_1} \rangle \cdot \langle f(x)_{j_1}, \A_{[j_1],l} \rangle,
    \end{align*}
    where the first step follows from the definition of the inner product, the second step follows from {\bf Part 8} of Lemma~\ref{lem:basic_derivatives}, the third step follows from {\bf Part 7} of Lemma~\ref{lem:basic_derivatives}, and the fourth and the fifth step follows from Fact~\ref{fac:vector_properties}.

    Therefore, the second term of Eq.~\eqref{eq:Lexp_dxixl} is:
    \begin{align}\label{eq:Lexp_dxixl:2}
        & ~ \frac{\d \langle c(x)_{j_1}, f(x)_{j_1} \rangle}{\d x_i} \cdot \langle f(x)_{j_1}, \A_{[j_1],i} \rangle \notag\\
        = & ~ \langle f(x)_{j_1} + c(x)_{j_1}, f(x)_{j_1} \circ \A_{[j_1],l} \rangle \cdot \langle f(x)_{j_1}, \A_{[j_1],i} \rangle \notag\\
        - & ~ \langle f(x)_{j_1} + c(x)_{j_1}, f(x)_{j_1} \rangle \cdot \langle f(x)_{j_1}, \A_{[j_1],l} \rangle \cdot \langle f(x)_{j_1}, \A_{[j_1],i} \rangle.
    \end{align}

    By applying the proof of Lemma~\ref{lem:second_derivative_alpha-1} (see Eq.~\eqref{eq:alpha_dxixl:second_term}), we can compute the third term of Eq.~\eqref{eq:Lexp_dxixl}
    \begin{align}\label{eq:Lexp_dxixl:3}
        & ~ \langle c(x)_{j_1}, f(x)_{j_1} \rangle \cdot \frac{\d\langle f(x)_{j_1}, \A_{[j_1],i}\rangle}{\d x_l} \notag \\
        = & ~ \langle c(x)_{j_1}, f(x)_{j_1} \rangle \cdot (\langle f(x)_{j_1}, \A_{[j_1],i} \circ \A_{[j_1],l} \rangle - \langle f(x)_{j_1}, \A_{[j_1],l} \rangle \cdot \langle f(x)_{j_1}, \A_{[j_1],i} \rangle) \notag \\
        = & ~ \langle c(x)_{j_1}, f(x)_{j_1} \rangle \cdot \langle f(x)_{j_1}, \A_{[j_1],i} \circ \A_{[j_1],l} \rangle - \langle c(x)_{j_1}, f(x)_{j_1} \rangle \cdot \langle f(x)_{j_1}, \A_{[j_1],l} \rangle \cdot \langle f(x)_{j_1}, \A_{[j_1],i} \rangle,
    \end{align}
    where the second step follows from simple algebra.

    Combining Eq.~\eqref{eq:Lexp_dxixl}, Eq.~\eqref{eq:Lexp_dxixl:1}, Eq.~\eqref{eq:Lexp_dxixl:2}, Eq.~\eqref{eq:Lexp_dxixl:3}, we have
    \begin{align*}
        \frac{\d^2 L_{\exp} }{\d x_i \d x_l} 
        = & ~ \sum_{j_1=1}^n (\langle f(x)_{j_1} \circ \A_{[j_1],l}, f(x)_{j_1} \circ \A_{[j_1],i} \rangle - \langle f(x)_{j_1}, f(x)_{j_1} \circ \A_{[j_1],i} \rangle \cdot \langle f(x)_{j_1}, \A_{[j_1],l} \rangle \notag\\
        & ~ + \langle c(x)_{j_1}, f(x)_{j_1} \circ \A_{[j_1],i} \circ \A_{[j_1],l} \rangle + \langle c(x)_{j_1}, f(x)_{j_1} \circ \A_{[j_1],i} \rangle \cdot \langle f(x)_{j_1}, \A_{[j_1],l} \rangle) \notag \\
        - & ~ \sum_{j_1=1}^n (\langle f(x)_{j_1} + c(x)_{j_1}, f(x)_{j_1} \circ \A_{[j_1],l} \rangle \cdot \langle f(x)_{j_1}, \A_{[j_1],i} \rangle \notag\\
        & ~ - \langle f(x)_{j_1} + c(x)_{j_1}, f(x)_{j_1} \rangle \cdot \langle f(x)_{j_1}, \A_{[j_1],l} \rangle \cdot \langle f(x)_{j_1}, \A_{[j_1],i} \rangle) \notag\\
        - & ~ \sum_{j_1=1}^n  (\langle c(x)_{j_1}, f(x)_{j_1} \rangle \cdot \langle f(x)_{j_1}, \A_{[j_1],i} \circ \A_{[j_1],l} \rangle \notag\\ 
        & ~ - \langle c(x)_{j_1}, f(x)_{j_1} \rangle \cdot \langle f(x)_{j_1}, \A_{[j_1],l} \rangle \cdot \langle f(x)_{j_1}, \A_{[j_1],i} \rangle).
    \end{align*}
\end{proof}

\subsection{Hessian of A Single Loss}
\label{sub:hessian:loss}
This section marks the initiation of our Hessian computation for a single loss. The subsequent result is prominently featured in the preceding study \cite{dls23}. Our presentation illustrates that our work is an expanded iteration of the identical problem, scaled by a factor of $n$. Indeed, our approach precisely constitutes a tensor-based rendition and proposes the Hessian property iteratively across $n$ instances.
\begin{lemma}\label{lem:hessian_single_loss}
We have
\begin{itemize}
    \item 
    {\bf Part 1.}
    \begin{align*}
    \frac{\d^2 L_{\exp} }{\d x_i^2} 
    = & ~ ( - \langle f(x), A_{*,i}\rangle \cdot f(x) + f(x) \circ A_{*,i})^\top ( - \langle f(x), A_{*, i} \rangle \cdot f(x) + f(x) \circ A_{*, i}) \\
    + & ~ c^\top (2\langle f(x), A_{*,i}\rangle^2 \cdot f(x) - \langle f(x) \circ A_{*,i}, A_{*, i} \rangle \cdot f(x) - 2\langle f(x), A_{*, i} \rangle \cdot f(x) \circ A_{*,i})\\
    + & ~ c^\top f(x) \circ A_{*,i} \circ A_{*, i}.
\end{align*}
    \item 
    {\bf Part 2.} 
    \begin{align*}
    \frac{\d^2 L_{\exp} }{\d x_i \d x_j} 
    = & ~ ( - \langle f(x), A_{*,j}\rangle \cdot f(x) + f(x) \circ A_{*,j})^\top (-\langle f(x), A_{*, i} \rangle \cdot f(x) + f(x) \circ A_{*, i}) \\
    + & ~ c^\top (2\langle f(x), A_{*,i}\rangle \cdot \langle f(x), A_{*,j}\rangle \cdot f(x) - \langle f(x) \circ A_{*,j}, A_{*, i} \rangle \cdot f(x) \\
    - & ~ \langle f(x), A_{*, i} \rangle \cdot f(x) \circ A_{*,j} - \langle f(x), A_{*,j}\rangle \cdot f(x) \circ A_{*, i} + f(x) \circ A_{*,j} \circ A_{*, i})
\end{align*}
\end{itemize}
\end{lemma}

\ifdefined\isarxiv

For the completeness, we still provide a proof.
\begin{proof}
{\bf Proof of Part 1.}

Note that in \cite{dls23},
\begin{align*}
    \frac{\d L_{\exp}}{\d x_i} = (f(x) - b)^\top (- \langle f(x), A_{*, i} \rangle \cdot f(x) + f(x) \circ A_{*, i})\\
    \frac{\d (f(x) - b)}{\d x_i} = \frac{\d f(x)}{\d x_i} = -\langle f(x), A_{*,i}\rangle \cdot f(x) + f(x) \circ A_{*,i}\\
\end{align*}

Therefore, we have
\begin{align*}
    \frac{\d^2 L_{\exp} }{\d x_i^2} 
    = & ~ \frac{\d}{\d x_i} ((f(x) - b)^\top (- \langle f(x), A_{*, i} \rangle \cdot f(x) + f(x) \circ A_{*, i}))\\
    = & ~ (\langle f(x), A_{*,i}\rangle \cdot f(x) + f(x) \circ A_{*,i})^\top (\langle f(x), A_{*, i} \rangle \cdot f(x) + f(x) \circ A_{*, i}) \\
    + & ~ (f(x) - b)^\top \frac{\d}{\d x_i} (- \langle f(x), A_{*, i} \rangle \cdot f(x) + f(x) \circ A_{*, i})
\end{align*}

Analyzing the second term of the above equation, we have
\begin{align*}
    \frac{\d}{\d x_i} (- \langle f(x), A_{*, i} \rangle \cdot f(x))
    = & ~ - \frac{\d}{\d x_i} ( \langle f(x), A_{*, i} \rangle) \cdot f(x) - \langle f(x), A_{*, i} \rangle \cdot (-\langle f(x), A_{*,i}\rangle \cdot f(x) + f(x) \circ A_{*,i})\\
    = & ~ - \langle -\langle f(x), A_{*,i}\rangle \cdot f(x) + f(x) \circ A_{*,i}, A_{*, i} \rangle \cdot f(x) \\
    + & ~ \langle f(x), A_{*, i} \rangle^2 \cdot f(x) - \langle f(x), A_{*, i} \rangle \cdot f(x) \circ A_{*,i}\\
    = & ~ \langle f(x), A_{*,i}\rangle^2 \cdot f(x) - \langle f(x) \circ A_{*,i}, A_{*, i} \rangle \cdot f(x) \\
    + & ~ \langle f(x), A_{*, i} \rangle^2 \cdot f(x) - \langle f(x), A_{*, i} \rangle \cdot f(x) \circ A_{*,i}.
\end{align*}

And, we have
\begin{align*}
    \frac{\d}{\d x_i} (f(x) \circ A_{*, i})
    = & ~ (-\langle f(x), A_{*,i}\rangle \cdot f(x) + f(x) \circ A_{*,i}) \circ A_{*, i}\\
    = & ~ -\langle f(x), A_{*,i}\rangle \cdot f(x) \circ A_{*, i} + f(x) \circ A_{*,i} \circ A_{*, i}.
\end{align*}

Combining everything together, we have
\begin{align*}
    \frac{\d^2 L_{\exp} }{\d x_i^2} 
    = & ~ ( - \langle f(x), A_{*,i}\rangle \cdot f(x) + f(x) \circ A_{*,i})^\top ( - \langle f(x), A_{*, i} \rangle \cdot f(x) + f(x) \circ A_{*, i}) \\
    + & ~ c^\top (2\langle f(x), A_{*,i}\rangle^2 \cdot f(x) - \langle f(x) \circ A_{*,i}, A_{*, i} \rangle \cdot f(x) - 2\langle f(x), A_{*, i} \rangle \cdot f(x) \circ A_{*,i} + f(x) \circ A_{*,i} \circ A_{*, i})
\end{align*}

{\bf Proof of Part 2.}

We have
\begin{align*}
    \frac{\d^2 L_{\exp} }{\d x_i \d x_j} 
    = & ~ \frac{\d}{\d x_j} ((f(x) - b)^\top (- \langle f(x), A_{*, i} \rangle \cdot f(x) + f(x) \circ A_{*, i}))\\
    = & ~ (-\langle f(x), A_{*,j}\rangle \cdot f(x) + f(x) \circ A_{*,j})^\top (-\langle f(x), A_{*, i} \rangle \cdot f(x) + f(x) \circ A_{*, i}) \\
    + & ~ (f(x) - b)^\top \frac{\d}{\d x_j} (- \langle f(x), A_{*, i} \rangle \cdot f(x) + f(x) \circ A_{*, i})
\end{align*}

Analyzing the second term of the above equation, we have
\begin{align*}
    \frac{\d}{\d x_j} (- \langle f(x), A_{*, i} \rangle \cdot f(x))
    = & ~ - \frac{\d}{\d x_j} ( \langle f(x), A_{*, i} \rangle) \cdot f(x) - \langle f(x), A_{*, i} \rangle \cdot (-\langle f(x), A_{*,j}\rangle \cdot f(x) + f(x) \circ A_{*,j})\\
    = & ~ - \langle -\langle f(x), A_{*,j}\rangle \cdot f(x) + f(x) \circ A_{*,j}, A_{*, i} \rangle \cdot f(x) \\
    + & ~ \langle f(x), A_{*, i} \rangle \cdot \langle f(x), A_{*, j} \rangle \cdot f(x) - \langle f(x), A_{*, i} \rangle \cdot f(x) \circ A_{*,j}\\
    = & ~ \langle f(x), A_{*,i}\rangle \cdot \langle f(x), A_{*,j}\rangle \cdot f(x) - \langle f(x) \circ A_{*,j}, A_{*, i} \rangle \cdot f(x) \\
    + & ~ \langle f(x), A_{*, i} \rangle \cdot \langle f(x), A_{*, j} \rangle \cdot f(x) - \langle f(x), A_{*, i} \rangle \cdot f(x) \circ A_{*,j}.
\end{align*}

And, we have
\begin{align*}
    \frac{\d}{\d x_j} (f(x) \circ A_{*, i})
    = & ~ (-\langle f(x), A_{*,j}\rangle \cdot f(x) + f(x) \circ A_{*,j}) \circ A_{*, i}\\
    = & ~ -\langle f(x), A_{*,j}\rangle \cdot f(x) \circ A_{*, i} + f(x) \circ A_{*,j} \circ A_{*, i}.
\end{align*}

Combining everything together, we have
\begin{align*}
    \frac{\d^2 L_{\exp} }{\d x_i \d x_j} 
    = & ~ ( - \langle f(x), A_{*,j}\rangle \cdot f(x) + f(x) \circ A_{*,j})^\top (-\langle f(x), A_{*, i} \rangle \cdot f(x) + f(x) \circ A_{*, i}) \\
    + & ~ c^\top (2\langle f(x), A_{*,i}\rangle \cdot \langle f(x), A_{*,j}\rangle \cdot f(x) - \langle f(x) \circ A_{*,j}, A_{*, i} \rangle \cdot f(x) \\
    - & ~ \langle f(x), A_{*, i} \rangle \cdot f(x) \circ A_{*,j} - \langle f(x), A_{*,j}\rangle \cdot f(x) \circ A_{*, i} + f(x) \circ A_{*,j} \circ A_{*, i})
\end{align*}

\end{proof}
\else 

\fi

\subsection{Checking \texorpdfstring{$B_1$}{} and \texorpdfstring{$B_2$}{}}
\label{sub:hessian:b1_b2}

In this section, we introduce new notations $B_1(x)$ and $B_2(x)$ to simplify the Hessian as \cite{dls23}.

\begin{lemma}[\cite{dls23}]\label{lem:dls23}
    If the following conditions hold
    \begin{itemize}
        \item Let $B_1(x) \in \R^{n \times n}$ be a matrix satisfying 
        \begin{align*}
            A_{*, i}^\top B_1(x) A_{*, j} := ( - \langle f(x), A_{*,j}\rangle \cdot f(x) + f(x) \circ A_{*,j})^\top (-\langle f(x), A_{*, i} \rangle \cdot f(x) + f(x) \circ A_{*, i}).
        \end{align*}
        \item Let $B_2(x) \in \R^{n \times n}$ be a matrix satisfying 
        \begin{align*}
            A_{*, i}^\top B_1(x) A_{*, j} 
            := & ~ c^\top (2\langle f(x), A_{*,i}\rangle \cdot \langle f(x), A_{*,j}\rangle \cdot f(x) - \langle f(x) \circ A_{*,j}, A_{*, i} \rangle \cdot f(x) \\
            - & ~ \langle f(x), A_{*, i} \rangle \cdot f(x) \circ A_{*,j} - \langle f(x), A_{*,j}\rangle \cdot f(x) \circ A_{*, i} + f(x) \circ A_{*,j} \circ A_{*, i}).
        \end{align*}
    \end{itemize}

        Then, we have
        \begin{itemize}
            \item Part 1. 
            \begin{align*}
                \frac{\d^2 L}{\d x_i^2} = A_{*,i}^\top B_1(x) A_{*,i} + A_{*,i}^\top B_2(x) A_{*,i}.
            \end{align*}
            \item Part 2. 
            \begin{align*}
                \frac{\d^2 L}{\d x_i \d x_j} = A_{*,i}^\top B_1(x) A_{*,j} + A_{*,i}^\top B_2(x) A_{*,j}.
            \end{align*}
        \end{itemize}
\end{lemma}
\begin{proof}
    This Lemma follows directly from Lemma~\ref{lem:hessian_single_loss}.
\end{proof}
\section{Sketching}
\label{sec:sketching}

In Section~\ref{sub:sketching:alg}, we introduce the iterative sketching-based federated learning algorithm. In Section~\ref{sub:sketching:sk}, we present the $\mathsf{sk}/\mathsf{desk}$ via coordinate-wise embedding. In Section~\ref{sub:sketching:related}, we introduce the related work of sketching. In Section~\ref{sub:sketching:preli}, we introduce the basic definition and property of sketching. In Section~\ref{sub:sketching:upper}, we prove the upper bound of $\|\wt{g}^{t,k}_r\|_2^2$. In Section~\ref{sub:sketching:lower}, we prove the lower bound of $\ip{\wt{u}^{t,k}_r-w^*}{\wt{g}^{t,k}_r}$. In Section~\ref{sub:sketching:induction}, we introduce the induction tools. In Section~\ref{sub:federated:converge_tool:short}, we give the formal proof to show the convergence of our gradient coin system.

\subsection{Iterative Sketching-based Federated Learning Algorithm}
\label{sub:sketching:alg}

In this section, we introduce the iterative sketching-based federated learning algorithm proposed in \cite{swyz23} (see Algorithm~\ref{alg:interactive_sketching}). The algorithm leverages sketching matrices to address communication efficiency issues, ensuring that our gradient coin system operates efficiently.

\begin{algorithm}[!ht]
\caption{Iterative sketching-based federated learning Algorithm with $K$ local steps}\label{alg:interactive_sketching}
\begin{algorithmic}[1]
\Procedure{\textsc{IterativeSketchingFL}}{}
\State Each client initializes $w_0$ with the same seed
\For{$t = 1 \to T$} \Comment{$T$ denotes the total number of global steps}
\State Let $\text{user}^{(c)}$ be the one which first finishes the computation of Proof of Work.
\State {\color{blue} /* $\text{user}^{(c)}$ */} 

\If{$t = 1$}
\State $u_c^{t, 0} \gets x_0$
\Else
\State $u_c^{t, 0} \gets x_{t - 1} + \mathsf{desk}_t(\Delta\wt{x}_{t - 1})$ \Comment{$\mathsf{desk}_t : \R^{b_{\mathrm{sketch}}} \to \R^d$ de-sketch the change}
\EndIf
\State $w_t \gets u_c^{t, 0}$
\For{$k = 1 \to K$}
\State $u_c^{t, k} \gets u_c^{t, k - 1} - \eta \cdot \nabla f_c(u_c^{t, k-1})$
\EndFor
\State $\Delta x_c(t) \gets u_c^{t, K} - x_t$
\State $\text{user}^{(c)}$ sends $\mathsf{sk}_t(\Delta x_c(t))$ to other users \Comment{$\mathsf{sk}_t : \R^d \to \R^{b_{\mathrm{sketch}}}$ sketch the change}
\State $\Delta x(t) \gets \Delta x_c(t)$
\State $\Delta \wt{x}_t \gets \eta \cdot \mathsf{sk}_t(\Delta x(t))$ \Comment{$\Delta \wt{x}_t \in \R^d$}
\State $\text{user}^{(c)}$ sends $\Delta \wt{x}_t$ to each client
\EndFor
\EndProcedure
\end{algorithmic}
\end{algorithm}

\subsection{\texorpdfstring{$\mathsf{sk}/\mathsf{desk}$}{}}

\label{sub:sketching:sk}

In this section, we introduce the $\mathsf{sk}/\mathsf{desk}$ via coordinate-wise embedding \cite{lsz19,sy21,jswz21,swyz23,qszz23,syyz23}. First, we give a formal definition of $a$-coordinate-wise embedding.

\begin{definition}[$a$-coordinate-wise embedding, Definition 4.1 in \cite{swyz23}]\label{def:coordinatewise_embedding}
    Let $R \in \R^{b_\mathrm{sketch} \times d}$ be a randomized matrix. 

    Let $g, h \in \R^d$ be two arbitrary vectors.

    $R$ satisfy $a$-coordinate wise embedding if
    \begin{align*}
        \E_{R \sim \Pi}[h^\top R^\top Rg] = h^\top g
    \end{align*}
    and 
    \begin{align*}
        \E_{R \sim \Pi}[(h^\top R^\top Rg)^2] \leq (h^\top g)^2 + \frac{a}{b_{\mathrm{sketch}}} \|h\|_2^2 \cdot \|g\|_2^2
    \end{align*}
\end{definition}

Definition~\ref{def:coordinatewise_embedding} can naturally connect the concept of coordinate-wise embedding with $\mathsf{sk}_t/\mathsf{desk}_t$ operators. This important definition may help us achieve the condition that any arbitrarily processed gradient $\mathsf{desk}_t \circ \mathsf{sk}_t(g)$ is ``close" to the true gradient of $g$ so that it can preserve the convergence of the algorithm. Typically, familiar sketching matrices tend to have a small constant value for their coordinate-wise embedding parameter $a$. If $h$ is a one-hot vector $e_i$, then the conditions of being $a$-coordinate wise embedding listed in Definition~\ref{def:coordinatewise_embedding} becomes
\begin{align*}
    \E_{R \sim \Pi}[R^\top Rg] = g
\end{align*}
and
\begin{align*}
    \E_{R \sim \Pi}[\|R^\top Rg\|_2^2] \leq (1 + a \cdot \frac{d}{b_{\mathrm{sketch}}}) \cdot \|g\|_2^2.
\end{align*}

Therefore, if we let the sketching be
\begin{align}\label{eq:sk_desk}
    \mathsf{sk}_t &= R_t \in \R^{b_{\mathrm{sketch}} \times d} 
\end{align}
and the de-sketching be
\begin{align}\label{eq:sk_desk_2}
    \mathsf{desk}_t &= R_t^\top \in \R^{d \times b_{\mathrm{sketch}}},
\end{align}
then for all iterations $t$ being greater than or equal to $1$, with independent random matrices $R_t$ having a sketching dimension of $b_{\mathrm{sketch}}$, we can get an unbiased sketching/de-sketching scheme and a variance which is bounded (see the following Theorem).

\begin{theorem}[Theorem 4.2 in \cite{swyz23}]\label{thm:sk_desk}

Let $t \in \Z_+$. 

Let $R_t$ be a list of arbitrary matrix in $\R^{b_{\mathrm{sketch}} \times d}$, and for each $t$, $R_t$ satisfies $a$-coordinate wise embedding property (see Definition~\ref{def:coordinatewise_embedding}).

Let $\mathsf{sk}_t$ and $\mathsf{desk}_t$ be defined by Eq.~\eqref{eq:sk_desk} and Eq.~\eqref{eq:sk_desk_2}.

Then, we can get that 1). for each iteration $t$, $(\mathsf{sk}_t, \mathsf{desk}_t)$ is independent, 2). $\mathsf{desk}_t$ and $\mathsf{sk}_t$ are both linear operators, 3). 
\begin{align*}
    \E[\mathsf{desk}_t(\mathsf{sk}_t(h))] = h,
\end{align*}
for each $h \in \R^d$, and 4). 
\begin{align*}
    \E[\|\mathsf{desk}_t(\mathsf{sk}_t(h))\|_2^2] \leq (1 + \alpha) \cdot \|h\|_2^2,
\end{align*}
for each $h \in \R^d$ and $\alpha = a \cdot d/b_{\mathrm{sketch}}$.

Additionally, for $\alpha > 0$, Table~\ref{tab:sketching_matrices} shows the typical sketching matrices.

    \begin{table}[!ht]
        \centering
        \begin{tabular}{|c|c|c|c|}
        \hline
           {\bf Reference}  & {\bf Sketching matrix}  & {\bf Definition}  & {\bf Param $\alpha$} \\
           
        \hline
            folklore  & Random Gaussian & Definition~\ref{def:Gaussian_matrix} &  $3d/b_{\mathrm{sketch}}$\\

        \hline
                   \cite{ldfu13}  & SRHT & Definition~\ref{def:SRHT} &  $2d/b_{\mathrm{sketch}}$\\
        \hline
           \cite{ams99}  & AMS sketch & Definition~\ref{def:AMS} &  $2d/b_{\mathrm{sketch}}$\\
           \hline
           \cite{ccfc02}  & Count-sketch & Definition~\ref{def:countsketch_matrix} &  $3d/b_{\mathrm{sketch}}$\\
     
        \hline
            \cite{nn13} & Sparse embedding & Definition~\ref{def:sparse_embedding_matrix_I}, \ref{def:sparse_embedding_matrix_II} &  $2d/b_{\mathrm{sketch}}$\\
          \hline

        \end{tabular}
        \caption{The $\alpha$ value (coordinate-wise embedding parameter) with the corresponding sketching matrix.}
        \label{tab:sketching_matrices}
    \end{table}
     
\end{theorem}

\subsection{Related Work}

\label{sub:sketching:related}

Sketching is a powerful tool that has been applied to numerous machine learning problems. Typically, there are two ways to apply sketching matrices. The first approach involves applying sketching once (or a constant number of times), known as ``sketch-and-solve". The second approach entails applying sketching in each iteration of the optimization algorithm while simultaneously designing a robust analysis framework. This is referred to as ``iterate-and-sketch". The present work falls into the second category.

Sketch-and-solve can be applied in various fields, including linear regression \cite{cw13,nn13}, low-rank approximation with Frobenius norm \cite{cw13,nn13,cswz23}, matrix CUR decomposition \cite{bw14,swz17,swz19}, weighted low-rank approximation \cite{rsw16}, entrywise $\ell_1$ norm low-rank approximation \cite{swz17,swz19_neurips_l1}, $\ell_p$ norm low-rank approximation \cite{cgk+17}, Schatten $p$-norm low rank approximation \cite{lw20}, $\ell_0$-norm low rank approximation \cite{bkw17}, tensor regression \cite{swyz21,rsz22,dssw18,djs+19}, tensor low-rank approximation \cite{swz19}, and general norm column subset selection \cite{swz19_neurips_general}.

Iterate-and-sketch has been applied to many fundamental problems, such as linear programming \cite{cls19,sy21,jswz21,dly21,gs22}, empirical risk minimization \cite{lsz19,qszz23}, support vector machines \cite{gsz23}, semi-definite programming \cite{gs22}, John's Ellipsoid computation \cite{ccly19,syyz22}, the Frank-Wolfe algorithm \cite{xss21,sxyz22}, reinforcement learning \cite{ssx23}, softmax-inspired regression \cite{dls23,gsy23_hyper,lsz23,ssz23}, federated learning \cite{swyz23}, the discrepancy problem \cite{dsw22,sxz22}, and non-convex optimization \cite{syz21,szz21,als+22,z22,gqsy22,hswz22}.

\subsection{Upper Bounding \texorpdfstring{$\|\wt{g}^{t,k}_r\|_2^2$}{}}
\label{sub:sketching:upper}

In this section, the upper bound of $\|\wt{g}^{t,k}_r\|_2^2$ is established.

\begin{lemma}\label{lem:gtk1}
Let $r \in [N]$.

Let $f_r:\R^d\rightarrow \R$ be a list of functions, and for each $r$, $f_r$ is $L$-smooth (see Definition~\ref{def:smooth}) and $\mu$-strongly convex (see Definition~\ref{def:app:strongly_convex}).

Then, we have
\begin{align*}
    \|\wt{g}^{t,k}_r\|_2^2 \leq ~4L (f(\wt{u}^{t,k}_r)-f(w^*))
\end{align*}

\end{lemma}
\begin{proof}

We have
\begin{align}\label{eq:grtk}
    \|\wt{g}^{t,k}_r\|_2^2 
    = & ~ \|\wt{g}^{t,k}_r-\nabla f(\wt{u}^{t,k}_r)+\nabla f(\wt{u}^{t,k}_r)\|_2^2 \notag\\
    \leq & ~ 2\|\wt{g}_r^{t,k}-\nabla f(\wt{u}^{t,k}_r)\|_2^2+2\|\nabla f(\wt{u}^{t,k}_r)\|_2^2,
\end{align}
where the first step is from simple algebra and the second step is by triangle inequality.

Note that 
\begin{align}\label{eq:grtk_1}
    \|\wt{g}^{t,k}_r-\nabla f(\wt{u}^{t,k}_r)\|_2^2 = 0.
\end{align}

Also, we have
\begin{align}\label{eq:grtk_2}
    \|\nabla f(\wt{u}^{t,k}_r)\|_2^2 
    = & ~ \|\nabla f(\wt{u}^{t,k}_r)-\nabla f(x^*)\|_2^2 \notag\\
    \leq & ~ 2L (f(\wt{u}^{t,k}_r)-f(x^*)).
\end{align}

Combining Eq.~\eqref{eq:grtk}, Eq.~\eqref{eq:grtk_1}, and Eq.~\eqref{eq:grtk_2}, we get
\begin{align*}
    \|\wt{g}^{t,k}_r\|_2^2 \leq & ~ 2L^2\|\wt{u}^{t,k}_r-x^*\|_2^2 \\
    \leq & ~ 4L (f(\wt{u}^{t,k}_r)-f(x^*))
\end{align*}
\end{proof}

\subsection{Lower Bounding \texorpdfstring{$\ip{\wt{u}^{t,k}_r-x^*}{\wt{g}^{t,k}_r}$}{}}
\label{sub:sketching:lower}

In this section, we find the lower bound of $\ip{\wt{u}^{t,k}_r-x^*}{\wt{g}^{t,k}_r}$.

\begin{lemma}\label{lem:gtk2}
Suppose each $f_r$ is $\mu$-strongly convex and L-smooth then
\begin{align*}
     \ip{\wt{u}^{t,k}_r-x^*}{\wt{g}^{t,k}_r} \geq & ~ f(\wt{u}^{t,k}_r)-f(x^*) + \frac{\mu}{2}\|\wt{u}^{t,k}_r-x^*\|_2^2
\end{align*}
\end{lemma}
\begin{proof}
 The lower bound on this inner product can be established as follows
\begin{align*}
    \ip{\wt{u}^{t,k}_r-x^*}{\wt{g}^{t,k}_r} = & ~ \ip{\wt{u}^{t,k}_r-x^*}{\nabla f_r(u_r^{t,k})} 
\end{align*}

We have
\begin{align*}
    \ip{\wt{u}_r^{t,k}-x^*}{\nabla f_r(\wt{u}_r^{t,k})} \geq & ~ f_r(\wt{u}_r^{t,k})-f_r(x^*) + \frac{\mu}{2}\|\wt{u}_r^{t,k}-x^*\|_2^2
\end{align*}

\end{proof}

\subsection{Induction Tools}
\label{sub:sketching:induction}
We introduce our induction tool in this section.
\begin{lemma}\label{lem:induction}
If the following conditions hold:
\begin{itemize}
    \item Suppose each $f_c$ satisfies Assumption~\ref{ass:app:strongly_convex}.
    \item   Let Theorem~\ref{thm:sk_desk} hold 
    \item $\eta \leq \frac{1}{8(1 + \alpha)LK}$, where $\alpha$ is defined as in Theorem~\ref{thm:sk_desk}.
    \item $R \sim \Pi$ is the distribution of sketching matrix.
\end{itemize}
for any $(t,k) \neq (1,0)$ and $r \sim [N]$, it follows that
\begin{align*}
    \E_{R \sim \Pi}[\| \wt{u}^{t,k}_r - x^*\|_2^2] 
    & ~ \leq (1 - \mu \eta) \E_{R \sim \Pi}[\| \wt{u}_r^{t, k-1} - x^* \|_2^2]  - \eta \E_{R \sim \Pi}[f(\wt{u}_r ^{t, k-1}) - f(x^*)] \\
    & ~  + {\bf 1}_{\{k=0\}} \eta^2 \alpha K ( 4L\sum_{i=0}^{K-1} \E_{R \sim \Pi}[f(\wt{u}_r^{t-1,i})-f(x^*)])
\end{align*}
    
\end{lemma}

\begin{proof}
We have for any $(t,k)\neq (1,0)$,
\begin{align*}
    \wt{u}_r^{t,k} = & ~ \wt{u}_r^{t,k-1} - \eta \cdot \wt{g}_r^{t,k-1} + 1_{\{k=0\}} \cdot \eta\cdot (I_d - \mathsf{desk}_t\circ\mathsf{sk}_t)(\sum_{i=0}^{K-1}\wt{g}_r^{t-1,i}).
\end{align*}
Therefore, denoting 
\begin{align}\label{eq:h}
    h^{t}:= (I_d - \mathsf{desk}_t\circ\mathsf{sk}_t)(\sum_{i=0}^{K-1}\wt{g}_r^{t-1,i})
\end{align}
we have
\begin{align}\label{eq:rt_step1}
    \|\wt{u}_r^{t,k} - x^*\|_2^2 
    = & ~ \| \wt{u}_r^{t,k-1} -x^* - \eta \cdot \wt{g}_r^{t,k-1} + 1_{\{k=0\}}\eta\cdot h^{t} \|_2^2\notag\\
    = & ~ \| \wt{u}_r^{t,k-1} -x^* \|_2^2 + \eta^2 \cdot \|\wt{g}_r^{t,k-1}\|_2^2 - 2\eta \ip{\wt{u}_r^{t,k-1} -x^*}{\wt{g}_r^{t,k-1}}\notag\\
    & ~ + 2\eta 1_{\{k=0\}}\ip{\wt{u}_r^{t,k-1} -x^*}{h^{t}} - 2 \eta^2 1_{\{k=0\}}\ip{\wt{g}_r^{t,k-1}}{h^{t}}\notag\\
    &~ + \eta^2 1_{\{k=0\}}\cdot \|h^{t}\|_2^2,
\end{align}
where the first step follows from the definition of $\wt{u}_r^{t,k}$ (see Algorithm~\ref{alg:interactive_sketching}), and the second step follows from the Pythagorean Theorem.

For any vector $h$, we have
\begin{align*}
    \E[ \mathsf{desk}_t( \mathsf{sk}_t ( h ) ) ]  = h, \qquad \E[ \| \mathsf{desk}_t( \mathsf{sk}_t ( h ) )\|_2^2] \leq   (1+\alpha) \cdot \| h \|_2^2
\end{align*}

Hence, we take expectation over Eq.~\eqref{eq:rt_step1},
\begin{align}\label{eq:E_u-w}
    \E[\|\wt{u}_r^{t,k}-x^*\|_2^2 ~|~ \mathcal{F}_{t}] 
    = & ~ \E[\| \wt{u}_r^{t,k-1} -x^* \|_2^2 ~|~ \mathcal{F}_{t}] + \eta^2 \cdot \E[\|\wt{g}_r^{t,k-1}\|_2^2 ~|~ \mathcal{F}_{t}] \notag\\
    & ~  - 2\eta \E[\ip{\wt{u}_r^{t,k-1} -x^*}{\wt{g}_r^{t,k-1}} ~|~ \mathcal{F}_{t}] + 1_{\{k=0\}}\cdot \eta^2 \cdot \E[\|h^{t}\|_2^2 ~|~ \mathcal{F}_{t}]
\end{align}

The two inner products involving $h^t$ vanishes due to the reason that $\E[h^t ~|~ {\cal F}_t]=0$.

Since
\begin{align*}
    \E[\|h^{t}\|_2^2 ~|~ \mathcal{F}_{t}]
    = &~\E[\|(I_d - \mathsf{desk}_t\circ\mathsf{sk}_t)(\sum_{i=0}^{K-1}\wt{g}_r^{t-1,i})\|_2^2 ~|~ \mathcal{F}_{t}]\\
    \leq &~ \alpha\E[\|\sum_{i=0}^{K-1}\wt{g}_r^{t-1,i}\|_2^2 ~|~ \mathcal{F}_{t}]\\
    \leq & ~ \alpha K\sum_{i=0}^{K-1}\E[\|\wt{g}_r^{t-1,i}\|_2^2 ~|~ \mathcal{F}_{t}],
\end{align*}
where the first step follows from the definition of $h^t$ (see Eq.~\eqref{eq:h}), the second step follows from $\|I_d - \mathsf{desk}_t\circ\mathsf{sk}_t)\|_2^2 \leq \alpha$, and the last step follows from the linearity property of expectation.

It follows that
\begin{align*}
    & ~ \E[\|\wt{u}_r^{t,k}-x^*\|_2^2] \\
    \leq & ~ \E[\| \wt{u}_r^{t,k-1} -x^* \|_2^2] + \eta^2 \cdot \E[\|\wt{g}_r^{t,k-1}\|_2^2]- 2\eta \E[\ip{\wt{u}_r^{t,k-1} -x^*}{\wt{g}_r^{t,k-1}}]\\
    & ~   + 1_{\{k=0\}}\cdot \eta^2 \cdot \alpha K\sum_{i=0}^{K-1}\E[\|\wt{g}_r^{t-1,i}\|_2^2]\\
    \leq & ~ \E[\| \wt{u}_r^{t,k-1} -x^* \|_2^2] + \eta^2 \cdot \E[4L (f(\wt{u}_r^{t,k-1})-f(x^*))]\\
    & ~ - 2\eta \E[f(\wt{u}_r^{t,k-1})-f(x^*)+ \frac{\mu}{2}\|\wt{u}_r^{t,k-1}-x^*\|_2^2] \\
    & ~ + 1_{\{k=0\}}\cdot \eta^2 \cdot \alpha K\sum_{i=0}^{K-1}\E[4L (f(\wt{u}_r^{t-1,i})-f(x^*))]\\
    \leq & ~ (1-\mu\eta)\E[\|\wt{u}_r^{t,k-1}-x^*\|_2^2]\\
    & ~ - 2\eta\cdot (1-2\eta L) \cdot \E[f(\wt{u}_r^{t,k-1}) - f(x^*)] \\
    & + 1_{\{k=0\}} \cdot \eta^2 \cdot \alpha  K\cdot  \Big(4L\sum_{i=0}^{K-1} \E[f(\ov{u}^{t-1,i}) - f(x^*)]\Big) 
\end{align*}
where the first step follows from Eq.~\eqref{eq:E_u-w}, the second step follows from Lemma~\ref{lem:gtk1} and Lemma~\ref{lem:gtk2}, and the last step follows from simple algebra.

Since $\eta\leq \frac{1}{4L}$, we have
\begin{align*}
    \E[\|\wt{u}_r^{t,k}-x^*\|_2^2] \leq & ~ (1-\mu\eta)\E[\|\wt{u}_r^{t,k-1}-x^*\|_2^2] - \eta \E[f(\wt{u}_r^{t,k-1}) - f(x^*)] \\
    & + 1_{\{k=0\}} \eta^2 \alpha K \Big(4L\sum_{i=0}^{K-1} \E[f(\wt{u}_r^{t-1,i}) - f(x^*)]\Big) 
\end{align*}
\end{proof}
\subsection{Convergence}
\label{sub:federated:converge_tool:short}

Once the aforementioned assumptions are established, we will ensure the convergence of our gradient coin design.

\begin{lemma}\label{thm:converge_loss:short}
    If the following conditions hold:
    \begin{itemize}
        \item Assumption~\ref{ass:app:strongly_convex} holds, where $\mu$ and $L$ are defined as in Assumption~\ref{ass:app:strongly_convex}.
        \item Let $K$ be the amount of the local steps.
        \item Let Theorem~\ref{thm:sk_desk} hold and $\eta \leq \frac{1}{8(1 + \alpha)LK}$, where $\alpha$ is defined as in Theorem~\ref{thm:sk_desk}.
        \item Let $x_0$, $x_{T + 1}$ be defined as in Algorithm~\ref{alg:interactive_sketching}.
        \item Let $\sigma^2 = \frac{1}{N} \sum_{c = 1}^N \|\nabla f_c(x^*)\|^2$. 
        \item $R \sim \Pi$ is the distribution of sketching matrix.
    \end{itemize}

    Then, we have
    \begin{align*}
        \E[f(x_{T + 1}) - f(x^*)] \leq & ~ \frac{L}{2} \E[\|x_0 - x^*\|_2^2] e^{- \mu \eta T}
    \end{align*}
    where $x^*$ is a minimizer of Definition~\ref{def:f(w)}.
\end{lemma}

\begin{proof}
By using Lemma~\ref{lem:induction} for $k$ times from $0$ to $K-1$, for any $t \geq 1$, it follows that
\begin{align*}
    & ~ (\E_{R \sim \Pi }[\| \wt{u}_r^{t + 1, 0} - x^* \|_2^2] + \sum_{k = 1}^{K - 1} \E_{R \sim \Pi }[\| \wt{u}_r^{t, k} - x^* \|_2^2]) - (1 - \mu \eta) \sum_{k = 0}^{K - 1} \E_{R \sim \Pi }[\| \wt{u}_r^{t, k} - x^* \|_2^2] \\
    \leq & ~ - \eta \sum_{k = 0}^{K - 1} \E_{R \sim \Pi }[f(\wt{u}_r^{t, k}) - f(x^*)] + \sum_{k = 0}^{K - 1} \mathbf{1}_{k = 0} \eta^2 \alpha K ( 4L \sum_{i = 0}^{K - 1} \E_{R \sim \Pi }[f(\wt{u}_r^{t, i}) - f(x^*)])\\
    = & ~ - \eta \sum_{k = 0}^{K - 1} \E_{R \sim \Pi }[f(\wt{u}_r^{t, k}) - f(x^*)]+ \eta^2 \alpha K( 4L \sum_{i = 0}^{K - 1} \E_{R \sim \Pi }[f(\wt{u}_r^{t, i}) - f(x^*)])\\
    = & ~ - \eta(1 - 4\eta \alpha LK) \sum_{k = 0}^{K - 1} \E_{R \sim \Pi }[f(\wt{u}_r^{t, k}) - f(x^*)]\\
    \leq & ~ - \frac{1}{2} \eta\sum_{k = 0}^{K - 1} \E_{R \sim \Pi }[f(\wt{u}_r^{t, k}) - f(x^*)],
\end{align*}
where the first step follows from Lemma~\ref{lem:induction}, the second step follows from simple algebra, the third step follows from simple algebra, and the last step follows from $\eta \leq \frac{1}{8 LK}$. 

Rearranging the terms, we obtain
\begin{align*}
    \E_{R \sim \Pi }[\| \wt{u}_r^{t + 1, 0} - x^* \|_2^2] \leq (1 - \mu \eta) \E_{R \sim \Pi }[\| \wt{u}_r^{t, 0} - x^* \|_2^2]
\end{align*}

Now, we will have
\begin{align*}
\E_{r \sim [N], R \sim \Pi}[\| \wt{u}_{r}^{t + 1, 0} - x^* \|_2^2] \leq (1 - \mu \eta) \E_{r \sim [N], R \sim \Pi}[\| \wt{u}_{r}^{t, 0} - x^* \|_2^2]
\end{align*}
implying
\begin{align}\label{eq:E_u-x*}
    \E_{r \sim [N],R \sim \Pi}[\| \wt{u}_r^{t + 1, 0} - x^* \|_2^2] \leq (1 - \mu \eta) (\E_{r \sim [N],R \sim \Pi}[\| \wt{u}_r^{t, 0} - x^* \|_2^2].
\end{align}

Therefore, we have
\begin{align}\label{eq:E_x_T+1_x*}
    \E_{r \sim [N],R \sim \Pi}[\| x^{T + 1} - x^* \|_2^2]
    \leq & ~ (1 - \mu \eta)^T (\E_{r \sim [N],R \sim \Pi}[\| x^0 - x^* \|_2^2]) \notag\\
    \leq & ~ \E_{r \sim [N],R \sim \Pi}[\| x^0 - x^* \|_2^2] e^{- \mu \eta T},
\end{align}
where the first step follows from the iterating Eq.~\eqref{eq:E_u-x*} $T$ times, and the second step follows from $(1 - \mu \eta)^T \leq e^{- \mu \eta T}, ~\forall T > 0$.

Finally, by $L$-smoothness of function $f$, we obtain
\begin{align*}
    \E_{r \sim [N],R \sim \Pi}[f(x^{T + 1}) - f(x^*)]
    \leq & ~ \frac{L}{2} \E_{r \sim [N],R \sim \Pi}[\| x^{T + 1} - x^* \|_2^2]\\
    \leq & ~ \frac{L}{2} \E_{r \sim [N],R \sim \Pi}[\| x^0 - x^* \|_2^2] e^{- \mu \eta T},
\end{align*}
where the first step follows from the definition of $L$-smoothness (see Definition~\ref{def:smooth}) and the second step follows from Eq.~\eqref{eq:E_x_T+1_x*}.

\end{proof}

\section{Distributed/Federated Learning}
\label{sec:federated}

In Section~\ref{sub:app:federated:def}, we introduce the definition of $\mu$-strongly convex and $M$-lipschitz. In Section~\ref{sub:federated:convex}, we adapt the properties of strongly convex and combine that with our result developed earlier in this paper. In Section~\ref{sub:federated:lip}, we adapt the properties of Lipschitz and combine that with our result developed earlier in this paper. In Section~\ref{sub:federated:tools}, we introduce some properties from previous work. 

\subsection{Definitions}
\label{sub:app:federated:def}

\begin{definition}[$\mu$-Strongly  Convex]\label{def:app:strongly_convex}
We say a function $L: \R^d \rightarrow \R$ is a $\mu$-strongly convex if
\begin{align*}
    \nabla^2 L(x) \succeq \mu \cdot I_d,
\end{align*}
where $\mu \in \R$.
\end{definition}

\begin{definition}[$l$-Smooth]\label{def:smooth}
Let $x$ and $y$ be two arbitrary elements in $\R^d$. 

Let $l > 0$ be a real number. 

We say a function $L: \R^d \rightarrow \R$ is $l$-smooth if
\begin{align*}
    \| \nabla L(x) - \nabla L(y) \|_2 \leq l \cdot \| x- y \|_2
\end{align*}
(It is equivalent to saying the gradient of $L$ is $l$-Lipschitz)
\end{definition}

\begin{definition}[$M$-Lipschitz]\label{def:lipschitz}
Let $x$ and $y$ be two arbitrary elements in $\R^d$. 

Let $M > 0$ be a real number. 

We say a function $L: \R^d \rightarrow \R$ is $M$-Lipschitz if
\begin{align*}
    | L(x) - L(y) | \leq M \cdot \| x - y \|_2
\end{align*}
\end{definition}

Upon comparing the Hessian result presented in Lemma~\ref{lem:hessian_single_loss} of our work with the Hessian result outlined in Lemma~\ref{lem:dls23} in \cite{dls23}, it becomes evident that each individual instance of our derived Hessian follows the identical structure as the Hessian discussed in \cite{dls23}. (Our Hessian result can be viewed as a summation of $n$ instances discussed in \cite{dls23}.)

Furthermore, the paper \cite{dls23} establishes the properties of Lipschitz continuity and strongly convex for a single instance. Building upon this foundation, we intend to extend these theoretical findings to encompass a series of $n$ iterations. By doing so, we anticipate the emergence of the following outcomes.
\subsection{Strongly Convex}
\label{sub:federated:convex}

\begin{lemma}
[Strongly Convex]\label{lem:convex_L}
If the following conditions hold
If the following conditions hold
\begin{itemize}
    \item Let $L_{j_1} : \R^{d^2} \rightarrow \R$ be defined as Definition~\ref{def:L}.
    \item Let $L : \R^{d^2} \rightarrow \R$ be defined as Definition~\ref{def:L}
    \item Let $W = \diag(w) \in \R^{n \times n}$.
    \item Let $\A \in \R^{n^2 \times d^2}$.
    \item Let $\A_{[j]} \in \R^{n \times d^2}$ denote the $j$-th block of $\A \in \R^{n^2 \times d^2}$. 
    \item Let $W^2 \in \R^{n \times n}$ denote the matrix that $i$-th diagonal entry is $w_i^2$.
    \item Let $\sigma_{\min}( \A_{ [j]} )$ denote the minimum singular value of $\A_{[j]}$ for matrix $\A_{[j]} \in \R^{n \times d^2}$ for all $j \in [n]$.
    \item Let $\min_{i \in [n]} w_i^2 \geq 4 + \mu / ( \sigma_{\min}^2 ( \A_{[j]} ) n )$ for all $j \in [n]$
\end{itemize}
Then, we have 
\begin{itemize}
    \item $L_j$ is $\mu$-strongly convex with parameter $\mu/n$ for all $j \in [n]$.
    \item $L$ is $\mu$-strongly convex with $\mu$.
\end{itemize}
\end{lemma}
\begin{proof}

{\bf Proof of Part 1.}
Based on Lemma~6.3 in page 30 in \cite{dls23}, we have
\begin{align*}
                \frac{\d^2 L}{\d x_i^2} = A_{*,i}^\top B_1(x) A_{*,i} + A_{*,i}^\top B_2(x) A_{*,i} \succeq  \mu/n \cdot I_d
\end{align*}
and
\begin{align*}
                \frac{\d^2 L}{\d x_i \d x_j} = A_{*,i}^\top B_1(x) A_{*,j} + A_{*,i}^\top B_2(x) A_{*,j} \succeq \mu/n \cdot I_d
\end{align*}
The $L_j$ is $\mu$-strongly convex now. We will focus on the second part of proof.

{\bf Proof of Part 2.}
By iterating over $n$ times, we obtain the following summary
\begin{align*}
    \nabla^2 L(x) = \sum_{j=1}^n \nabla^2 L_j(x).
\end{align*}
And then we can have
\begin{align*}
     \nabla^2 L(x) \succeq \mu \cdot I_d
\end{align*}
 Thus the loss function $L$ is $ \mu $-strongly convex. 
 
 Now the proof is complete now.
\end{proof}

\subsection{Lipschitz}
\label{sub:federated:lip}

\begin{lemma}
[Lipschitz]\label{lem:smooth_L}
If the following conditions hold
\begin{itemize}
    \item Let $L_{j_1}$ be defined as Definition~\ref{def:L}.
    \item Let $L$ be defined as Definition~\ref{def:L}.
    \item Let $R > 4$.
    \item Let $\A \in \R^{n^2 \times d^2}$.
    \item Let $l = \exp( O(R^2 + \log (nd)))$.
\end{itemize}
Then, we have 
\begin{itemize}
    \item $L_{j_1}$ is $l/n$-smooth for all $j_1 \in [n]$.
    \item $L$ is $l$-smooth
\end{itemize}
\end{lemma}
\begin{proof}
{\bf Proof of Part 1.}

Using Part 1 of Corollary 2.3 in \cite{gsx23}, we have
\begin{align}\label{eq:L:lip}
    \| \nabla L_{j_1}(x) - \nabla L_{j_1}(y) \|_2 \leq l/n \cdot \| x - y \|_2
\end{align}
Now $L_{j_1}$ is $l/n$-smooth.

{\bf Proof of Part 2.}
Now will focus on the smooth property of our loss.

We have 
\begin{align*}
    \| \nabla L(x) - \nabla L(y)\|_2 
    \leq  & ~ \| \sum_{j_1 =1}^n \nabla L_{j_1}(x) - \sum_{j_1 =1}^n\nabla L_{j_1}(y) \|_2 \\
    \leq & ~ \sum_{j_1 =1}^n \|  \nabla L_{j_1}(x) - \nabla L_{j_1}(y) \|_2 \\
    \leq & ~ l \cdot \| x - y \|_2
\end{align*}
where the first step is due to  Lemma~\ref{lem:hessian_single_loss}, 
the second step follows from triangle inequality, and the third step is from Eq.~\eqref{eq:L:lip}.

Now $L$ is $l$-smooth and our proof is complete
\end{proof}

\subsection{Tools from previous work}
\label{sub:federated:tools}

\begin{definition}\label{def:f(w)}
    Consider a federated learning scenario  
    with $N$ clients and corresponding local losses $f_c : \R^d \to \R$, our goal is to find 
    \begin{align}
        \min_{w \in \R^d} f(w) := \frac{1}{N}\sum_{c = 1}^N f_c(w).
    \end{align}
\end{definition}

\begin{assumption}[Assumption 3.1 in \cite{swyz23}]\label{ass:app:strongly_convex}
    Each $f_c$ is $\mu$-strongly convex for $\mu \geq 0$ and $L$-smooth. That is, for all $x, y \in \R^d$,
    \begin{align*} 
        f_c(y) - f_c(x) + \langle y - x, \nabla f_c(x) \rangle \geq & ~ \frac{\mu}{2} \|y - x\|_2^2  \\
        f_c(y) - f_c(x) + \langle y - x, \nabla f_c(x) 
        \leq & ~ \frac{L}{2} \|y - x\|_2^2.
    \end{align*}
    (Note that by definition of strongly convex and convex, $\mu > 0$ denotes strongly convex, and $\mu = 0$ denotes convex.)
\end{assumption}

\section{Gradient Coin Analysis}
\label{sec:gradient_coin}
In this section, we use the induction method to demonstrate the correctness of gradient computation \cite{bs23}. 
\begin{definition}
    We define the following as the block gradient
    \begin{align*}
        \Delta x_1, \Delta x_2\cdots \Delta x_t 
    \end{align*}
\end{definition}
\begin{lemma}[Induction of Gradient Computation]Given the block gradient $\Delta x_1, \Delta x_2,\cdots,\Delta x_{t-1}$ and $x_0$, We have the following facts 
\begin{itemize}
    \item We can compute $\Delta x_t$ as the current step's gradient. 
\end{itemize}
\end{lemma}
\begin{proof}
We can obtain the current weight by
\begin{align*}
    x_t = x_0 + \eta_{\mathrm{local}}\sum_{i=1}^{t-1}\Delta w_i
\end{align*}
And then $user^{(c)}$ can compute $K$ steps gradients by $u_c^{t, k} \gets u_c^{t, k - 1} - \eta_{\mathrm{local}} \cdot \nabla f_c(u_c^{t, k-1})$.

Finally, we can have $\Delta x_t$.
\end{proof}

\begin{algorithm}[!ht]
  \caption{ Gradient Block}\label{alg:gradient_block}
  \begin{algorithmic}[1]
    \State {\bf  datastructure} \textsc{GradientBlock} \Comment{See Definition~\ref{def:gradinent_block}}
   \State  {\bf members}
    \State  \hspace{4mm} \textsc{GradientBlock} prevhash \Comment{Used to link the prior block}
    \State \hspace{4mm} $\Delta x_t \in \R^{d \times d}$
    \State \hspace{4mm} $t \in \R$ \Comment{The index number of the gradient block}
    \State  \hspace{4mm} $\{\text{transaction}^{(i)}\}_{i=1}^k$ \Comment{List of transactions}
    \State {\bf end member}
    \Procedure{Initialize}{$t$,$x$, \textsc{GradientBlock} prevhash}
    \State prevhash $\gets$ prevhash
    \State $t \gets t$
    \State $\Delta x_t \gets x$ 
    \EndProcedure
    \Procedure{AddTrans}{\textsc{User} \text{user},$\{\text{transactions}^{(j)}\}_{j=1}^k$}
    \State $\text{user}$ collects $\{\text{transactions}^{(j)}\}_{j=1}^k$ into this block
    \EndProcedure
    \State {\bf end datastructure}
  \end{algorithmic}

\end{algorithm}

\begin{algorithm}[!ht]
  \caption{ Chain of Gradient Block}\label{alg:chain_gradient_block}
  \begin{algorithmic}[1]
    \State {\bf  datastructure} \textsc{GradBlockChain} \Comment{See Definition~\ref{def:chain_gradient_block}}
   \State  {\bf members}
    \State  \hspace{4mm} \textsc{GradientBlock} frontblock
    \State \hspace{4mm}  \textsc{GradientBlock} currentblock
    \State \hspace{4mm}   $t \in \R$ \Comment{Current Step}
    \State {\bf end members}
    \Procedure{Add}{$t$,$w$}
    \State prevbock $\gets$ currentblock
    \State \textsc{GradientBlock} block
    \State block.$\textsc{Initialize}$($w$,$t$,prevblock)
    \State currentblock $\gets$ block
    \State $t \gets t+1$
    \EndProcedure
    \State {\bf end datastructure}
  \end{algorithmic}
\end{algorithm}


\ifdefined\isarxiv
\bibliographystyle{alpha}
\bibliography{ref}
\else

\fi


\graphicspath{{./figs/}}



\end{document}